\documentclass[10pt]{article} 
\usepackage[preprint]{tmlr}

\usepackage{amsmath,amsfonts,bm}

\def\eqref#1{equation~\ref{#1}}

\def\1{\bm{1}}

\DeclareMathAlphabet{\mathsfit}{\encodingdefault}{\sfdefault}{m}{sl}
\SetMathAlphabet{\mathsfit}{bold}{\encodingdefault}{\sfdefault}{bx}{n}

\usepackage{hyperref}
\usepackage{url}            
\usepackage{booktabs}       
\usepackage{amsfonts}       
\usepackage{nicefrac}       
\usepackage{microtype}      
\usepackage{xcolor}         
\usepackage{graphicx}
\usepackage{tikz}
\usepackage{amsmath}
\usepackage{adjustbox}
\usepackage{multirow}
\usepackage{makecell}
\usepackage{bbm}
\usepackage{subcaption}
\usepackage{amsthm}
\usepackage{threeparttable}

\definecolor{codegreen}{rgb}{0,0.6,0}
\definecolor{codegray}{rgb}{0.5,0.5,0.5}
\definecolor{codepurple}{rgb}{0.58,0,0.82}
\definecolor{backcolour}{rgb}{0.97,0.97,0.94}
\definecolor{c0}{HTML}{1d1d1d}
\definecolor{c1}{HTML}{173359}
\definecolor{c2}{HTML}{8f5a39}
\definecolor{c3}{HTML}{a8332b}
\definecolor{c4}{HTML}{4c825c}
\definecolor{c5}{HTML}{0037a6}

\usepackage{listings}
\lstdefinestyle{mystyle}{
    language=Python,
    backgroundcolor=\color{backcolour},   
    commentstyle=\it\ttfamily\color{codegreen},
    keywordstyle=\color{c3},
    numberstyle=\tiny\color{codegray},
    stringstyle=\color{c5},
    showstringspaces=false,
    basicstyle=\ttfamily\scriptsize,
    breakatwhitespace=true, 
    keepspaces=true,                 
    breaklines=true,
    numbers=none,                                     
    morekeywords={nn,torch, init, empty, eye_, requires_grad, requires_grad_, zeros_, __init__, matmul, Module, roberta, encoder, layer, attention},
    emph={tt_config},
    emphstyle={\color{codepurple}}
}
\usepackage{tcolorbox}
\usepackage{algorithm,algpseudocode, soul}
\algrenewcommand{\algorithmiccomment}[1]{\hfill \textcolor{gray}{$\vartriangleright$ \textit{#1}}}

\usepackage[nameinlink,capitalize]{cleveref}

\definecolor{revcolor}{RGB}{0,0,200}
\newcommand{\rev}[1]{{\color{revcolor}#1}}
\renewcommand{\rev}[1]{#1}
\usepackage{tikz}
\usepackage{xspace}

\newcommand{\rbase}{RoBERTa$_\text{base}$\xspace}
\newcommand{\rlarge}{RoBERTa$_\text{large}$\xspace}

\makeatletter
\def\hlinewd#1{%
\noalign{\ifnum0=`}\fi\hrule \@height #1 \futurelet
\reserved@a\@xhline}
\makeatother

\crefname{equation}{Equation}{Eqs.}
\crefname{figure}{Figure}{Figs.}
\crefname{table}{Table}{Tabs.}
\crefname{appendix}{Appendix}{Apps.}
\crefname{algorithm}{Algorithm}{Algs.}

\title{MetaTT: A Global Tensor-Train Adapter for Parameter-Efficient Fine-Tuning}

\author{%
  Javier Lopez-Piqueres\thanks{Email: \texttt{javier.lopezpiqueres@jpmchase.com}} \quad Pranav Deshpande \quad Archan Ray \\
  Mattia J. Villani \quad Marco Pistoia \quad Niraj Kumar\thanks{Principal investigator. Email: \texttt{niraj.x7.kumar@jpmchase.com}} \\
  \\
  Global Technology Applied Research, JPMorgan Chase, New York, NY 10001, USA
}

\def\month{06}  
\def\year{2026} 
\def\openreview{\url{https://openreview.net/forum?id=1HdcPWfA9s}} 

\begin{document}

\maketitle

\begin{abstract}

We present MetaTT, a Tensor Train (TT) adapter framework for fine-tuning of pre-trained transformers. MetaTT enables flexible and parameter-efficient model adaptation by using a single shared TT to factorize transformer sub-modules. This factorization indexes key structural dimensions, including layer and matrix type, and can optionally incorporate heads and tasks. This design allows MetaTT’s parameter count to scale with the sum, rather than the product, of the modes, resulting in a substantially more compact adapter. Our benchmarks compare MetaTT with LoRA along with recent state-of-the-art matrix and tensor decomposition based fine-tuning methods. We observe that when tested on single-task standard language modeling benchmarks, MetaTT achieves competitive parameter efficiency to accuracy tradeoff. We further demonstrate that MetaTT performs competitively when compared to state-of-the-art methods on multi-task learning. Finally, we leverage the TT decomposition to design a rank adaptive optimizer inspired by the DMRG method from many-body physics. Our results demonstrate that integrating this approach with AdamW enhances optimization performance for a specified target rank. 
\end{abstract}

\section{Introduction}\label{sec:intro}

The sheer size of today's pre-trained models (e.g., LLaMA-$3$~\cite{grattafiori2024llama}, Gemini-$1.5$~\cite{team2024gemini}, GPT-$4$o~\cite{hurst2024gpt}, Falcon-$40$B~\cite{almazrouei2023falcon}, Mistral-$7$B~\cite{jiang2023mistral7b}, BERT~\cite{devlin2019bert}) coupled with the rapid rise of adapting them for specific tasks has proven to be a catalyst for the research on parameter efficient fine-tuning (PEFT) methods. Since \cite{aghajanyan2020intrinsic} demonstrated that pre-trained language models can effectively learn on a given task even when subjected to a random projection onto a smaller subspace, starting from LoRA \cite{hu2021lora}, a flurry of works on PEFT have demonstrated significant parameter reduction for fine-tuning large models on simpler and often single tasks \cite{karimi2021compacter, zhang2023adalora, zi2023delta, zhang2024riemannian, albert2025randlora}. 

To push beyond local layer-wise parameter reduction, sharing low-rank adapters across transformer layers have shown great promise. VeRA \cite{kopiczko2023vera} shares single pair of low-rank matrices across all layers and learns small scaling vectors; NOLA \cite{koohpayegani2023nola} re-parameterizes low-rank matrices as linear combinations of random bases, decoupling parameter count from rank and architecture; and VB-LoRA \cite{li2024vb} and Uni-LoRA \cite{li2025uni} construct all adapters from a global vector bank, achieving extreme parameter efficiency.

Subsequently, a wider class of methods have started considering the weight matrices (individually or a combination of them) as higher order tensors and then designing decompositions. To this end, several lines of work have emerged, partly because it is not obvious which components of the model benefit from tensor decompositions, and partly because unlike matrices, it is not known how to decompose tensors optimally \cite{kolda2009tensor}. As a result it remains open, even empirically, to understand how tensor-based decompositions can further improve the balance between the number of trainable parameters and downstream task performance during fine-tuning.

Methods compressing per layer adapters via tensor decompositions include LoRETTA \cite{yang2024loretta}, which at each layer replaces LoRA’s trainable matrices with a tensor train (TT) decomposition; TT-LoRA \cite{anjum2024tensor}, which similarly first folds each trainable matrix into a tensor and then factors them into TT decomposition. QuanTA \cite{chen2024quanta} and Quantum-PEFT \cite{koike2025quantum} further decompose adapters into tensor networks shaped as quantum circuits.

A promising avenue is to exploit \emph{both} shared adapters and tensor decompositions, which can a priori lead to higher compression rates compared to per-layer tensor decompositions, at higher expressibility compared to shared low-rank matrices. 
In FacT \cite{jie2023fact}, the authors use 3D TT and Tucker decompositions to capture parameter sharing across layers in the context of vision transformers. This idea is extended in the context of LLMs in LoTR \cite{bershatsky2024lotr}. Finally, LoRTA \cite{hounie2024lorta} decomposes the various linear layers in a transformer using a CP-decomposition. CP decompositions for higher order tensors generally have unique decompositions \cite{kolda2009tensor}, and thus finding the right decomposition may seem to be harder during fine-tuning.
Given the plethora of work on matrix and tensor decomposition based adapters we ask,
\begin{center}
    \emph{Can we achieve further parameter efficiency when fine-tuning transformer models?}
\end{center}
\paragraph{A global adapter via tensor trains:} Building on LoTR and FacT-TT, which stack and decompose all adapter layers into a single 3D TT, we further separate the layer and matrix type dimensions to form a 4D TT, enabling greater parameter reduction. Decomposing the multi-head self-attention (MHSA) output dimension into head dimension and number of heads yields a 5D tensor, as in LoRTA. Both 4D and 5D TTs maintain parameter efficiency, scaling linearly with the number of modes, unlike Tucker decomposition, which scales exponentially. Moreover, TTs can represent a broader class of tensors than CP decomposition, often requiring exponentially fewer parameters \cite{khrulkov2017expressive}. 

\paragraph{Multi-task learning via tensor trains:} In the context of adapter-based methods, single-task fine-tuning has been widely studied. Only recently the need for multi-task learning (MTL) has gained prominence. This is partly due to the size of the pre-trained models and partly due to the fact that often there are common modalities across datasets and tasks. Modifications to LoRA have been shown to work well for MTL, e.g, ensembling multiple LoRA adapters \cite{wang2023multilora}, using a mixture of experts (MoE-LoRA) \cite{liu2024moe}, and sharing one parameter across multiple tasks (MTL-LoRA) \cite{yang2025mtl}. We observe that a tensor based structure automatically extends to MTL and so given a tensor based adapter one can construct a unified adapter, which may lead to further parameter efficiency via parameter sharing across tasks, which has been shown to yield benefits in the setting of hypernetwork adapters \cite{karimi2021parameter}. Once adapters are represented in tensor form, this structure naturally extends to MTL: the task index can be incorporated as an additional tensor mode, yielding a unified adapter that promotes parameter efficiency. Building on this idea, we efficiently extend the TT adapters to the multi-task setting, and refer to this unified family of TT-based adapters as \emph{MetaTT}, reflecting their modular design that seamlessly accommodates the extra task dimension. 

Since the construction of such \textit{global} adapters can also be achieved efficiently by means of other tensor decompositions (such as CP), it becomes natural to ask,  
\begin{center}
    \emph{Do we achieve anything beyond parameter reduction when using MetaTT's architecture?}
\end{center}
To address this, we examine the unique optimization advantages offered by the TT structure.

\textbf{Rank adaptive training:} Unlike other tensor decompositions, TTs are equipped with powerful optimization routines that exploit the TT structure. Specifically, we apply a rank adaptive scheme inspired by the Density-Matrix Renormalization Group (DMRG) optimization \cite{schollwock2011density, verstraete2023density}, a method widely used in the context of quantum many-body physics, to improve training in the presence of many TT cores and adaptively choose the TT ranks during fine-tuning. Such a method does not trivially extend beyond the TT architecture.

\begin{figure}
    \centering
    \includegraphics[width=0.9\linewidth]{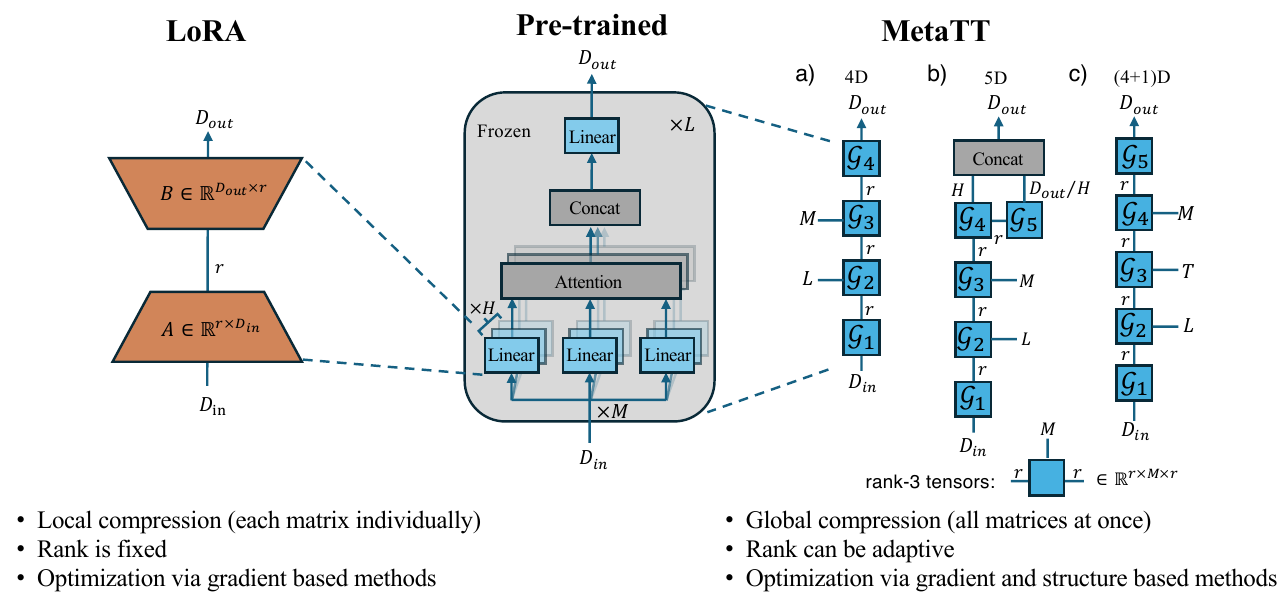}
    \caption{\textbf{Comparison between LoRA and MetaTT adapters.} While LoRA parameterizes each weight matrix individually, MetaTT parameterizes all linear maps in the transformer architecture jointly as a TT (here shown only for a MHSA block). We propose two architectures for single-task fine-tuning: a) MetaTT-4D decomposes the entire set of linear maps into a TT of order $4$ along the input/output dimensions (as in LoRA) as well as along the layer dimension, $L$, and the set of projection matrices, $M$. b) MetaTT-5D further decomposes the output dimension along the head dimension and number of heads. To capture task dependencies in multi-task learning, we extend MetaTT by adding an additional tensor core with mode dimension $T$ corresponding to the number of tasks, resulting in c) MetaTT-(4+1)D. Unlike LoRA, TT ranks in MetaTT can adapt during fine-tuning, providing both parameter efficiency and optimization flexibility.
    }
    \label{fig:main_fig}
    \vspace{-1em}
\end{figure}

\section{Meta-Adapter with Tensor Networks}
Tensor networks are mathematical structures that can represent high-dimensional tensors in a more manageable form. This is achieved by decomposing a large order tensor into a network of interconnected, generally lower-dimensional, tensors. This decomposition reduces the storage and computational requirements, making tensor networks suitable for applications involving big data.
\subsection{Tensor-Train Decomposition}
Among the various types of tensor networks, tensor trains (TTs) offer a particularly efficient representation. A TT decomposes a tensor $\mathcal{G} \in \mathbb{R}^{n_1\times \cdots \times n_d}$ of order $d$ into a set of rank-3 tensors as follows 
\begin{equation}
    \mathcal{G}[i_1,\cdots, i_d]=\mathcal{G}_1[i_1]\mathcal{G}_2[i_2]\cdots \mathcal{G}_d[i_d],
\end{equation}
where $\mathcal{G}_k[i_k]\in \mathbb{R}^{r_{k-1}\times r_k}$, $i_k=1,\cdots, n_k$, are matrices, except the first and last, which are row and column vectors, respectively. It is also customary to see $\mathcal{G}_{k=2,\cdots, d-1}$ as rank-3 tensors, also known as cores. The parameters $r_i$ are known as TT-ranks. The complexity of the TT decomposition is $\mathcal{O}(dr^2n)$ with $r=\max_k r_k$, $n=\max_k n_k$. A TT therefore offers a controllable trade-off between expressivity (via $r$) and storage. In what follows, we assume the TT-ranks are all of equal value $r$.

\subsection{Tensor Based Adapters}

LoRA type adapters inject a matrix $\Delta W \in \mathbb{R}^{D_{\rm in} \times D_{\rm out}}$ at every layer and a subset of projection matrices in the MHSA module. The set of all such adapters can be viewed as a $4$-dimensional tensor,
\begin{equation}
    \Delta \mathcal{W}_{4D}=\{\{\Delta W_{l,m}\}_{l=1}^L\}_{m=1}^M \in \mathbb{R}^{D_{\rm in} \times L \times M \times D_{\rm out}},
\end{equation}
where $L$ is the number of layers, and $M$ is the number of projection matrices, which can be between $1$ and $4$ (corresponding to $Q,K,V$, and $O$ matrices). One can choose to include the MLP matrices in this tensor after properly reshaping them. For instance in the BERT family of models, the two MLP matrices are of size $4D_{\rm in}\times D_{\rm out}$. However, including these MLP layers would increase $M$ to potentially $12$. This increases the computational overhead should we wish to construct a $3$-dimensional tensor by stacking the $L$-dimension on top of $M$. This approach was followed in FacT \cite{jie2023fact} and LoTR \cite{bershatsky2024lotr}.

Moreover, the output dimension in the MHSA is further split into $H$ number of heads. Thus implicitly, such a architecture allows for the construction of even a $5$-dimensional tensor,
\begin{equation}
    \Delta \mathcal{W}_{5D}=\{\{\{\Delta W_{l,m,h}\}_{l=1}^L\}_{m=1}^M\}_{h=1}^H \in \mathbb{R}^{D_{\rm in} \times L \times M \times H\times D_{\rm out}/H}.
\end{equation}
Tensors $\Delta \mathcal{W}_{4D}$ and $\Delta \mathcal{W}_{5D}$ potentially capture all adapters one could include for a transformer-like model. Furthermore, this idea extends beyond single transformer adapters. For instance, one could include an extra dimension capturing task dependency in a multi-task learning (MTL) setting, where different transformer adapters are used for different tasks
\begin{equation}
    \Delta \mathcal{W}_{6D}=\{\{\{\{\Delta W_{l,m,h,t}\}_{l=1}^L\}_{m=1}^M\}_{h=1}^H \}_{t=1}^T \in \mathbb{R}^{D_{\rm in} \times L \times M \times H \times T\times D_{\rm out}/H},
\end{equation}
where $T$ represents the total number of tasks.

\subsection{MetaTT Adapter} 
Consider the TT decomposition of $\Delta \mathcal{W}_{4D}$, which we refer to as MetaTT-4D, shown pictorially in \cref{fig:main_fig}. For fixed layer $l$ and $m$-th projection matrix, the TT decomposition results in a list of $4$ matrix multiplications of rank $r$ (assuming for simplicity each bond has the same fixed rank). It is low-rank if $r \ll \min\{D_{\rm in}, D_{\rm out}\}$. In other words, for an input batch $X\in \mathbb{R}^{N\times D_{\rm in}}$ and output batch $Y \in \mathbb{R}^{N \times D_{\rm out}}$, for every layer $l$ and $m$-th projection matrix we have:
\begin{align} \label{eq:TT_data}
\begin{split}
    Y&=X\cdot W_{l,m}^T+\alpha X\cdot\operatorname{ TT}(\Delta \mathcal{W}_{4D})_{l,m}\\
    &=X\cdot W_{l,m}^T+\alpha X\cdot\mathcal{G}_1\mathcal{G}_2[l]\mathcal{G}_3[m]\mathcal{G}_4,
    \end{split}
\end{align}
where $W_{l,m}^T$ is the transposed frozen linear layer (from the pre-trained model). While one can choose any permutations of the ordering of the TT cores, we present the arrangement that leads to the most compressed form. This entails assigning the input and output dimensions to each end of the TT since these are usually the largest dimensions in a transformer architecture. This is because they are only coupled by a single bond, thereby incurring an additive $O(D\times r)$ cost to the overall complexity while also being quadratic in $r$ for the other dimensions, which are usually orders of magnitude less than the input and output dimensions. 

The extension to capture other dimensions such as heads or tasks follows straightforwardly from this construction. The only difference is when accounting for heads one would further need to concatenate the number of heads and the head dimension into the output dimension. This is shown in \cref{fig:main_fig}. Importantly, minimal reshaping is required throughout this process, as shown explicitly for MetaTT-4D in \eqref{eq:TT_data}. This is in contrast to other tensor decompositions. The input data is processed in its original format and outputs dimensions are also consistent with the original model, facilitating the use of optimized matrix-vector GPU kernels and allowing for performance and scalability enhancement. While further compression can be achieved by further unrolling the input and output dimensions, it is crucial to avoid these, as they can complicate the decomposition and reduce computational efficiency \cite{monturiol2025tensorization, lu2025fetta}. 

\paragraph{Complexity Analysis.}
MetaTT-4D has $2Dr +(L+M)r^2$ parameters for $D=\max\{D_{\rm in}, D_{\rm out}\}$. Similarly, MetaTT-5D has $(D+D/H)r+(L+M+H)r^2$ parameters. This is substantially better than the LoRA adapter which requires at least $2LMDr$ parameters. Thus, by introducing a small series of $r \times r$ matrices we are able to significantly compress the tensor otherwise obtained by using LoRA. Note that for fixed TT-rank $r$, MetaTT-4D is more efficient than MetaTT-5D whenever $r>D/H(1-1/H)$. 

In the MTL setting, we consider the variant MetaTT-(4+1)D as shown in \cref{fig:main_fig}. This has parameter count $2Dr+(L+M+T)r^2$, which remains substantially smaller than having $T$ independent LoRAs (with $2MTLr$ number of parameters). Hence, MetaTT scales gracefully with task dimension.

Training times of MetaTT adapters are very competitive with LoRA. At each linear layer one performs $2(D\times r)+ 2(r\times r)$ matrix multiplications (for MetaTT-4D), where the complexity is dominated by $(D\times r)$ since $D\gg r$. As such the total time required to train the adapter is very similar to that of a LoRA adapter. During inference, one can match the speeds of LoRA by adding a single pre-computation step where one can merge the middle tensor cores with $\mathcal{G}_1$ or $\mathcal{G}_4$ (for MetaTT-4D) once the adapters are trained. 

\subsection{DMRG-Inspired Sweep: A Rank Adaptive Training Algorithm}

While the gold standard of training PEFT adapters has been gradient descent using optimizers like Adam \cite{kingma2014adam}, these methods fail to take advantage of the tensor decomposition structure. For matrices and other small order tensor decompositions, Adam works remarkably well. However, for higher order tensor decompositions, e.g., tensor networks with many cores, training using gradient descent can be unstable \cite{barratt2022improvements}. We propose the use of a rank adaptive scheme inspired by the DMRG method \cite{schollwock2011density, verstraete2023density}, a variational algorithm widely used in quantum many-body physics to optimize TTs (also known as matrix product states in that context) representing quantum wavefunctions.

Starting with a sufficiently high-rank TT, we train with Adam for a few epochs and then apply a compression layer composed of a series of SVD decompositions on neighboring merged tensor cores and keep only vectors corresponding to the largest $r$ singular values. While we use full SVD decomposition one can use approximate SVD \cite{halko2011finding, musco2015randomized, tropp2023randomized}. Alternatively, more sophisticated importance scores can also be used to compute low-rank approximations \cite{cohen2017input, musco2017recursive, zhang2023adalora}. We successively compute these low-rank approximations until a desired rank is achieved. We state this procedure in \cref{alg:DMRG} (this particular version of the algorithm is analogous to a TT-rounding sweep but involving two sites as opposed to one). Note that since the ranks change after calling DMRG, and thus the number of trainable weights, one must reinitialize Adam moments after each truncation.
\begin{algorithm}
    \caption{DMRG-inspired sweep}
    \label{alg:DMRG}
    \begin{algorithmic}[1]
        \Require MetaTT $\rm TT_{dD}(\Delta \mathcal{W})$ with ranks $r_0$, $d$ represents number of TT cores, target ranks $r$, and truncated SVD function $\operatorname{tSVD}$.
        \For{$i = 1$ to $d - 1$}
            \State $M\gets \operatorname{MERGE}(\mathcal{G}_i,\mathcal{G}_{i+1})$ \Comment{Merge adjacent cores, reshape into matrix $M$}
            \State $U,S,V^T = \operatorname{tSVD}(M; r)$  \Comment{Rank $r$ approximation using SVD}
            \State $\mathcal{G}_i \gets U; \mathcal{G}_{i+1} \gets SV^T$
        \EndFor
        \For{$i = d$ to $2$}
            \State $M\gets \operatorname{MERGE}(\mathcal{G}_{i-1},\mathcal{G}_i)$ \Comment{Merge adjacent cores, reshape into matrix $M$}
            \State $U,S,V^T = \operatorname{tSVD}(M; r)$ \Comment{Rank $r$ approximation using SVD}
            \State $\mathcal{G}_{i-1} \gets US; \mathcal{G}_{i} \gets V^T$
        \EndFor
        \Ensure MetaTT with ranks $r$. 
    \end{algorithmic}
\end{algorithm}

\section{Experiments}\label{sec:exps}
In this section, we perform three sets of experiments. In \cref{sec:stl} we test the performance of MetaTT in the context of single-task fine-tuning against state-of-the-art methods. Our focus here is on commonsense reasoning tasks using the setup of~\cite{hu2023llm}, and natural language understanding~\cite{wang2018glue}. In \cref{sec:mtl} we compare the performance of MetaTT when adding an extra tensor for capturing task-specific knowledge in the context of MTL. Finally, in \cref{sec:dmrg} we demonstrate that optimizing MetaTT using a variant of AdamW alternating with \cref{alg:DMRG} can further boost the performance of fine-tuning using MetaTT. 

\subsection{Single-Task Fine-Tuning}\label{sec:stl}

In this section we discuss the performance of various PEFT adapters along with MetaTT on single-task fine-tuning.
\begin{table}[ht]
    \centering
    \begin{adjustbox}{max width=\textwidth}
    \begin{tabular}{|c|lc*{9}{c}|}
        \hline
        & \textbf{Method} & \textbf{Param$\times 10^5$} &  \textbf{ARC-c} & \textbf{ARC-e} & \textbf{BoolQ} & \textbf{HellaSwag} & \textbf{OBQA} & \textbf{PIQA} & \textbf{SIQA} & \textbf{WinoGrande} & \textbf{Avg} \\
        \hline
        \multirow{8}{*}{\rotatebox{90}{Llama-2-7b}} 
        & Zero Shot  & -- & $46.5$ & $74.5$ & $74.7$ & $75.9$ & $47.0$ & $78.8$ & $46.1$ & $69.5$ & $64.1$\\
        & LoRA (r=8) & $41.9$ & $\mathbf{53.3(4)}$ & $79.1(1)$ & $\mathbf{80.8(5)}$ & $76.05(7)$ & $\mathbf{60.7(2)}$ & $80.0(1)$ & $\mathbf{54.5(6)}$ & $\mathbf{74.5(2)}$ & $\mathbf{69.86(2)}$ \\
        & LoRA (r=16) & $83.9$ & $\mathbf{52.6(4)}$ & $77.6(2)$ & $\mathbf{81(1)}$ & $\mathbf{76.3(3)}$ & $\mathbf{62(4)}$ & $\mathbf{80.1(2)}$ & $\mathbf{55.3(4)}$ & $\mathbf{75.5(5)}$ & $\mathbf{70.0(5)}$ \\
        & VeRA (r=1024) & $3.27$ & $47.8(3)$ & $76.2(3)$ & $75.8(6)$ & $\mathbf{76.26(3)}$ & $52.5(2)$ & $78.9(2)$ & $47.68(6)$ & $69.5(3)$ & $65.60(1)$ \\
        & LoTR (r=16) & $1.47$ & $51.2(4)$ & $\mathbf{80.4(1)}$ & $78(2)$ & $75.6(1)$ & $57(2)$ & $79.9(1)$ & $52.0(8)$ & $71.6(4)$ & $68.1(5)$  \\
        & \rev{LoTR (r=64)} & \rev{$7.86$} & \rev{$52.1(5)$} & \rev{$78.9(8)$} & \rev{$78(1)$} & \rev{$75.3(3)$} & \rev{$57(2)$} & \rev{$79.3(3)$} & \rev{$54.1(6)$} & \rev{$73.7(6)$} & \rev{$68.6(4)$} \\
        & MetaTT-4D (r=16) & $1.40$ & $50.2(3)$ & $79.15(8)$ & $77(2)$ & $75.4(1)$ & $58(3)$ & $79.7(2)$ & $51.1(2)$ & $71.8(4)$ & $67.8(4)$  \\
        & MetaTT-4D (r=64) & $6.6$ & $51.8(3)$ & $\mathbf{80.6(9)}$ & $80(1)$ & $75.6(2)$ & $58(2)$ & $\mathbf{80.3(2)}$ & $53.8(3)$ & $73.1(5)$ & $69.2(2)$ \\
        \hline
        \multirow{8}{*}{\rotatebox{90}{Llama-2-13b}} 
        & Zero Shot  & -- & $48.9$ & $77.6$ & $71.0$ & $79.4$ & $49.4$ & $80.3$ & $47.2$ & $72.1$ & $65.7$\\
        & LoRA (r=8) & $65.5$ & $\mathbf{57.6(5)}$ & $80.7(7)$ & $83.4(5)$ & $79.0(3)$ & $57(3)$ & $81.3(1)$ & $\mathbf{56(1)}$ & $\mathbf{78(1)}$ & $\mathbf{71.6(3)}$ \\
        & LoRA (r=32) & $262.1$ & $\mathbf{57.8(5)}$ & $80(1)$ & $\mathbf{84(1)}$ & $79.0(3)$ & $56(1)$ & $\mathbf{81.4(2)}$ & $\mathbf{57.0(4)}$ & $\mathbf{78.4(5)}$ & $\mathbf{71.9(2)}$ \\
        & VeRA (r=256) & $4.3$ & $53.2(5)$ & $79.8(9)$ & $80.4(1)$ & $77.73(7)$ & $57.7(2)$ & $80.80(3)$ & $49.7(2)$ & $74.55(8)$ & $69.25(4)$ \\    
        & \rev{LoTR (r=16)} & \rev{$1.84$} & \rev{$55.5(3)$} & \rev{$80.9(4)$} & \rev{$83.1(8)$} & \rev{$\mathbf{79.3(1)}$} & \rev{$55(3)$} & \rev{$81.0(2)$} & \rev{$52.7(3)$} & \rev{$74.5(3)$} & \rev{$70.2(4)$} \\
        & LoTR (r=64) & $9.83$ & $57(1)$ & $\mathbf{81.5(9)}$ & $\mathbf{83.6(4)}$ & $78.9(1)$ & $\mathbf{58(2)}$ & $\mathbf{81.5(1)}$ & $55.1(6)$ & $76(1)$ & $71.5(5)$ \\
        & MetaTT-4D (r=16) & $1.75$ & $55.2(2)$ & $80.8(5)$ & $81(2)$ & $\mathbf{79.20(6)}$ & $55(2)$ & $81.2(3)$ & $53.9(3)$ & $75.0(2)$ & $70.2(5)$  \\
        & MetaTT-4D (r=64) & $8.3$ & $56.9(2)$ & $\mathbf{81.9(5)}$ & $83(1)$ & $79.0(3)$ & $\mathbf{61(2)}$ & $80.8(4)$ & $54.11(5)$ & $75.0(4)$ & $71.5(3)$  \\
        \hline
    \end{tabular}
    \end{adjustbox}
    \caption{\textbf{Comparison of fine-tuning Llama-2-7b and Llama-2-13b.} \rev{We compare the performance of fine-tuning Llama-2-7b and Llama-2-13b on Commonsense170k \cite{hu2023llm}.} We show in bold the two best accuracies per task. \rev{All results are reported as mean $\pm$ standard error over 3 independent seeds; values in parentheses denote standard error on the last digit.} We observe that MetaTT-4D trails very closely to LoRA while often outperforming VeRA while using $\approx30$x and $\approx3$x less trainable parameters respectively. \rev{We also compare MetaTT-4D against LoTR at both $r\!=\!16$ and $r\!=\!64$: MetaTT-4D matches or outperforms LoTR at each rank while consistently using fewer parameters.}}
    \label{tab:commonsense}
    \vspace{-1em}
\end{table}
\paragraph{Commonsense reasoning.}
We present results in \cref{tab:commonsense} on the performance of MetaTT-4D against LoRA and other parameter-sharing methods introduced earlier in the context of commonsense reasoning: VeRA \cite{kopiczko2023vera} and LoTR \cite{bershatsky2024lotr}. We follow the same setup from \cite{hu2023llm} and first train on the Commonsense170k dataset, and assess results across eight different downstream tasks. For fine-tuning, we utilize Llama-2 models \cite{touvron2023llama} with 7B and 13B parameters as our pre-trained models. We report best accuracy results for each of the methods across two epochs (see \cref{app:exp-deets} for more details on selection of hyper-parameters). \rev{All methods are evaluated over 3 independent seeds and we report mean $\pm$ standard error.} Accuracies are evaluated using the lm-evaluation-harness framework \cite{eval-harness}. 

We observe that MetaTT-4D closely trails LoRA in terms of average performance across almost all the commonsense tasks for both Llama2-7b and Llama2-13b, while using significantly fewer parameters: up to $\approx30$x fewer parameters as compared to LoRA with less than $1\%$ drop in average accuracy. For both Llama2-7b and Llama2-13b, MetaTT-4D performs very similarly to LoTR while using fewer parameters and outperforms VeRA in almost all datasets while using $\approx3$x fewer trainable parameters. \rev{We additionally include LoTR at both $r\!=\!16$ and $r\!=\!64$ on each model for a direct comparison: MetaTT-4D matches or exceeds LoTR at matched ranks while consistently using fewer parameters (e.g., on Llama-2-7b at $r\!=\!64$, MetaTT-4D achieves avg $69.2(2)$ vs.\ LoTR at $68.6(4)$ with $16\%$ fewer parameters).}

\paragraph{Language understanding.} We compare fine-tuning RoBERTa based models with \rev{MetaTT-4D} against several baseline methods on GLUE Benchmark datasets - CoLA, MNLI, MRPC, QNLI, QQP, RTE, SST2 and STS-B \cite{wang2018glue} in \cref{tab:roberta-combined}. To isolate the performance of the shared adapters we only fine-tune the encoder adapter weights for the attention modules and not the classifier or regression heads for the corresponding downstream tasks. We defer the reader to \cref{app:exp-deets} for a detailed exposition on hyper-parameter tuning, adapter target modules, and the final set of hyper-parameters used to produce \cref{tab:roberta-combined}.

Our results indicate that \rev{MetaTT-4D} is competitive with other state-of-the-art methods. It outperforms LoRETTA across both models, and matches or outperforms VeRA (the margin on RoBERTa-large being within error). We also observe that at sufficient rank, LoTR, LoRTA, and \rev{MetaTT-4D} come close to LoRA in terms of accuracy across tasks, while using significantly fewer parameters. Finally, we reiterate that the accuracies reported in \cref{tab:roberta-combined} were achieved by only fine-tuning the encoder adapter weights of the attention modules (for LoRETTA, which additionally adapts the feed-forward output projection, the encoder adapters but not the classifier or regression heads), and not the final classifier or regression heads. We expect significant gains on top of these accuracies if the final classifier or regression heads are also trained. Note, the relative under-performance of VeRA is consistent with its restricted adaptation: VeRA learns only per-layer diagonal scalings over fixed, randomly-initialized, shared projection bases, whereas higher-order tensor decompositions learn structured factors shared across the layer and matrix-type modes. At a fixed trainable-parameter budget, the latter therefore expose a more expressive, task-adapted update subspace. Among the LoRETTA variants, we find that the adapter-based method consistently outperforms the version that reparameterizes the input model. Therefore, we chose to report only the results for the adapter-based approach.

\begin{table}[ht!]
\centering
\begin{adjustbox}{max width=\textwidth}
\begin{tabular}{|c|lcc*{9}{c}|}
\hline
\multirow{2}{*}{} & \multirow{2}{*}{\textbf{Method}} & \multirow{2}{*}{\shortstack{\textbf{Param}\\$\times 10^3$}} & \multirow{2}{*}{\textbf{Rank}} & \multicolumn{9}{c|}{\textbf{Metric (\%)}} \\
& & & &  \textbf{CoLA} &  \textbf{MNLI} &  \textbf{MRPC} &  \textbf{QNLI} &  \textbf{QQP} &  \textbf{RTE} &  \textbf{SST2} &  \textbf{STS-B} & \rev{\textbf{Avg}} \\
\hline
\multirow{12}{*}{\rotatebox{90}{{\rbase}}} 
& FT & $125\text{k}$ & -- & $61(1)$ & $87.6$ & $89.3(9)$ & $92.6(1)$ & $91.9$ & $79(2)$ & $94.1(1)$ & $90.4(2)$ & \rev{$85.7$} \\
\cline{2-13}
& LoRA & $295$ & $8$ & $\mathbf{61.1(6)}$ & $\mathbf{87.3(2)}$ & $88(1)$ & $91.3(2)$ & $\mathbf{90.1(1)}$ & $73(2)$ & $\mathbf{94.2(2)}$ & $90.7(2)$ & $\mathbf{84.5(3)}$ \\
\cline{2-13}
& VeRA & $43$ & $1024$ & $58(1)$ & $81(3)$ & $87.2(7)$ & $89.6(4)$ & $85.87(2)$ & $73.4(9)$ & $92.2(4)$ & $88.7(4)$ & \rev{$82.0(4)$}\\
\cline{2-13}
& $\text{LoRETTA}_{{adp}}$ & $57$ & $64,5$\textsuperscript{4} & $57.9(1)$ & $84.6(0)$ & $86.4(1)$ & $92.0(0)$ & $88.0(0)$ & $70.3(2)$ & $93.3(0)$ & --(--)\textsuperscript{3} & \rev{$81.8(0)^\dagger$} \\
\cline{2-13}
& \multirow{2}{*}{LoRTA} & $6.9$ & $8$ & $55.9(1)$  & $84.1(0)$ & $86.9(1)$ & $91.1(1)$ & $86.7(0)$ & $70.2(1)$ & $93.0(1)$ & $86.6(0)$ & \rev{$81.8(0)$} \\
& & $55$ & $64$ & $58.6(1)$ & $\mathbf{86.1(0)}$ & $88.0(2)$ & $\mathbf{92.2(0)}$ & $\mathbf{89.0(0)}$ & $75.0(2)$ & $93.6(0)$ & $89.3(0)$ & $\mathbf{84.0(0)}$ \\
\cline{2-13}
& \multirow{3}{*}{LoTR}& $100$ & $40$ & $58(2)$ & $85.2(2)$ & $88(1)$ & $\mathbf{92.5(3)}$ & $87.6(0)$ & $53(14)$ & $93.8(7)$ & $89.8(5)$ & \rev{$81(2)$} \\
& & $276$ & $80$ & $61(2)$ & $84.6(1)$ & $\mathbf{89.0(0)}$ & $92.1(5)$ & $86.8(0)$ & $71(3)$ & $93.4(1)$ & $\mathbf{90.9(2)}$ & \rev{$83.6(8)$} \\
& & $321$ & $88$ & $\mathbf{61.3(6)}$ & $84.7(0)$ & $88.0(9)$ & $92.0(4)$ & $86.9(0)$ & $67(13)$ & $93.3(2)$ & $\mathbf{91.0(1)}$ & \rev{$83(2)$} \\
\cline{2-13}
& \multirow{3}{*}{MetaTT-4D} & $13$ & $8$ & $58.8(5)$ & $84.2(1)$ & $87.6(2)$ & $90.4(1)$ & $86.9(1)$ & $72.9(5)$ & $92.0(1)$ & $89.1(2)$ & \rev{$82.7(1)$} \\
& & $45$ & $24$ & $59.7(7)$ & $85.5(1)$ & $88.6(4)$ & $91.0(1)$ & $87.8(1)$ & $74.2(4)$ & $92.3(2)$ & $89.9(2)$ & \rev{$83.6(1)$} \\
& & $156$ & $64$ & $61(1)$ & $85.9(1)$ & $\mathbf{88.9(3)}$ & $77(12)$\textsuperscript{1} & $88.5(1)$ & $\mathbf{77.5(7)}$ & $72(15)$\textsuperscript{1} & $90.1(2)$ & \rev{$80(2)^*$} \\
\hlinewd{1.5pt}
\multirow{8}{*}{\rotatebox{90}{\rlarge}}
& & & & & & & & & & & & \\
& FT & 355k & - & 68 & 90.2 & 91 & 94.7 & 92.2 & 87 & 96.4 & 92.4 & \rev{$89.0$} \\
\cline{2-13}
& LoRA & 786 & 8 & $\mathbf{68.0(7)}$\textsuperscript{2} & $\mathbf{90.6(2)}$ & $84(5)$\textsuperscript{2} & $\mathbf{94.8(3)}$ & $\mathbf{91.6(2)}$ & $\mathbf{87.0(8)}$\textsuperscript{2} & $\mathbf{95.7(2)}$ & $\mathbf{91.9(4)}$ & $\mathbf{88.0(6)}$ \\
\cline{2-13}
& VeRA & $61$ & $256$ & $64(2)$ & $88.8(2)$ & $89.4(4)$ & $93.1(2)$ & $87.62(8)$ & $83(1)$ & $95.1(2)$ & $91.5(1)$ & \rev{$86.6(3)$} \\
\cline{2-13}
& $\text{LoRETTA}_{adp}$ & $133$ & $64,5$\textsuperscript{4} & $61.0(1)$ & $89.69(0)$ & $88.1(1)$ & $94.08(1)$ & $89.6(0)$ & $72.0(4)$ & $95.5(0)$ & --(--)\textsuperscript{3} & \rev{$84.3(1)^\dagger$} \\
\cline{2-13}
& LoRTA & $9.1$ & $8$ & $58.9(1)$  &  $88.4(0)$ & $87.3(1)$ & $94.0(0)$ & $88.1(1)$ & $66.6(10)$ & $95.3(0)$ & $91.1(0)$ & \rev{$83.7(1)$} \\
\cline{2-13}
& LoTR & $328$ & $64$ & $61.3(9)$ & $\mathbf{90.3(0)}$ & $89.0(5)$ & $\mathbf{94.8(1)}$ & $\mathbf{89.2(1)}$ & $84(2)$ & $\mathbf{95.9(1)}$ & $91.6(1)$ & $\mathbf{87.0(3)}$ \\
\cline{2-13}
& \multirow{2}{*}{MetaTT-4D} & $39$ & $16$ & $62.8(5)$ & $89.6(1)$ & $88.6(3)$ & $93.8(1)$ & $88.5(1)$ & $84.2(5)$ & $95.2(2)$ & $91.8(1)$ & \rev{$86.8(1)$} \\
& & $92$ & $32$ & $64.0(1)$ & $90.0(1)$ & $\mathbf{90.1(3)}$ & $94.4(2)$ & $76(9)$\textsuperscript{1} & $\mathbf{84.8(6)}$ & $95.3(2)$ & $\mathbf{92.2(1)}$ & \rev{$86(1)^*$} \\
\hline
\end{tabular}
\end{adjustbox}

\par\vspace{3pt}
\begin{list}{}{%
  \setlength{\leftmargin}{1.6em}
  \setlength{\labelwidth}{1.6em}
  \setlength{\itemsep}{1pt}\setlength{\parsep}{0pt}\setlength{\topsep}{0pt}}
\footnotesize
\item[\textsuperscript{1}] For specifically these runs, we found that the model does not train for values of alpha less than $1$ (which were the ideal values found in the hyper-parameter search). We believe this is partly due to the initialization of tensor cores and partly because training TTs can be more challenging (more on this later in \cref{sec:dmrg}). We defer further exposition on this to \cref{app:exp-deets}.
\item[\textsuperscript{2}] In LoTR \cite{bershatsky2024lotr} these values of LoRA failed to train successfully. For better comparison we re-run LoRA for these datasets using the same random seeds as for MetaTT. We found that on one run in MRPC LoRA failed to train successfully as well.
\item[\textsuperscript{3}] We were unable to find the right set of hyper-parameters for STS-B when freezing the final regression layers.
\item[\textsuperscript{4}] In LoRETTA \cite{yang2024loretta}, the bottleneck size is set as $64$ for RoBERTa models and the TT rank is set as $5$ for the adapter based method.
\end{list}

\caption{\textbf{Comparison of MetaTT-4D against other PEFT techniques on \rbase and \rlarge}. Results for LoTR and LoRA are reported from \cite{bershatsky2024lotr}.  
For each dataset, we highlight the two best PEFT methods (FT is not considered for this ranking and we only list it as a benchmark). For CoLA, the metric is Matthew's correlation, for STS-B it is the Spearman's rank-correlation coefficient, and for all other datasets it is accuracy. Observe that variants of MetaTT sometimes outperform or match the performance of LoRA for a much lower parameter count (between $20$x and $2$x less parameters when compared to LoRA). Value in parenthesis is a standard error rounded up to the last single significant digit. \rev{Avg is the mean across all 8 tasks; $^\dagger$\,average over 7 tasks (STS-B excluded); $^*$\,average includes footnoted failure runs.} \rev{We also explored MetaTT-5D, which further decomposes the output dimension along heads; however, we found 4D to be more stable and parameter-efficient in practice (see \cref{tab:roberta-5d} in \cref{sec:5d-results}).}}
\label{tab:roberta-combined}
\end{table}

\subsection{Multi-Task Learning}\label{sec:mtl}
The modular architecture of TT based adapters allows for the inclusion of an extra tensor core capable of capturing task dependent features by simply assigning a low-rank rank-$3$ tensor along the TT chain. We explore this modification of the architecture by adding an extra core on MetaTT-4D placed at the middle of the TT, so that the ordering becomes $(D,L,T,M,D)$ where $T$ is the number of tasks on which the model is trained. 
We specifically choose this for symmetry of the tensor cores and not necessarily any particular reason.
We compare this adapter, henceforth named as MetaTT-(4+1)D, against four baselines -- a single LoRA adapter for all tasks and a MetaTT-4D adapter (which can be seen as a MetaTT-(4+1)D with the task dependent core frozen and set to identity), MTL-LoRA, and MoE-LoRA. \rev{Note that MetaTT-(4+1)D requires the task identity to be known at both training and inference time, since each sample's task label $t$ selects the corresponding slice $\mathcal{G}_T[t]$. This is a key difference from MoE-LoRA, which learns a routing gate and does not require explicit task labels at inference. However, in many practical MTL scenarios (e.g., serving multiple known downstream tasks), the task label is generally available.}

In the context of MTL, we distinguish between the following two approaches:
\begin{itemize}
    \item \textbf{Sequential Learning.} This approach involves first fine-tuning a model on a specific task, transferring the adapter to a new task for further fine-tuning, and then transferring the adapter back to the original task. The core idea is to leverage the features learned from the second task to enhance performance on the first task. However, a significant challenge with sequential learning is the risk of catastrophic forgetting or training interference, where the model may lose previously acquired knowledge or experience negative interactions between tasks, respectively. These issues have been extensively studied and documented in the literature (see e.g., \cite{zhang2024slca++}) and aligns with our observations.

    \item\textbf{Joint Training.} Alternatively, joint training aims to minimize a composite loss function that aggregates losses from multiple tasks at each epoch, i.e., the model is trained with the loss function $\mathcal{L}=\sum_{k=1}^{T} \mathcal{L}_k $, where $\mathcal{L}_k$ is the loss function for $k$\textsuperscript{th} task. 
\end{itemize}

\paragraph{Experimental Setup.} We fine-tune \rbase and \rlarge jointly on CoLA, MRPC, and RTE datasets from the GLUE benchmark datasets (MTL for 3 tasks). 
A notable issue in joint training is the disparity in dataset sizes, such as approximately $8000$ training samples in CoLA compared to around $3000$ in MRPC. To address this, we downsample each dataset to either the size of each dataset or a maximum of $5000$ samples per dataset, whichever is smaller. This forms the training set. For the evaluation set, we retain either $500$ samples or the full size of the validation/test set, whichever is smaller. For each trial, we first compute the mean performance across the three datasets at every epoch (yielding $20$ points per trial) and then select the best mean among those $20$ epochs. Finally, we report the average of these best means over 3 independent trials.

\paragraph{Empirical results and observations.} The results from our experiments on MTL for 3 tasks are shown in \cref{tab:MTL}.
We first observe that a single LoRA adapter can work remarkably well across different datasets and pre-trained models. This had already been documented in \cite[Table 1]{yang2025mtl}. For fine-tuning \rbase, we also observe that MoE-LoRA performs well across different datasets. \rev{On \rbase, MetaTT-(4+1)D achieves avg $70.5\%$ compared to LoRA at $74.9\%$ and MoE-LoRA at $77.2\%$, a $4$--$7$ point accuracy gap. In this case, the primary contribution of MetaTT-(4+1)D is thus an $\approx 22\times$ parameter reduction ($13.4$k vs.\ $\approx 300$k parameters), rather than competitive accuracy.} However, \rev{on \rlarge the picture is more favorable:} MetaTT-(4+1)D performs within $1\%$ of LoRA \rev{(avg $79.2\%$ vs.\ $80.0\%$)}, and outperforms both MoE-LoRA and MTL-LoRA, while requiring about $18.2$k trainable parameters (an $\approx43$x parameter reduction when compared to other baseline methods). Furthermore, we observe that in case of \rbase, MetaTT-4D performs very similar to MetaTT-(4+1) while using about $200$ less parameters, and in case of \rlarge, MetaTT-4D performs very similar to MoE-LoRA and MTL-LoRA while using about $200$ less parameters than MetaTT-(4+1)D (and using $\approx43$x less trainable parameters when compared to other methods). We defer further experiments and a discussion of how MetaTT-(4+1)D captures task dependent information for MTL to \cref{app:MTL}.

\begin{table}
\centering
\begin{adjustbox}{max width=\textwidth}
\begin{tabular}{|llcc*{4}{c}|}
\hline
\multirow{2}{*}{\textbf{Model}} & \multirow{2}{*}{\textbf{Method}} & \multirow{2}{*}{\shortstack{\textbf{Param}\\$\times 10^3$}} & \multirow{2}{*}{\textbf{Rank}} & \multicolumn{4}{c|}{\textbf{Metric} (\%)} \\
& & & &  {\textbf{CoLA}} &   {\textbf{MRPC}} &  {\textbf{RTE}} & {\textbf{Avg}} \\
\hline
         & LoRA & $295$ & $8$ & $\mathbf{60.7(8)}$ & $86.5(2)$ & $\mathbf{77.6(2)}$ & $\mathbf{74.9(2)}$ \\
         & MTL-LoRA \cite{yang2025mtl} & $296$ & $4$ & $53.0(1)$ & $\mathbf{87.8(1)}$ & $71.6(2)$ & $70.8(2)$ \\
         \rbase & MoE-LoRA \cite{liu2024moe} & $307$ & $8$ & $\mathbf{60.1(1)}$ & $\mathbf{88.6(1)}$ & $\mathbf{82.8(3)}$ & $\mathbf{77.2(2)}$ \\
         & MetaTT-4D & $13.2$ & $8$ & $53.2(2)$ & $85.9(4)$ & $72(2)$ & $70.3(8)$ \\
         & MetaTT-(4+1)D & $13.4$ & $8$ & $54(1)$  & $86.0(5)$ & $71.5(5)$ & $70.5(8)$ \\
         \hline
        & LoRA & $786$ & $8$ & $\mathbf{68(2)}$ & $\mathbf{89.3(6)}$ & $\mathbf{83.0(5)}$ & $\mathbf{80.0(3)}$ \\
        & MTL-LoRA \cite{yang2025mtl} & $789$ & $4$ & $58.8(1)$ & $88.7(2)$ & $82.1(2)$ & $76.5(2)$\\
        \rlarge & MoE-LoRA \cite{liu2024moe} & $811$ & $8$ & $61.1(0)$ & $\mathbf{89.9(1)}$ & $82.6(1)$ & $77.8(1)$ \\
        & MetaTT-4D & $18.0$ & $8$ & $59.5(5)$ & $88.4(5)$ & $81.1(8)$ & $76.3(6)$ \\
         & MetaTT-(4+1)D & $18.2$ & $8$ & $\mathbf{64.0(8)}$ & $89.0(6)$ & $\mathbf{84.4(4)}$ & $\mathbf{79.2(4)}$\\
        \hline
    \end{tabular}
    \end{adjustbox}
    \caption{\textbf{Results of MTL with 3 tasks.} We observe that across both \rbase and \rlarge, MetaTT-(4+1)D outperforms single MetaTT-4D adapters for almost all of the datasets, while using about $200$ more trainable parameters. For \rbase MetaTT-(4+1)D performs comparably to MTL-LoRA while using $\approx22$x less parameters, and for \rlarge MetaTT-(4+1)D outperforms both MTL-LoRA and MoE-LoRA on average, and is within $1\%$ of average accuracy of LoRA, while using $\approx43$x less parameters. We show in bold the two best accuracies per task.
    }
    \label{tab:MTL}
    \vspace{-1em}
\end{table}

\subsection{Rank Adaptive Fine-tuning via DMRG-inspired Sweep}\label{sec:dmrg} 
\begin{figure}
    \centering
    \includegraphics[width=\textwidth]{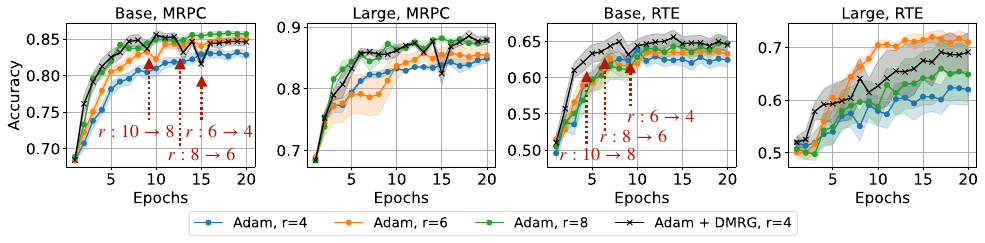}
    \vspace{-1em}
    \caption{\textbf{Comparison of AdamW and AdamW+\cref{alg:DMRG} sweeps applied at certain epochs.} Results are shown for MetaTT-5D on MRPC and RTE for \rbase and \rlarge. In Adam we fix the rank throughout.  
    For AdamW+\cref{alg:DMRG} we start with a $r=10$ TT and progressively decrease ranks until we reach $r=4$ as indicated by arrows on the plots for the base model, with the same schedule followed by the large counterparts. Error bars in both panels correspond to standard errors. The learning rate used across all the optimizers is $5e-4$ with $0$ weight decay.}
    \label{fig:dmrg}
\end{figure}

\paragraph{Empirical evaluations.} We present comparisons of training \rbase and \rlarge using AdamW \cite{loshchilov2017decoupled} and interdispersing DMRG-inspired sweeps as in \cref{alg:DMRG} on the MRPC and RTE dataset in \cref{fig:dmrg}. For AdamW \cite{loshchilov2017decoupled} we fine-tune on fixed ranks $\{4,6,8\}$ for a given learning rate. We observe that one can adaptively change the ranks in the training phase without any major performance degradation. For \rbase, we show that using AdamW together with DMRG-inspired sweeps we achieve higher accuracy
at rank $r=4$ when compared to the accuracy achieved by AdamW and $r=4$. We also show that the performance improvement with \cref{alg:DMRG} when fine-tuning \rlarge is even more significant (see \cref{tab:dmrg-tab}). For both models, one can observe that the accuracy reduces significantly when truncated SVD is applied, followed by a rapid climb, with deeper gorges as we go to smaller ranks. Each DMRG sweep is applied right after each training epoch, before the evaluation on the validation dataset. Thus, it removes a significant amount of information, across all bonds of the TT, and so performance degradation is expected, before AdamW is able to readjust to its new weight space at the next epoch. This problem is exacerbated when DMRG is applied to smaller ranks, as the relative change in ranks (current rank divided by target rank) substantially increase as we go from higher ranks to lower ranks. We defer further discussions on the experiments with DMRG to the \cref{app:DMRG} and \cref{sec:ada_vs_dmrg_sec}. \rev{We additionally demonstrate in \cref{sec:adalora_vs_dmrg} that DMRG sweeps are effective at the Llama-2-7b scale on a downsampled dataset (Commonsense15k). Scaling DMRG to larger datasets (e.g., the full Commonsense170k) remains an important direction for future work.}
\begin{table}[h]
    \centering
    \begin{adjustbox}{max width=\textwidth}
    \begin{tabular}{|lcc|}
        \hline
        \textbf{Model} & \textbf{AdamW} & \textbf{AdamW + DMRG} \\
        \hline
        Base, MRPC & 0.839 & 0.852 \\
        Large, MRPC & 0.854 & 0.887 \\
        Base, RTE  & 0.652 & 0.658 \\
        Large, RTE & 0.640 & 0.701 \\
        \hline
    \end{tabular}
    \end{adjustbox}
    \caption{\textbf{AdamW vs. AdamW+DMRG.} Comparison of the average of per-trial maximum accuracies (computed over 20 epochs), for 10 trials (\rbase) and 4 trials (\rlarge), between AdamW and AdamW + DMRG optimizers at target rank $r=4$.}
    \label{tab:dmrg-tab}
\end{table}

\section{Conclusions}

In this work, we have introduced MetaTT, a novel approach to parameter-efficient fine-tuning of large language models using TT decompositions. By leveraging the TT architecture, MetaTT achieves significant reductions in the number of trainable parameters while maintaining competitive performance compared to state-of-the-art methods. Our empirical evaluations demonstrate that MetaTT can achieve significant parameter reduction with similar accuracy on standard language modeling benchmarks when compared to these methods. 

The TT representation provides a compact and globally shared core, allowing for efficient parameter sharing across all components of a transformer network. Unlike methods that compress each weight matrix in isolation, MetaTT factorizes all linear sub-modules into a single shared TT, capturing structural axes such as layer, matrix-type, and optionally heads and tasks. This global compression leads to higher compression rates and improved scalability, making MetaTT a promising solution for fine-tuning large models. However, for single task learning, we observe that MetaTT often performs similar to other tensor based decompositions (including other variants of the TT decomposition).

To differentiate beyond single task fine-tuning, we observe that tensor based adapters can be easily extended to perform joint-MTL. Notably, we demonstrate that extending a simple modification to the architecture for single task learning, one can use MetaTT for joint-MTL. This is because the modular architecture of MetaTT enables extension to shared adapters across multiple tasks or expert partitions, without the need to redesign the core tensor. This had remained unexplored prior to our work. We further hypothesize that similar extensions could be applied to other tensor-based architectures as well. Furthermore, the TT structure benefits from mature optimization routines, such as DMRG-style alternating minimization, which simplifies rank tuning. \rev{We have demonstrated DMRG-inspired rank adaptation on RoBERTa (MRPC, RTE) and on Llama-2-7b with Commonsense15k, where it yields clear improvements over fixed-rank training. Scaling this approach to larger datasets remains an open direction; the theoretical cost advantage of SVD sweeps on TT bonds (vs.\ per-layer SVDs in LoRA-based methods) is model-size-independent and becomes more favorable as models grow (see \cref{sec:compl_ada_vs_dmrg}).}

Our results suggest that assuming low-rankness in the manifold of shared parameters is a viable strategy for parameter-efficient fine-tuning. The TT decomposition captures this manifold effectively, providing a robust framework for reducing computational overhead while preserving model performance. Future work may explore other tensor networks that better capture parameter sharing, including quantum-circuit inspired tensor network that may lift the low-rankness description while maintaining efficient parameter count. 

While our focus here has been on fine-tuning, we anticipate MetaTT to find extensions to other contexts including the design of new foundation models with shared parameters and for model compression. Moreover, DMRG-inspired techniques can offer a principled way to compress TTs during training. Finding applications where compression during training phase or alternative DMRG-inspired techniques extending beyond those discussed in this work, presents an exciting avenue.

\subsubsection*{Reproducibility Statement}

We provide all the codes and pseudocodes required for verifying experiments with MetaTT in \cref{app:metatt-codes}. Furthermore, the grids for hyper-parameter search and the final set of hyper-parameters required to reproduce the results of \cref{tab:commonsense}, \cref{tab:roberta-combined}, and \cref{tab:MTL} are reported in \cref{app:exp-deets} and \cref{app:MTL}. The experimental details for DMRG based experiments are reported in \cref{sec:dmrg}.

\subsubsection*{Disclaimer}

This paper was prepared for informational purposes by the Global Technology Applied Research center of JPMorgan Chase \& Co. This paper is not a merchandisable/sellable product of the Research Department of JPMorgan Chase \& Co. or its affiliates. Neither JPMorgan Chase \& Co. nor any of its affiliates makes any explicit or implied representation or warranty and none of them accept any liability in connection with this paper, including, without limitation, with respect to the completeness, accuracy, or reliability of the information contained herein and the potential legal, compliance, tax, or accounting effects thereof. This document is not intended as investment research or investment advice, or as a recommendation, offer, or solicitation for the purchase or sale of any security, financial instrument, financial product or service, or to be used in any way for evaluating the merits of participating in any transaction.

\bibliography{refs}

@article{meng2024pissa,
  title={Pissa: Principal singular values and singular vectors adaptation of large language models},
  author={Meng, Fanxu and Wang, Zhaohui and Zhang, Muhan},
  journal={Advances in Neural Information Processing Systems},
  volume={37},
  pages={121038--121072},
  year={2024}
}

@article{buyukakyuz2024olora,
  title={Olora: Orthonormal low-rank adaptation of large language models},
  author={B{\"u}y{\"u}kaky{\"u}z, Kerim},
  journal={arXiv preprint arXiv:2406.01775},
  year={2024}
}

@article{karimi2021parameter,
  title={Parameter-efficient multi-task fine-tuning for transformers via shared hypernetworks},
  author={Karimi Mahabadi, Rabeeh  and Ruder, Sebastian and Dehghani, Mostafa and Henderson, James},
  journal={arXiv preprint arXiv:2106.04489},
  year={2021}
}

@book{krishnamoorthy2006handbook,
  title={Handbook of statistical distributions with applications},
  author={Krishnamoorthy, Kalimuthu},
  year={2006},
  publisher={Chapman and Hall/CRC}
}

@article{karimi2021compacter,
  title={Compacter: Efficient low-rank hypercomplex adapter layers},
  author={Karimi Mahabadi, Rabeeh and Henderson, James and Ruder, Sebastian},
  journal={Advances in neural information processing systems},
  volume={34},
  pages={1022--1035},
  year={2021}
}

@article{wang2023multilora,
  title={Multilora: Democratizing lora for better multi-task learning},
  author={Wang, Yiming and Lin, Yu and Zeng, Xiaodong and Zhang, Guannan},
  journal={arXiv preprint arXiv:2311.11501},
  year={2023}
}

@inproceedings{wolf-etal-2020-transformers,
    title = "Transformers: State-of-the-Art Natural Language Processing",
    author = "Thomas Wolf and Lysandre Debut and Victor Sanh and Julien Chaumond and Clement Delangue and Anthony Moi and Pierric Cistac and Tim Rault and Rémi Louf and Morgan Funtowicz and Joe Davison and Sam Shleifer and Patrick von Platen and Clara Ma and Yacine Jernite and Julien Plu and Canwen Xu and Teven Le Scao and Sylvain Gugger and Mariama Drame and Quentin Lhoest and Alexander M. Rush",
    booktitle = "Proceedings of the 2020 Conference on Empirical Methods in Natural Language Processing: System Demonstrations",
    month = oct,
    year = "2020",
    address = "Online",
    publisher = "Association for Computational Linguistics",
    url = "https://www.aclweb.org/anthology/2020.emnlp-demos.6",
}

@inproceedings{liu2024moe,
  title={When {MoE} meets {LLM}s: Parameter efficient fine-tuning for multi-task medical applications},
  author={Liu, Qidong and Wu, Xian and Zhao, Xiangyu and Zhu, Yuanshao and Xu, Derong and Tian, Feng and Zheng, Yefeng},
  booktitle={Proceedings of the 47th International ACM SIGIR Conference on Research and Development in Information Retrieval},
  year={2024}
}

@article{li2025uni,
  title={Uni-LoRA: One Vector is All You Need},
  author={Li, Kaiyang and Han, Shaobo and Su, Qing and Li, Wei and Cai, Zhipeng and Ji, Shihao},
  journal={arXiv preprint arXiv:2506.00799},
  year={2025}
}

@inproceedings{yin2022batude,
  title={Batude: Budget-aware neural network compression based on tucker decomposition},
  author={Yin, Miao and Phan, Huy and Zang, Xiao and Liao, Siyu and Yuan, Bo},
  booktitle={Proceedings of the AAAI Conference on Artificial Intelligence},
  year={2022}
}

@article{kim2015compression,
  title={Compression of deep convolutional neural networks for fast and low power mobile applications},
  author={Kim, Yong-Deok and Park, Eunhyeok and Yoo, Sungjoo and Choi, Taelim and Yang, Lu and Shin, Dongjun},
  journal={arXiv preprint arXiv:1511.06530},
  year={2015}
}

@article{li2024vb,
  title={Vb-lora: Extreme parameter efficient fine-tuning with vector banks},
  author={Li, Yang and Han, Shaobo and Ji, Shihao},
  journal={Advances in Neural Information Processing Systems},
  year={2024}
}

@inproceedings{yang2025mtl,
  title={Mtl-lora: Low-rank adaptation for multi-task learning},
  author={Yang, Yaming and Muhtar, Dilxat and Shen, Yelong and Zhan, Yuefeng and Liu, Jianfeng and Wang, Yujing and Sun, Hao and Deng, Weiwei and Sun, Feng and Zhang, Qi and others},
  booktitle={Proceedings of the AAAI Conference on Artificial Intelligence},
  year={2025}
}

@inproceedings{lhoest21,
    title = "Datasets: A Community Library for Natural Language Processing",
    author = "Lhoest, Quentin  and
      Villanova del Moral, Albert  and
      Jernite, Yacine  and
      Thakur, Abhishek  and
      von Platen, Patrick  and
      Patil, Suraj  and
      Chaumond, Julien  and
      Drame, Mariama  and
      Plu, Julien  and
      Tunstall, Lewis  and
      Davison, Joe  and
      {\v{S}}a{\v{s}}ko, Mario  and
      Chhablani, Gunjan  and
      Malik, Bhavitvya  and
      Brandeis, Simon  and
      Le Scao, Teven  and
      Sanh, Victor  and
      Xu, Canwen  and
      Patry, Nicolas  and
      McMillan-Major, Angelina  and
      Schmid, Philipp  and
      Gugger, Sylvain  and
      Delangue, Cl{\'e}ment  and
      Matussi{\`e}re, Th{\'e}o  and
      Debut, Lysandre  and
      Bekman, Stas  and
      Cistac, Pierric  and
      Goehringer, Thibault  and
      Mustar, Victor  and
      Lagunas, Fran{\c{c}}ois  and
      Rush, Alexander  and
      Wolf, Thomas",
    booktitle = "Proceedings of the 2021 Conference on Empirical Methods in Natural Language Processing: System Demonstrations",
    month = {nov},
    year = "2021",
    address = "Online and Punta Cana, Dominican Republic",
    publisher = "Association for Computational Linguistics",
    url = "https://aclanthology.org/2021.emnlp-demo.21",
    pages = "175--184",
    eprint={2109.02846},
    archivePrefix={arXiv},
    primaryClass={cs.CL},
}

@article{loshchilov2017decoupled,
  title={Decoupled weight decay regularization},
  author={Loshchilov, Ilya and Hutter, Frank},
  journal={arXiv preprint arXiv:1711.05101},
  year={2017}
}

@article{musco2017recursive,
  title={Recursive sampling for the nystrom method},
  author={Musco, Cameron and Musco, Christopher},
  journal={Advances in neural information processing systems},
  year={2017}
}

@inproceedings{cohen2017input,
  title={Input sparsity time low-rank approximation via ridge leverage score sampling},
  author={Cohen, Michael B and Musco, Cameron and Musco, Christopher},
  booktitle={Proceedings of the Twenty-Eighth Annual ACM-SIAM Symposium on Discrete Algorithms},
  year={2017}
}

@article{musco2015randomized,
  title={Randomized block krylov methods for stronger and faster approximate singular value decomposition},
  author={Musco, Cameron and Musco, Christopher},
  journal={Advances in neural information processing systems},
  year={2015}
}

@article{koike2025quantum,
  title={Quantum-PEFT: Ultra parameter-efficient fine-tuning},
  author={Koike-Akino, Toshiaki and Tonin, Francesco and Wu, Yongtao and Wu, Frank Zhengqing and Candogan, Leyla Naz and Cevher, Volkan},
  journal={arXiv preprint arXiv:2503.05431},
  year={2025}
}

@article{zhang2024riemannian,
  title={Riemannian preconditioned lora for fine-tuning foundation models},
  author={Zhang, Fangzhao and Pilanci, Mert},
  journal={arXiv preprint arXiv:2402.02347},
  year={2024}
}

@article{zi2023delta,
  title={Delta-lora: Fine-tuning high-rank parameters with the delta of low-rank matrices},
  author={Zi, Bojia and Qi, Xianbiao and Wang, Lingzhi and Wang, Jianan and Wong, Kam-Fai and Zhang, Lei},
  journal={arXiv preprint arXiv:2309.02411},
  year={2023}
}

@article{zhang2023adalora,
  title={Adalora: Adaptive budget allocation for parameter-efficient fine-tuning},
  author={Zhang, Qingru and Chen, Minshuo and Bukharin, Alexander and Karampatziakis, Nikos and He, Pengcheng and Cheng, Yu and Chen, Weizhu and Zhao, Tuo},
  journal={arXiv preprint arXiv:2303.10512},
  year={2023}
}

@article{aghajanyan2020intrinsic,
  title={Intrinsic dimensionality explains the effectiveness of language model fine-tuning},
  author={Aghajanyan, Armen and Zettlemoyer, Luke and Gupta, Sonal},
  journal={arXiv preprint arXiv:2012.13255},
  year={2020}
}

@misc{jiang2023mistral7b,
      title={Mistral 7B}, 
      author={Albert Q. Jiang and Alexandre Sablayrolles and Arthur Mensch and Chris Bamford and Devendra Singh Chaplot and Diego de las Casas and Florian Bressand and Gianna Lengyel and Guillaume Lample and Lucile Saulnier and Lélio Renard Lavaud and Marie-Anne Lachaux and Pierre Stock and Teven Le Scao and Thibaut Lavril and Thomas Wang and Timothée Lacroix and William El Sayed},
      year={2023},
      eprint={2310.06825},
      archivePrefix={arXiv},
      primaryClass={cs.CL},
      url={https://arxiv.org/abs/2310.06825}, 
}

@article{almazrouei2023falcon,
  title={The falcon series of open language models},
  author={Almazrouei, Ebtesam and Alobeidli, Hamza and Alshamsi, Abdulaziz and Cappelli, Alessandro and Cojocaru, Ruxandra and Debbah, M{\'e}rouane and Goffinet, {\'E}tienne and Hesslow, Daniel and Launay, Julien and Malartic, Quentin and others},
  journal={arXiv preprint arXiv:2311.16867},
  year={2023}
}

@article{hurst2024gpt,
  title={Gpt-4o system card},
  author={Hurst, Aaron and Lerer, Adam and Goucher, Adam P and Perelman, Adam and Ramesh, Aditya and Clark, Aidan and Ostrow, AJ and Welihinda, Akila and Hayes, Alan and Radford, Alec and others},
  journal={arXiv preprint arXiv:2410.21276},
  year={2024}
}

@article{team2024gemini,
  title={Gemini 1.5: Unlocking multimodal understanding across millions of tokens of context},
  author={Team, Gemini and Georgiev, Petko and Lei, Ving Ian and Burnell, Ryan and Bai, Libin and Gulati, Anmol and Tanzer, Garrett and Vincent, Damien and Pan, Zhufeng and Wang, Shibo and others},
  journal={arXiv preprint arXiv:2403.05530},
  year={2024}
}

@article{grattafiori2024llama,
  title={The Llama 3 herd of models},
  author={Grattafiori, Aaron and Dubey, Abhimanyu and Jauhri, Abhinav and Pandey, Abhinav and Kadian, Abhishek and Al-Dahle, Ahmad and Letman, Aiesha and Mathur, Akhil and Schelten, Alan and Vaughan, Alex and others},
  journal={arXiv preprint arXiv:2407.21783},
  year={2024}
}

@article{brown2020language,
  title={Language models are few-shot learners},
  author={Brown, Tom and Mann, Benjamin and Ryder, Nick and Subbiah, Melanie and Kaplan, Jared D and Dhariwal, Prafulla and Neelakantan, Arvind and Shyam, Pranav and Sastry, Girish and Askell, Amanda and others},
  journal={Advances in neural information processing systems},
  year={2020}
}

@inproceedings{runwal2025peft,
  title={From peft to deft: Parameter efficient finetuning for reducing activation density in transformers},
  author={Runwal, Bharat and Pedapati, Tejaswini and Chen, Pin-Yu},
  booktitle={Proceedings of the AAAI Conference on Artificial Intelligence},
  year={2025}
}

@article{albert2025randlora,
  title={RandLoRA: Full-rank parameter-efficient fine-tuning of large models},
  author={Albert, Paul and Zhang, Frederic Z and Saratchandran, Hemanth and Rodriguez-Opazo, Cristian and Hengel, Anton van den and Abbasnejad, Ehsan},
  journal={arXiv preprint arXiv:2502.00987},
  year={2025}
}

@inproceedings{zanella2024low,
  title={Low-rank few-shot adaptation of vision-language models},
  author={Zanella, Maxime and Ben Ayed, Ismail},
  booktitle={Proceedings of the IEEE/CVF Conference on Computer Vision and Pattern Recognition},
  year={2024}
}

@article{wistuba2024choice,
  title={Choice of peft technique in continual learning: Prompt tuning is not all you need},
  author={Wistuba, Martin and Sivaprasad, Prabhu Teja and Balles, Lukas and Zappella, Giovanni},
  journal={arXiv preprint arXiv:2406.03216},
  year={2024}
}

@article{han2024parameter,
  title={Parameter-efficient fine-tuning for large models: A comprehensive survey},
  author={Han, Zeyu and Gao, Chao and Liu, Jinyang and Zhang, Jeff and Zhang, Sai Qian},
  journal={arXiv preprint arXiv:2403.14608},
  year={2024}
}

@inproceedings{zhou2022conditional,
  title={Conditional prompt learning for vision-language models},
  author={Zhou, Kaiyang and Yang, Jingkang and Loy, Chen Change and Liu, Ziwei},
  booktitle={Proceedings of the IEEE/CVF conference on computer vision and pattern recognition},
  year={2022}
}

@article{zhou2022learning,
  title={Learning to prompt for vision-language models},
  author={Zhou, Kaiyang and Yang, Jingkang and Loy, Chen Change and Liu, Ziwei},
  journal={International Journal of Computer Vision},
  year={2022}
}

@inproceedings{khattak2023self,
  title={Self-regulating prompts: Foundational model adaptation without forgetting},
  author={Khattak, Muhammad Uzair and Wasim, Syed Talal and Naseer, Muzammal and Khan, Salman and Yang, Ming-Hsuan and Khan, Fahad Shahbaz},
  booktitle={Proceedings of the IEEE/CVF international conference on computer vision},
  year={2023}
}

@inproceedings{zhang2024dept,
  title={{DePT}: Decoupled prompt tuning},
  author={Zhang, Ji and Wu, Shihan and Gao, Lianli and Shen, Heng Tao and Song, Jingkuan},
  booktitle={Proceedings of the IEEE/CVF Conference on Computer Vision and Pattern Recognition},
  year={2024}
}

@article{wang2018glue,
  title={{GLUE}: A multi-task benchmark and analysis platform for natural language understanding},
  author={Wang, Alex and Singh, Amanpreet and Michael, Julian and Hill, Felix and Levy, Omer and Bowman, Samuel R},
  journal={arXiv preprint arXiv:1804.07461},
  year={2018}
}

@article{monturiol2025tensorization,
  title={Tensorization of neural networks for improved privacy and interpretability},
  author={Monturiol, Jos{\'e} Ram{\'o}n Pareja and Pozas-Kerstjens, Alejandro and P{\'e}rez-Garc{\'\i}a, David},
  journal={arXiv preprint arXiv:2501.06300},
  year={2025}
}

@article{koohpayegani2023nola,
  title={{NOLA}: Compressing {LoRA} using linear combination of random basis},
  author={Koohpayegani, Soroush Abbasi and Navaneet, KL and Nooralinejad, Parsa and Kolouri, Soheil and Pirsiavash, Hamed},
  journal={arXiv preprint arXiv:2310.02556},
  year={2023}
}

@inproceedings{kopiczko2023vera,
  title={Ve{RA}: Vector-based Random Matrix Adaptation},
  author={Dawid Jan Kopiczko and Tijmen Blankevoort and Yuki M Asano},
  booktitle={The Twelfth International Conference on Learning Representations},
  year={2024},
  url={https://openreview.net/forum?id=NjNfLdxr3A}
}

@article{khrulkov2017expressive,
  title={Expressive power of recurrent neural networks},
  author={Khrulkov, Valentin and Novikov, Alexander and Oseledets, Ivan},
  journal={arXiv preprint arXiv:1711.00811},
  year={2017}
}

@misc{eval-harness,
  author       = {Gao, Leo and Tow, Jonathan and Abbasi, Baber and Biderman, Stella and Black, Sid and DiPofi, Anthony and Foster, Charles and Golding, Laurence and Hsu, Jeffrey and Le Noac'h, Alain and Li, Haonan and McDonell, Kyle and Muennighoff, Niklas and Ociepa, Chris and Phang, Jason and Reynolds, Laria and Schoelkopf, Hailey and Skowron, Aviya and Sutawika, Lintang and Tang, Eric and Thite, Anish and Wang, Ben and Wang, Kevin and Zou, Andy},
  title        = {The Language Model Evaluation Harness},
  month        = 07,
  year         = 2024,
  publisher    = {Zenodo},
  version      = {v0.4.3},
  doi          = {10.5281/zenodo.12608602},
  url          = {https://zenodo.org/records/12608602}
}

@article{touvron2023llama,
  title={Llama 2: Open foundation and fine-tuned chat models},
  author={Touvron, Hugo and Martin, Louis and Stone, Kevin and Albert, Peter and Almahairi, Amjad and Babaei, Yasmine and Bashlykov, Nikolay and Batra, Soumya and Bhargava, Prajjwal and Bhosale, Shruti and others},
  journal={arXiv preprint arXiv:2307.09288},
  year={2023}
}

@article{tropp2023randomized,
  title={Randomized algorithms for low-rank matrix approximation: Design, analysis, and applications},
  author={Tropp, Joel A and Webber, Robert J},
  journal={arXiv preprint arXiv:2306.12418},
  year={2023}
}

@article{halko2011finding,
  title={Finding structure with randomness: Probabilistic algorithms for constructing approximate matrix decompositions},
  author={Halko, Nathan and Martinsson, Per-Gunnar and Tropp, Joel A},
  journal={SIAM review},
  year={2011}
}

@article{hu2023llm,
  title={Llm-adapters: An adapter family for parameter-efficient fine-tuning of large language models},
  author={Hu, Zhiqiang and Wang, Lei and Lan, Yihuai and Xu, Wanyu and Lim, Ee-Peng and Bing, Lidong and Xu, Xing and Poria, Soujanya and Lee, Roy Ka-Wei},
  journal={arXiv preprint arXiv:2304.01933},
  year={2023}
}

@inproceedings{devlin2019bert,
  title={{BERT}: Pre-training of deep bidirectional transformers for language understanding},
  author={Devlin, Jacob and Chang, Ming-Wei and Lee, Kenton and Toutanova, Kristina},
  booktitle={Proceedings of the 2019 conference of the North American chapter of the association for computational linguistics: human language technologies, volume 1 (long and short papers)},
  year={2019}
}

@article{hu2021lora,
  title={{LoRA}: Low-rank adaptation of large language models},
  author={Hu, Edward J and Shen, Yelong and Wallis, Phillip and Allen-Zhu, Zeyuan and Li, Yuanzhi and Wang, Shean and Wang, Lu and Chen, Weizhu},
  journal={arXiv preprint arXiv:2106.09685},
  year={2021}
}

@article{yang2024loretta,
  title={{LoRETTA}: Low-Rank Economic Tensor-Train Adaptation for Ultra-Low-Parameter Fine-Tuning of Large Language Models},
  author={Yang, Yifan and Zhou, Jiajun and Wong, Ngai and Zhang, Zheng},
  journal={arXiv preprint arXiv:2402.11417},
  year={2024}
}

@article{anjum2024tensor,
  title={Tensor Train Low-rank Approximation ({TT-LoRA}): Democratizing AI with Accelerated LLMs},
  author={Anjum, Afia and Eren, Maksim E and Boureima, Ismael and Alexandrov, Boian and Bhattarai, Manish},
  journal={arXiv preprint arXiv:2408.01008},
  year={2024}
}

@article{chen2024quanta,
  title={{QuanTA}: Efficient High-Rank Fine-Tuning of LLMs with Quantum-Informed Tensor Adaptation},
  author={Chen, Zhuo and Dangovski, Rumen and Loh, Charlotte and Dugan, Owen and Luo, Di and Solja{\v{c}}i{\'c}, Marin},
  journal={arXiv preprint arXiv:2406.00132},
  year={2024}
}

@article{bershatsky2024lotr,
  title={{LoTR}: Low tensor rank weight adaptation},
  author={Bershatsky, Daniel and Cherniuk, Daria and Daulbaev, Talgat and Mikhalev, Aleksandr and Oseledets, Ivan},
  journal={arXiv preprint arXiv:2402.01376},
  year={2024}
}

@inproceedings{jie2023fact,
  title={{FacT}: Factor-tuning for lightweight adaptation on vision transformer},
  author={Jie, Shibo and Deng, Zhi-Hong},
  booktitle={Proceedings of the AAAI conference on artificial intelligence},
  year={2023}
}

@article{hounie2024lorta,
  title={LoRTA: Low Rank Tensor Adaptation of Large Language Models},
  author={Hounie, Ignacio and Kanatsoulis, Charilaos and Tandon, Arnuv and Ribeiro, Alejandro},
  journal={arXiv preprint arXiv:2410.04060},
  year={2024}
}

@article{barratt2022improvements,
  title={Improvements to gradient descent methods for quantum tensor network machine learning},
  author={Barratt, Fergus and Dborin, James and Wright, Lewis},
  journal={arXiv preprint arXiv:2203.03366},
  year={2022}
}

@article{stoudenmire2016supervised,
  title={Supervised learning with tensor networks},
  author={Stoudenmire, Edwin and Schwab, David J},
  journal={Advances in neural information processing systems},
  volume={29},
  year={2016}
}

@article{novikov2016exponential,
  title={Exponential machines},
  author={Novikov, Alexander and Trofimov, Mikhail and Oseledets, Ivan},
  journal={arXiv preprint arXiv:1605.03795},
  year={2016}
}

@article{kingma2014adam,
  title={Adam: A method for stochastic optimization},
  author={Kingma, Diederik P and Ba, Jimmy},
  journal={arXiv preprint arXiv:1412.6980},
  year={2014}
}

@article{schollwock2011density,
  title={The density-matrix renormalization group in the age of matrix product states},
  author={Schollw{\"o}ck, Ulrich},
  journal={Annals of physics},
  year={2011}
}

@article{verstraete2023density,
  title={Density matrix renormalization group, 30 years on},
  author={Verstraete, Frank and Nishino, Tomotoshi and Schollw{\"o}ck, Ulrich and Ba{\~n}uls, Mari Carmen and Chan, Garnet K and Stoudenmire, Miles E},
  journal={Nature Reviews Physics},
  year={2023}
}

@article{kolda2009tensor,
  title={Tensor decompositions and applications},
  author={Kolda, Tamara G and Bader, Brett W},
  journal={SIAM review},
  year={2009}
}

@article{novikov2015tensorizing,
  title={Tensorizing neural networks},
  author={Novikov, Alexander and Podoprikhin, Dmitrii and Osokin, Anton and Vetrov, Dmitry P},
  journal={Advances in neural information processing systems},
  year={2015}
}

@article{garipov2016ultimate,
  title={Ultimate tensorization: compressing convolutional and fc layers alike},
  author={Garipov, Timur and Podoprikhin, Dmitry and Novikov, Alexander and Vetrov, Dmitry},
  journal={arXiv preprint arXiv:1611.03214},
  year={2016}
}

@article{lu2025fetta,
  title={FETTA: Flexible and Efficient Hardware Accelerator for Tensorized Neural Network Training},
  author={Lu, Jinming and Tian, Jiayi and Li, Hai and Young, Ian and Zhang, Zheng},
  journal={arXiv preprint arXiv:2504.06474},
  year={2025}
}

@article{paszke2019pytorch,
  title={Pytorch: An imperative style, high-performance deep learning library},
  author={Paszke, Adam and Gross, Sam and Massa, Francisco and Lerer, Adam and Bradbury, James and Chanan, Gregory and Killeen, Trevor and Lin, Zeming and Gimelshein, Natalia and Antiga, Luca and others},
  journal={Advances in neural information processing systems},
  volume={32},
  year={2019}
}

@Misc{peft,
  title =        {PEFT: State-of-the-art Parameter-Efficient Fine-Tuning methods},
  author =       {Sourab Mangrulkar and Sylvain Gugger and Lysandre Debut and Younes Belkada and Sayak Paul and Benjamin Bossan},
  howpublished = {\url{https://github.com/huggingface/peft}},
  year =         {2022}
}

@article{wolf2019huggingface,
  title={Huggingface's transformers: State-of-the-art natural language processing},
  author={Wolf, Thomas and Debut, Lysandre and Sanh, Victor and Chaumond, Julien and Delangue, Clement and Moi, Anthony and Cistac, Pierric and Rault, Tim and Louf, R{\'e}mi and Funtowicz, Morgan and others},
  journal={arXiv preprint arXiv:1910.03771},
  year={2019}
}

@article{zhang2024slca++,
  title={SLCA++: Unleash the Power of Sequential Fine-tuning for Continual Learning with Pre-training},
  author={Zhang, Gengwei and Wang, Liyuan and Kang, Guoliang and Chen, Ling and Wei, Yunchao},
  journal={arXiv preprint arXiv:2408.08295},
  year={2024}
}
\bibliographystyle{tmlr}

\appendix

\section{Other Related Works}\label{sec:other-related-works}
In this section, we explore various works that are pertinent to our study.

\paragraph{Alternatives to tuning weights for PEFT.} Among the relevant works, we briefly highlight research that investigates alternatives to tuning transformer weights for adapting to new tasks and/or datasets. Notably, few-shots in-context learning methods have been demonstrated to perform less effectively than PEFT methods \cite{brown2020language, runwal2025peft}. 
Alternative approaches, such as prompt tuning, aim to isolate and preserve the shared knowledge subspaces \cite{zhang2024dept, khattak2023self}, while also learning context vector representations for prompts \cite{zhou2022learning, zhou2022conditional, zhang2024dept}. 
While these works significantly reduce the amount of learnable parameters, additional processing of input data can increase inference latency. 
Moreover, prompt tuning is often limiting beyond the realms of few-shots learning \cite{han2024parameter, wistuba2024choice} and can be outperformed by appropriate low-rank fine-tuning in presence of more data \cite{zanella2024low}. 
As such similar to prior work \cite{albert2025randlora} on this area of parameter efficient fine-tuning, we compare our work to algorithms for weight tuning only.

\paragraph{Tensor networks in machine learning.}
Early work on tensor networks proposed these as standalone models for supervised learning \cite{novikov2016exponential, stoudenmire2016supervised}. Parallel to this, research exploited them for weight compression in convolution neural networks (CNNs) and recurrent neural networks (RNNs) \cite{novikov2015tensorizing, garipov2016ultimate, kim2015compression, yin2022batude}. These techniques have motivated much of the work around tensor networks for fine-tuning. We remark that while tensor networks can significantly compress individual neural-network (NN) layers, they present notable drawbacks in terms of computational efficiency and latency on GPUs due to the need to manage tensor contraction and reshaping \cite{monturiol2025tensorization, lu2025fetta}. 

\section{Further Experiments on MTL using MetaTT-(4+1)D}\label{app:MTL}
In this section we complement the results on MTL from \cref{sec:mtl}. In \cref{tab:MTL-4-task} we analyze MetaTT's performance in MTL for 4 tasks from the GLUE dataset (COLA, MRPC, RTE, QNLI) using the same setup considered in the main text. We observe similar patterns when compared to the results for MTL 3 tasks. LoRA again works remarkably well across different datasets and pre-trained models. MoE-LoRA outperforms methods other than LoRA across different tasks for \rbase, and MetaTT-(4+1)D outperforms methods other than LoRA across all tasks when \rlarge is fine-tuned. However, we observe that on average MetaTT-4D marginally outperforms MetaTT-(4+1)D when \rbase is fine-tuned and is about $1\%$ worse when \rlarge is fine-tuned. The compression ratios achieved by MetaTT-(4+1)D remain analogous to the 3 tasks MTL experiments from the main text (about $22$x compression rate in RoBERTa$_{\text{base}}$ and $43$x compression rate in RoBERTa$_{\text{large}}$, when compared to LoRA).

\begin{table}
\centering
\begin{adjustbox}{max width=\textwidth}
\begin{tabular}{|llcc*{5}{c}|}
\hline
\multirow{2}{*}{\textbf{Model}} & \multirow{2}{*}{\textbf{Method}} & \multirow{2}{*}{\shortstack{\textbf{Param}\\$\times 10^3$}} & \multirow{2}{*}{\textbf{Rank}} & \multicolumn{5}{c|}{\textbf{Metric} (\%)} \\
& & & & {\textbf{CoLA}} &   {\textbf{MRPC}} &  {\textbf{RTE}} & {\textbf{QNLI}} & {\textbf{Avg}} \\
\hline
         & LoRA & $295$ & $8$ & $\mathbf{58.6(9)}$ & $87.1(5)$ & $75(1)$ & $\mathbf{88.0(5)}$ & $\mathbf{77.3(3)}$ \\
         & MTL-LoRA \cite{yang2025mtl} & $296$ & $4$ & $51.5 (1)$ & $87 (1)$ & $\mathbf{76.4 (2)}$ & $87.6(5)$ & $75.6 (5)$ \\
         \rbase & MoE-LoRA \cite{liu2024moe} & $309$ & $8$ & $\mathbf{54.5 (3)}$ & $\mathbf{87.4 (8)}$ & $\mathbf{75.6 (9)}$ & $\mathbf{88.2 (3)}$ & $\mathbf{76.4 (6)}$ \\
         & MetaTT-4D & $13.2$ & $8$ & $53.8(5)$ & $84.8(3)$ & $75(2)$ & $85.9(8)$ & $74.8(7)$ \\
         & MetaTT-(4+1)D & $13.5$ & $8$ & $54(1)$  & $\mathbf{87.3(9)}$ & $70(2)$ & $85.8(4)$ & $74.5(4)$ \\
         \hline
        & LoRA & $786$ & $8$ & $\mathbf{65(1)}$ & $\mathbf{89(1)}$ & $80.8(8)$ & $\mathbf{92.9(4)}$ & $\mathbf{81.7(5)}$ \\
        & MTL-LoRA \cite{yang2025mtl} & $789$ & $4$ & $60 (1)$ & $88.6 (8)$ & $82 (1)$ & $90.2 (3)$ & $80.4 (9)$\\
        \rlarge & MoE-LoRA \cite{liu2024moe} & $814$ & $8$ & $63 (2)$ & $\mathbf{89 (2)}$ & $\mathbf{83 (2)}$ & $90.2 (1)$ & $81 (1)$ \\
        & MetaTT-4D & $18.0$ & $8$ & $59(1)$ & $88.2(5)$ & $82.4(4)$ & $90.3(5)$ & $80.0(5)$ \\
         & MetaTT-(4+1)D & $18.3$ & $8$ & $\mathbf{63(1)}$ & $87.6(8)$ & $\mathbf{83.4(9)}$ & $\mathbf{91.1(2)}$ & $\mathbf{81.2(5)}$\\
        \hline
    \end{tabular}
    \end{adjustbox}
    \caption{\textbf{Results of MTL with 4 tasks.} We observe that unlike results in \cref{tab:MTL} MetaTT-(4+1)D outperforms MetaTT-4D only when using \rlarge. However, across both models, MetaTT-(4+1)D uses about only $300$ more trainable parameters. For \rbase MetaTT-(4+1)D performs within $1\%$ of MTL-LoRA while using $\approx22$x less parameters, and for \rlarge MetaTT-(4+1)D outperforms both MTL-LoRA and MoE-LoRA on average, and is within $0.5\%$ of average accuracy of LoRA, while using $\approx43$x less parameters. We show in bold the two best accuracies per task.
    }
    \label{tab:MTL-4-task}
    \vspace{-1em}
\end{table}

\begin{figure}[h]
    \centering
    \includegraphics[width=1\linewidth]{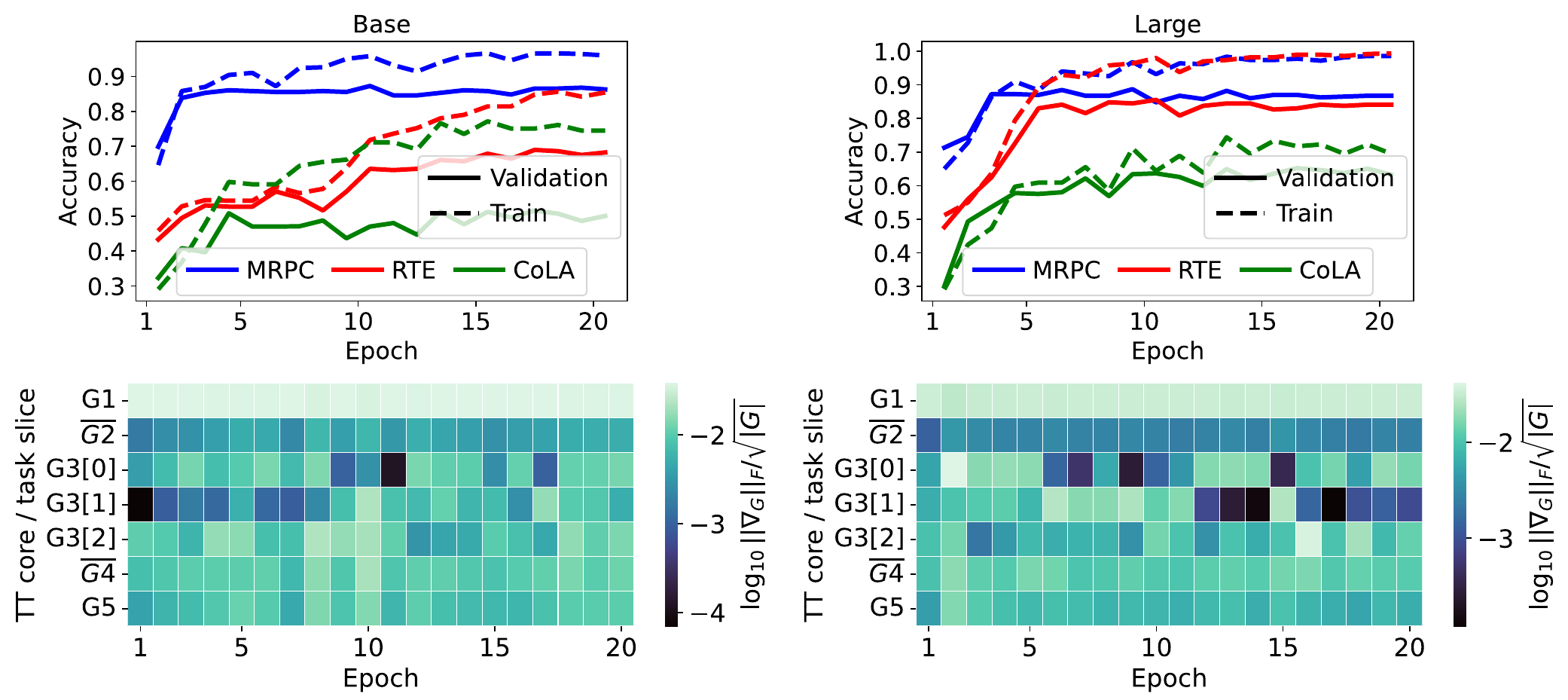}
    \vspace{-1em}
    \caption{\textbf{Influence of task-dependent TT core in MTL.} (Left): (Top): accuracy of MetaTT-(4+1)D as a function of epochs for RoBERTa$_{\text{Base}}$ for a single training realization (in the case of CoLA we compute Matthew's correlation instead). (Bottom): Corresponding normalized gradients across all tensors as a function of epochs (see \cref{app:MTL}). Task labels correspond to $0$: MRPC, $1$: RTE, $2$: CoLA. (Right): Same as in left but for RoBERTa$_{\text{Large}}$ as pretrained model.}
    \label{fig:gradient_heatmap}
\end{figure}

\begin{figure}[h]
    \centering
    \includegraphics[width=1\linewidth]{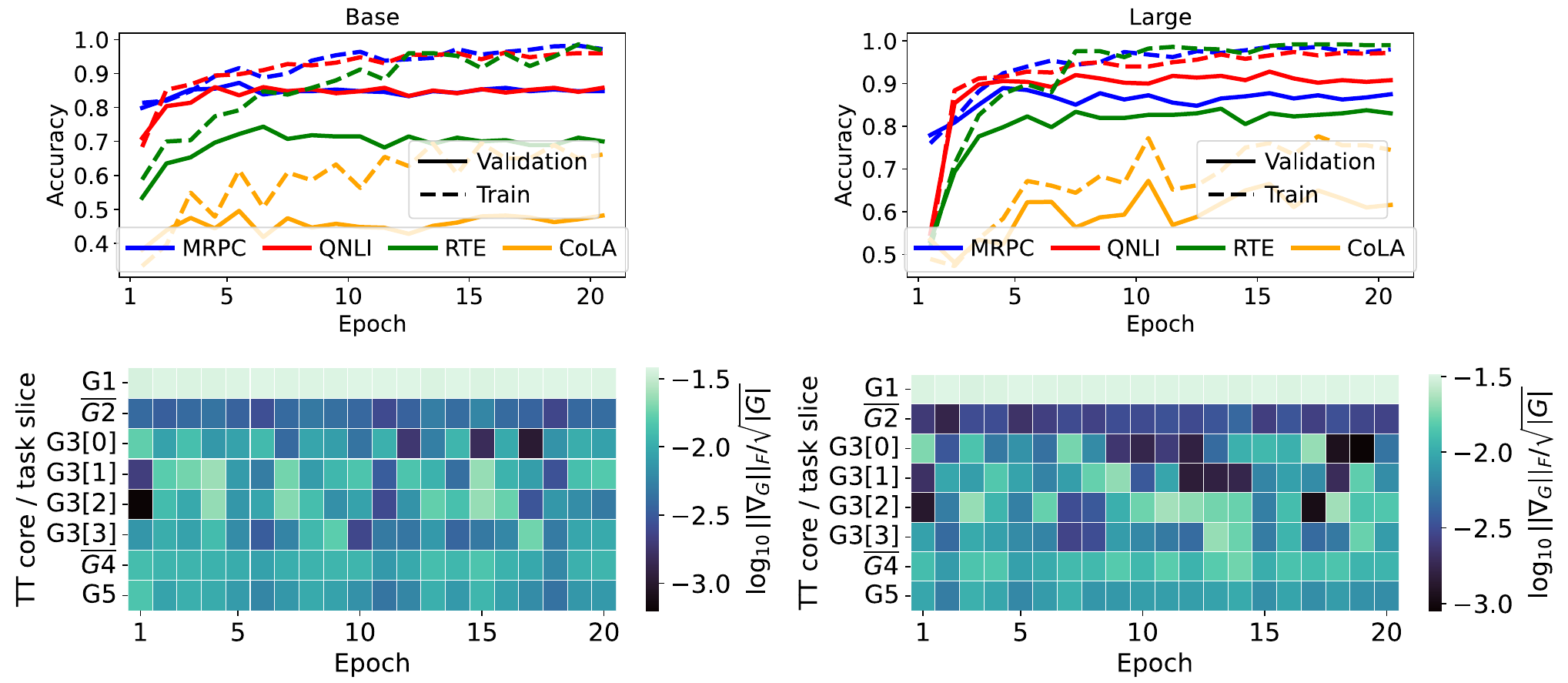}
    \vspace{-1em}
    \caption{\textbf{Influence of task-dependent TT core in MTL.} (Left): (Top): accuracy of MetaTT-(4+1)D as a function of epochs for RoBERTa$_{\text{Base}}$ for a single training realization (in the case of CoLA we compute Matthew's correlation instead). (Bottom): Corresponding normalized gradients across all tensors as a function of epochs (see \cref{app:MTL}). Task labels correspond to $0$: MRPC, $1$: QNLI, $2$: RTE, $3$: CoLA. (Right): Same as in left but for RoBERTa$_{\text{Large}}$ as pretrained model.}
    \label{fig:gradient_heatmap_v2}
\end{figure}

Next, we provide further evidence that the task-related tensor cores in MetaTT-(4+1)D used in play a significant role. For any given layer index $l$, matrix-type index $m$, and task index $t$, a given input batched vector gets updated as 
\begin{equation}
    X \gets X\cdot W_{l,t,m}^T+\alpha X\cdot\mathcal{G}_1\mathcal{G}_2[l]\mathcal{G}_3[t]\mathcal{G}_4[m]\mathcal{G}_5.
\end{equation}
To show the impact of the inclusion of task-dependent TT cores, in \cref{fig:gradient_heatmap} and \cref{fig:gradient_heatmap_v2} we plot heatmaps of the gradients across each tensor in the TT for a particular realization, for the 3 and 4 tasks MTL experiments, respectively. Since the boundary cores $\mathcal{G}_1$, $\mathcal{G}_5$ are much larger than the rest of the cores, we normalize the gradients across each TT core by the number of non-zero elements as follows -- $||\nabla_{\mathcal{G}}||_F/\sqrt{|\mathcal{G}|}$ where $\|\cdot\|_F$ is the Frobenius norm and $|\mathcal{G}|$ is the number of non-zero elements of the tensor $\mathcal{G}$. For tensors $\mathcal{G}_2$ and $\mathcal{G}_4$ we plot the average gradients across all layers and matrix-types, respectively. We observe that indeed the tensor $\mathcal{G}_3$ is acquiring significant gradient updates, especially for RoBERTa$_{\text{Large}}$. Moreover, for certain epochs, we find that $\mathcal{G}_3$ in fact acquires the largest gradients across all tensors. Interestingly, we find that the task core with label $2$ in \cref{fig:gradient_heatmap} and label $3$ in \cref{fig:gradient_heatmap_v2} corresponding in both cases to CoLA receive the largest gradient update. This is expected since CoLA is the hardest task among the chosen sets of tasks. Similar to \cref{sec:mtl}, the rank chosen in these experiments is $8$ across all bonds. Other relevant hyper-parameters used are: $\text{batch size} = 16$, $\alpha=2$, $\text{learning rate} = 5e-4$. Furthermore, we perform gradient clipping with a maximum gradient value of $3.0$.

We complement plots of gradients with the downstream task performance per epoch on each of the plots. While it is generally hard to make direct comparisons between gradients observed and downstream task performance, in \cref{fig:gradient_heatmap} we observe that for both RoBERTa$_{\text{Base}}$ and RoBERTa$_{\text{Large}}$ and the RTE dataset, the gradients observed at each epoch at tensor core $\mathcal{G}_3[1]$ correlate with the downstream task performance of the model. We also see similar trends for CoLA in \cref{fig:gradient_heatmap_v2}.

\section{Further Details on the Experiments with DMRG}\label{app:DMRG}
The results of \cref{sec:dmrg} show that, for a given target rank ($r=4$ for both MRPC and RTE datasets), interspersing DMRG-inspired sweeps to progressively bring down the TT ranks from a high enough rank ($r=10$ in this case) leads to higher accuracies than training via AdamW with that fixed target rank. Interestingly, the rank schedule philosophy used here in DMRG is the mirror image of the one commonly used in many-body physics: there, one starts with small ranks and progressively increments these so as to capture more precisely the target \textit{ground state} (see \cite{schollwock2011density} and references therein). Instead, in ML settings such as ours, the TT rank serves as a regularizer; pruning redundant directions after the optimiser has identified them improves generalization and reduces memory, whereas too high rank risks overfitting. 

The choice of rank schedule was done heuristically, with the only consideration in mind that the ranks should be reduced slowly so that the model can adapt to the new weight space more efficiently. We see two potential extensions that find such rank schedules in more principled ways and that we leave open for future work.

First, for our experiments we used the magnitude of the singular values across TT bonds as diagnostic to shrink the ranks (even if they all remained high relative to each other). One improvement could come in the form of considering other \textit{importance scores} that take into account the sensitivity of those singular values to the loss function. An approach similar in spirit was done in the context of LoRA type adapters in AdaLoRA \cite{zhang2023adalora}. We remark here that one advantage of performing rank adaptive schemes based on SVDs in MetaTT over LoRA type adapters is that a much smaller fraction of SVDs are needed in MetaTT than in LoRA. This series of SVDs at the end of certain epochs result in a small overhead. This is in contrast to performing SVDs on all LoRA type adapters across the transformer architecture. It is for this reason that the orthogonality condition on the isometry factors stemming from SVDs are enforced through regularizers in AdaLoRA, which encourages low-rank structure but does not strictly guarantee rank shrinkage at the end of training. In \cref{sec:adalora_vs_dmrg} we show results of MetaTT with DMRG-inspired sweeps \textit{vs.} AdaLoRA in the context of fine-tuning Llama-2 models, and in \cref{sec:compl_ada_vs_dmrg} we present a complexity analysis of performing SVDs on LoRA-type adapters \textit{vs.} SVDs on MetaTT-type adapters.

A second approach, which follows the original DMRG algorithm closer in spirit is to use powerful local optimizers to minimize directly the loss function with respect to each merged tensor at each step of the DMRG-inspired sweep in \cref{alg:DMRG}. This would not only enable rank adaptation across each TT bond, but also directly optimize the loss function which may result in a powerful optimizer.

\section{Stability analysis}

We compare stability of the performance of LoRA, VeRA, and MetaTT-4D for fine-tuning \rbase and \rlarge on the validation data of CoLA, MRPC and RTE. \rev{We also include MetaTT-5D in this analysis for completeness (see \cref{tab:roberta-5d} in \cref{sec:5d-results} for the corresponding accuracy results).} We study the stability of these adapters when ranks and learning rates are varied while keeping other hyper-parameters constant. 

\subsection{Stability during training using the best set of hyperparameters}
\paragraph{Fixed hyper-parameters.} For these experiments, we fix the random seeds to $\{1,2,3,4,5,6,7,8\}$ across all algorithms and models. The hyper-parameters used for LoRA is reported in \cref{tab:LoRA-hpams-stability}. Note, \cite{bershatsky2024lotr} which was used to report LoRA values in \cref{tab:roberta-combined} only provided the range of hyper-parameter values swept, and not the final set of hyper-parameters, and so we ran our own search on a wider set of seeds. The hyper-parameters for other methods are consistent with the best set of hyper-parameters reported in \cref{app:exp-deets}
\begin{table}[h]
    \centering
    \begin{tabular}{|l| l |c |c| c| c|}
        \hline
        \textbf{Model} & \textbf{Task} & \textbf{Rank} & $\alpha$ & \textbf{Learning Rate} & \textbf{Batch Size} \\
        \hline
        & CoLA & 4 & 8.0 & $5e-4$ & 16\\
        \rbase  & MRPC & 32  & 64.0 & $2e{-4}$ & 8  \\
        & RTE  & 8  & 16.0 & $2e{-4}$ & 16 \\
        \hline
        & CoLA & 64 & 128.0 & $2e-4$ & 16 \\
        \rlarge & MRPC & 32 & 64.0 & $2e{-4}$ & 8  \\
        & RTE  & 8  & 16.0 & $2e{-4}$ & 16 \\
        \hline
        \end{tabular}
        \caption{Best hyperparameters for each model and task combination for LoRA.}
        \label{tab:LoRA-hpams-stability}
\end{table}

\paragraph{Stability score.} To quantify stability of each adapters on the best corresponding set of hyper-parameters, we report margin of error (also known as the half-width of the $95\%$ confidence interval) \cite{krishnamoorthy2006handbook} defined as
\begin{align}
    \operatorname{Margin\ of\ error} \approx 1.96 \frac{\sigma}{\sqrt{n}},
\end{align}
where $1.96$  is the critical value expressed as a $z$-score and corresponds to $95\%$ confidence level, assuming that the data is normally distributed. Note, the margin of error reported here is at most a factor of $2$ worse than the standard error.

\begin{figure}[!h]
    \centering
    \begin{subfigure}[b]{0.48\linewidth}
        \centering    
        \includegraphics[width=\linewidth]{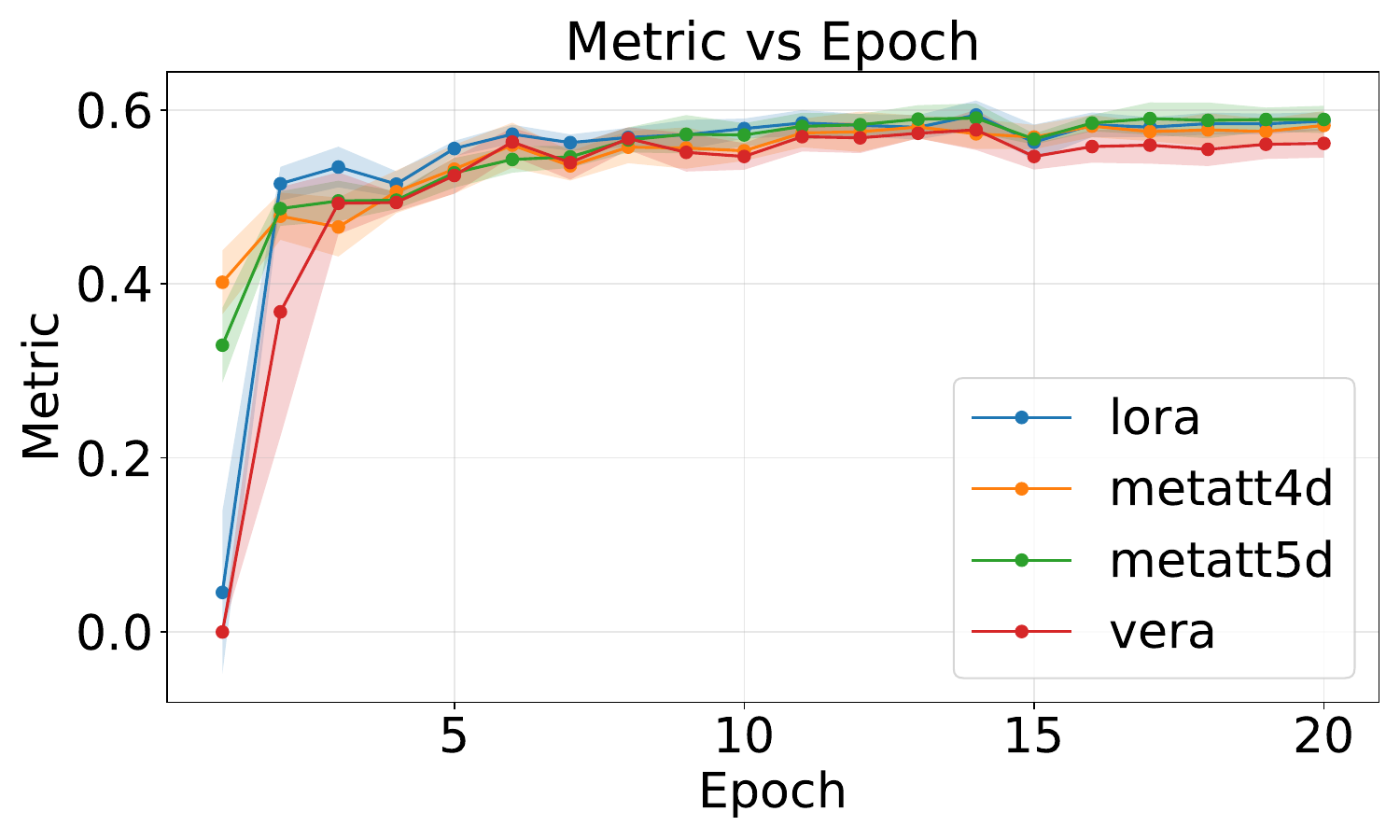}
        \caption{\rbase, CoLA}
    \end{subfigure}
    \begin{subfigure}[b]{0.48\linewidth}
        \centering
        \includegraphics[width=\linewidth]{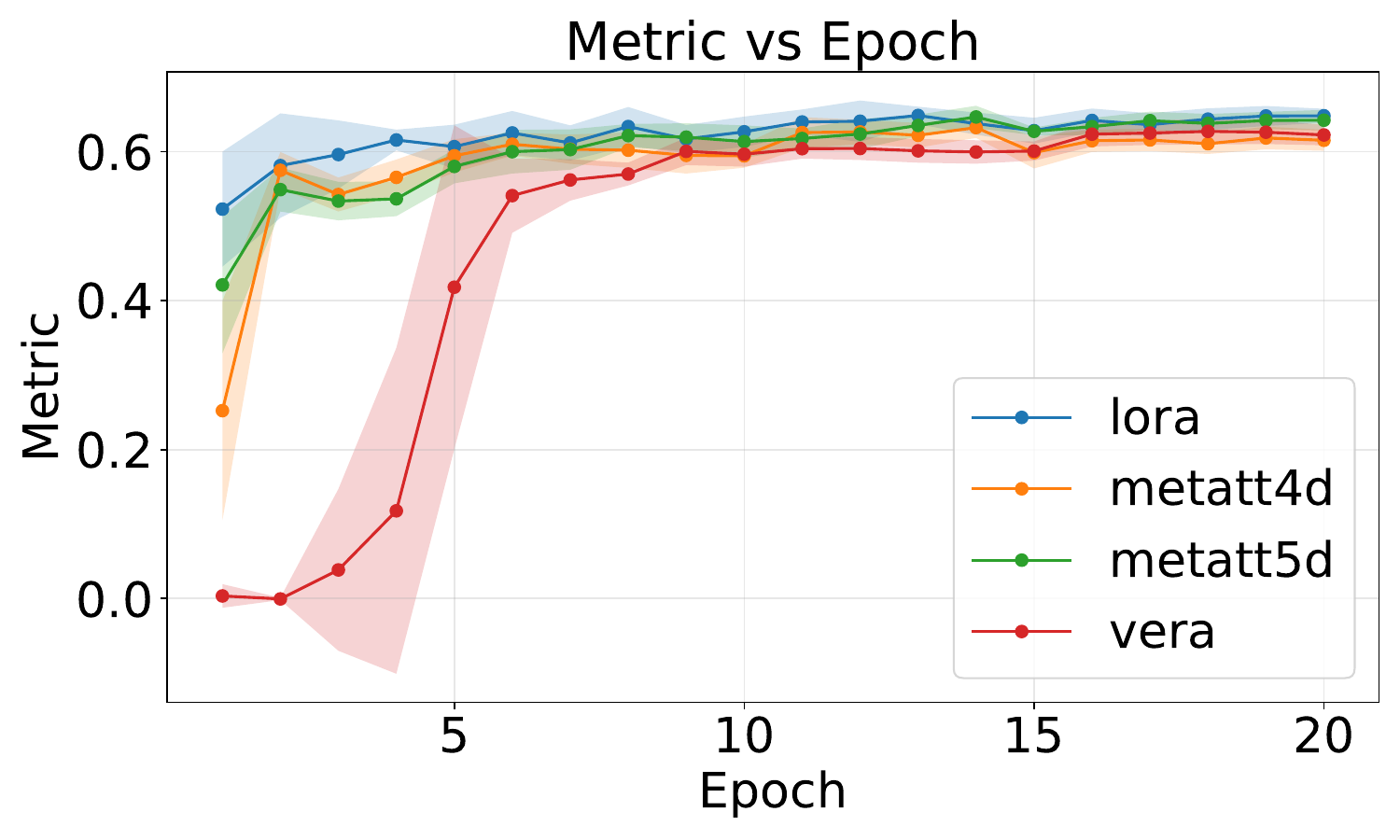}
        \caption{\rlarge, CoLA}
    \end{subfigure}
    \begin{subfigure}[b]{0.48\linewidth}
        \centering    
        \includegraphics[width=\linewidth]{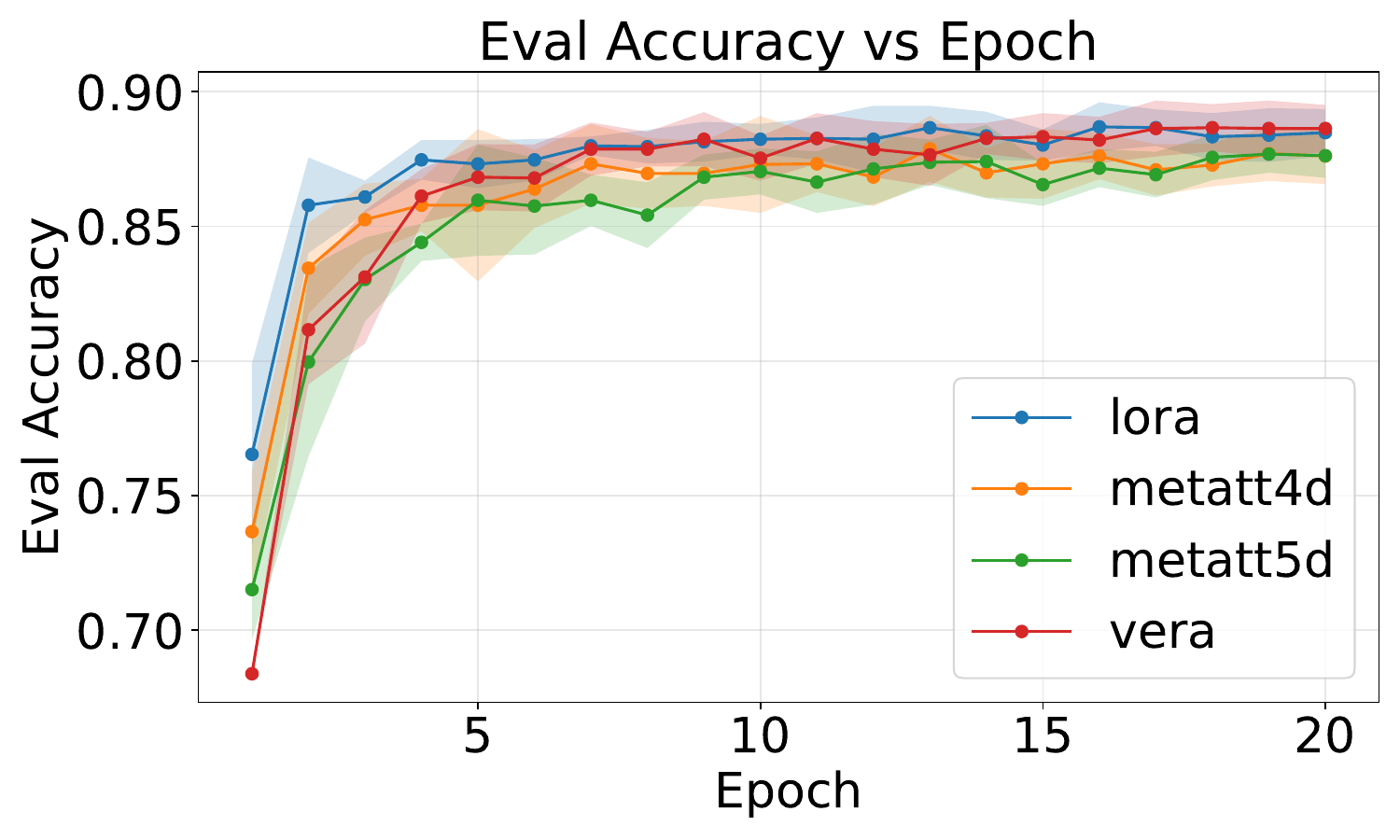}
        \caption{\rbase, MRPC}
    \end{subfigure}
    \begin{subfigure}[b]{0.48\linewidth}
        \centering
        \includegraphics[width=\linewidth]{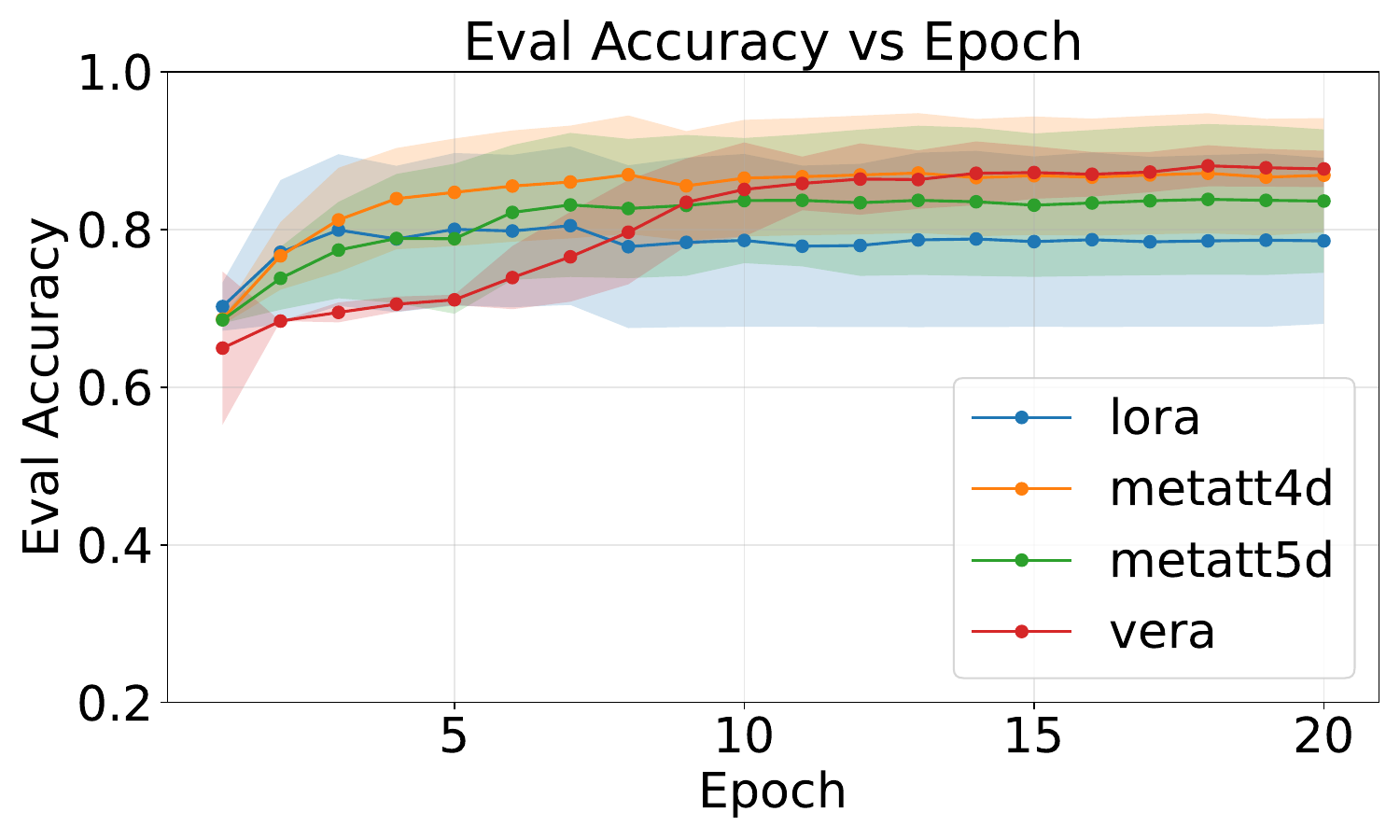}
        \caption{\rlarge, MRPC}
    \end{subfigure}
    \begin{subfigure}[b]{0.48\linewidth}
        \centering    
        \includegraphics[width=\linewidth]{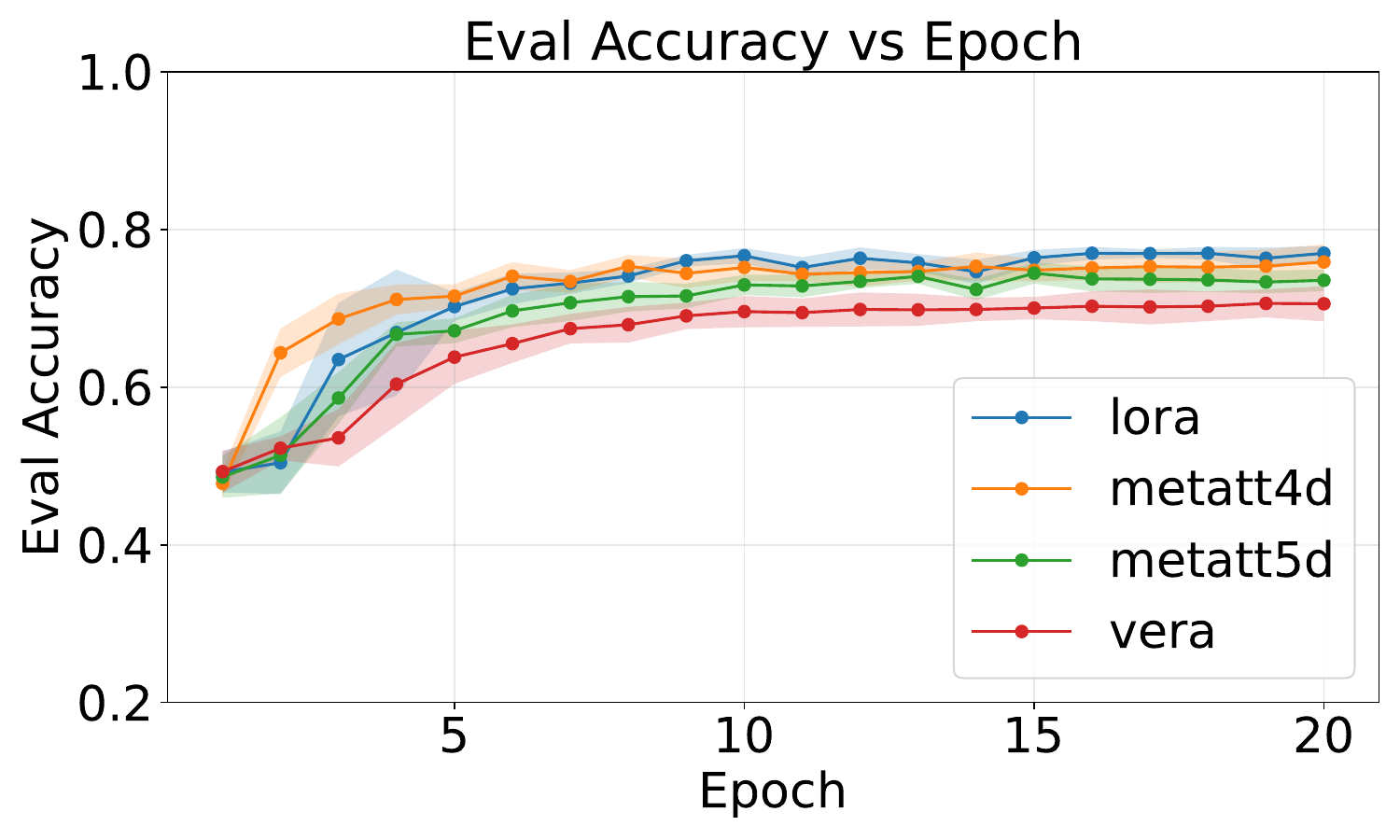}
        \caption{\rbase, RTE}
    \end{subfigure}
    \begin{subfigure}[b]{0.48\linewidth}
        \centering    
        \includegraphics[width=\linewidth]{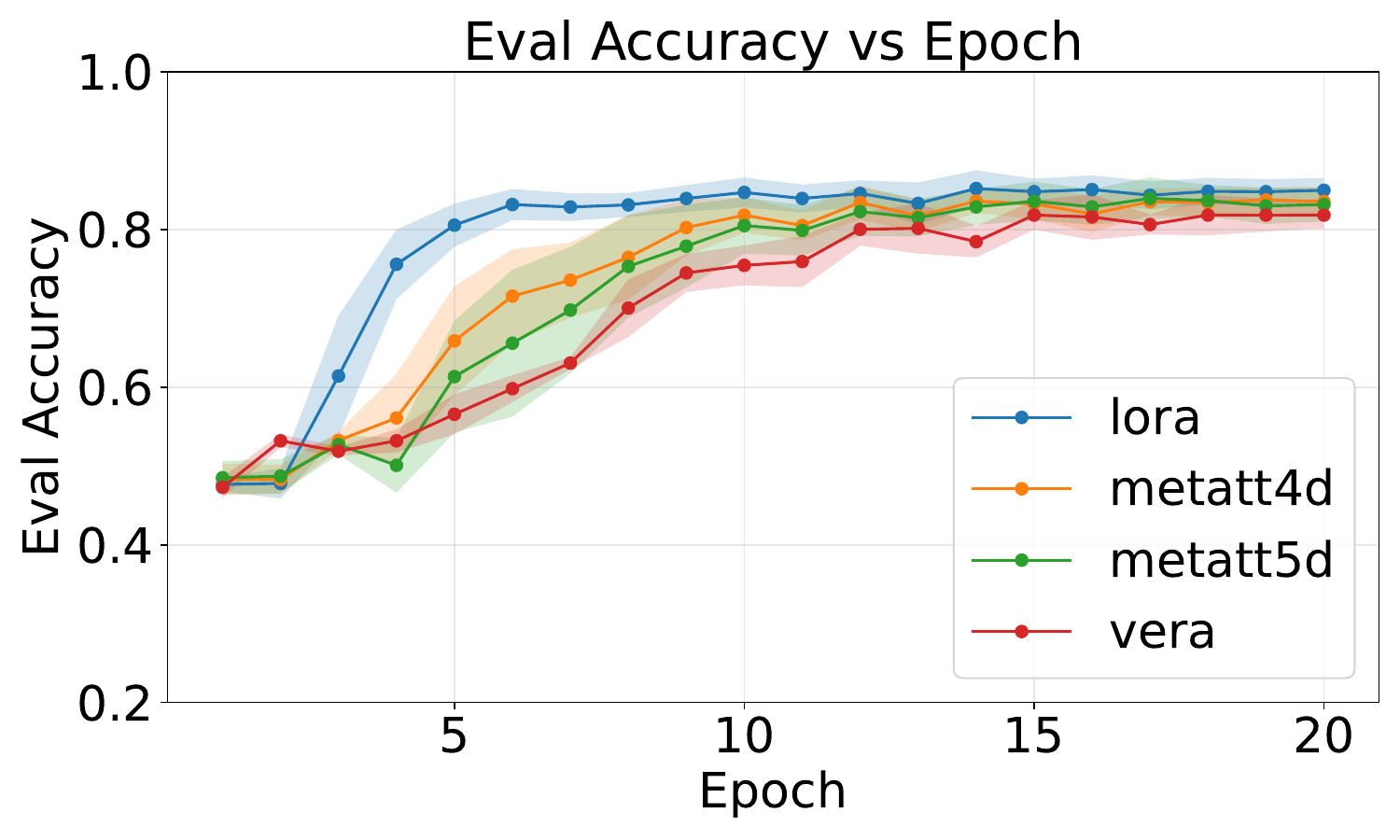}
        \caption{\rlarge, RTE}
    \end{subfigure}
    \caption{\textbf{Variance during training.} Here we plot the evaluation accuracy (Matthew's correlation coefficient for CoLA) as we train the respective model on specific finetuning tasks using LoRA, VeRA, MetaTT-4D, and MetaTT-5D adapters. We observe that on average the variance across LoRA, MetaTT-4D, and MetaTT-5D are approximately similar, except for \rbase on CoLA where MetaTT-5D demonstrates significantly greater variance in training across seeds. \rev{This increased instability of MetaTT-5D is one reason we recommend MetaTT-4D as the default variant.}}
    \label{fig:variance-during-training}
\end{figure}

\paragraph{Observations.} In \cref{fig:variance-during-training} we plot the evaluation accuracy as we increase epochs during training. We observe that all methods demonstrate similar variance, across seeds and epochs. To measure the stability of the performance of the final model, we measure corresponding stability scores and report them in \cref{tab:stability-best-hpams}. We observe that between models, tasks and adapters, the half width of the $95\%$ confidence interval does not vary by much. As a result, we can conclude that for the set of hyper-parameters that yield the highest evaluation accuracies across methods, the stability during training across adapter variants remains relatively similar.
\begin{table}[h]
    \centering
    \begin{tabular}{|l|l|c|c|c|c|}
         \hline
         \textbf{Model} & \textbf{Task} & \textbf{LoRA} & \textbf{VeRA} & \textbf{MetaTT-4D} & \textbf{MetaTT-5D}\\
         \hline
         & CoLA & \textbf{0.009} & 0.012 & 0.012 & \textbf{0.011} \\
         \rbase & MRPC & \textbf{0.006} & \textbf{0.006} & \textbf{0.006} & \textbf{0.006} \\
         & RTE & \textbf{0.007} & 0.016 & 0.011 & \textbf{0.010} \\
         \hline
         & CoLA & \textbf{0.007} & 0.013 & 0.154 & \textbf{0.010} \\
         \rlarge & MRPC & \textbf{0.004} & 0.007 & \textbf{0.006} & 0.065 \\
         & RTE & 0.011 & \textbf{0.008} & \textbf{0.010} & 0.093 \\
         \hline
    \end{tabular}
    \caption{\textbf{Margin of error across seeds.} We report the half-width of the $95\%$ confidence interval for finetuning \rbase and \rlarge using LoRA, VeRA, MetaTT-4D, and MetaTT-5D adapters. The lowest two values in each row are shown in bold. }
    \label{tab:stability-best-hpams}
\end{table}

\subsection{Stability across learning-rates}\label{sec:stability-lr}

\begin{figure}[!h]
    \centering
    \begin{subfigure}[b]{0.4\linewidth}
        \centering
        \includegraphics[width=\linewidth]{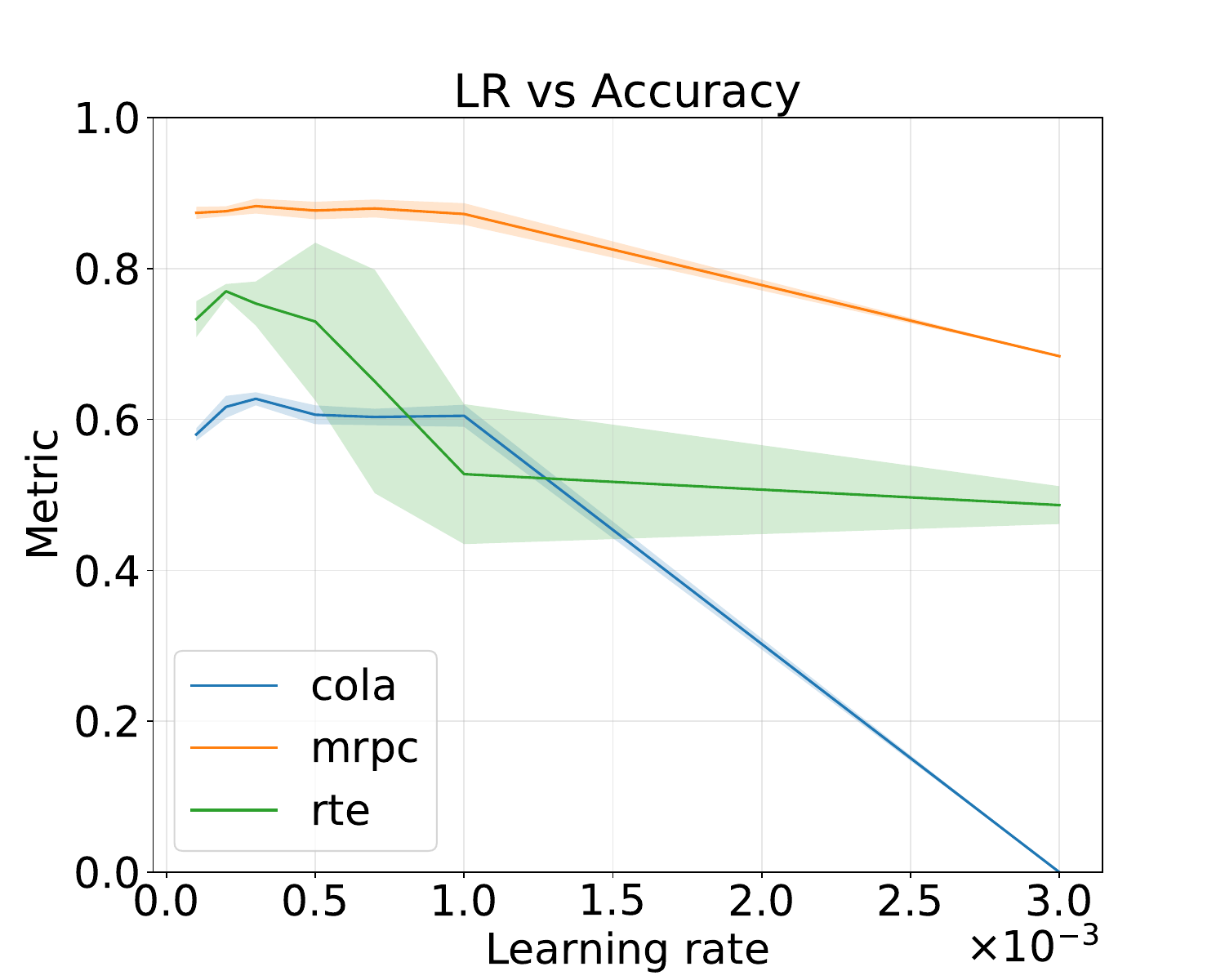}
        \caption{\rbase with LoRA.}
    \end{subfigure}
    \begin{subfigure}[b]{0.4\linewidth}
        \centering
        \includegraphics[width=\linewidth]{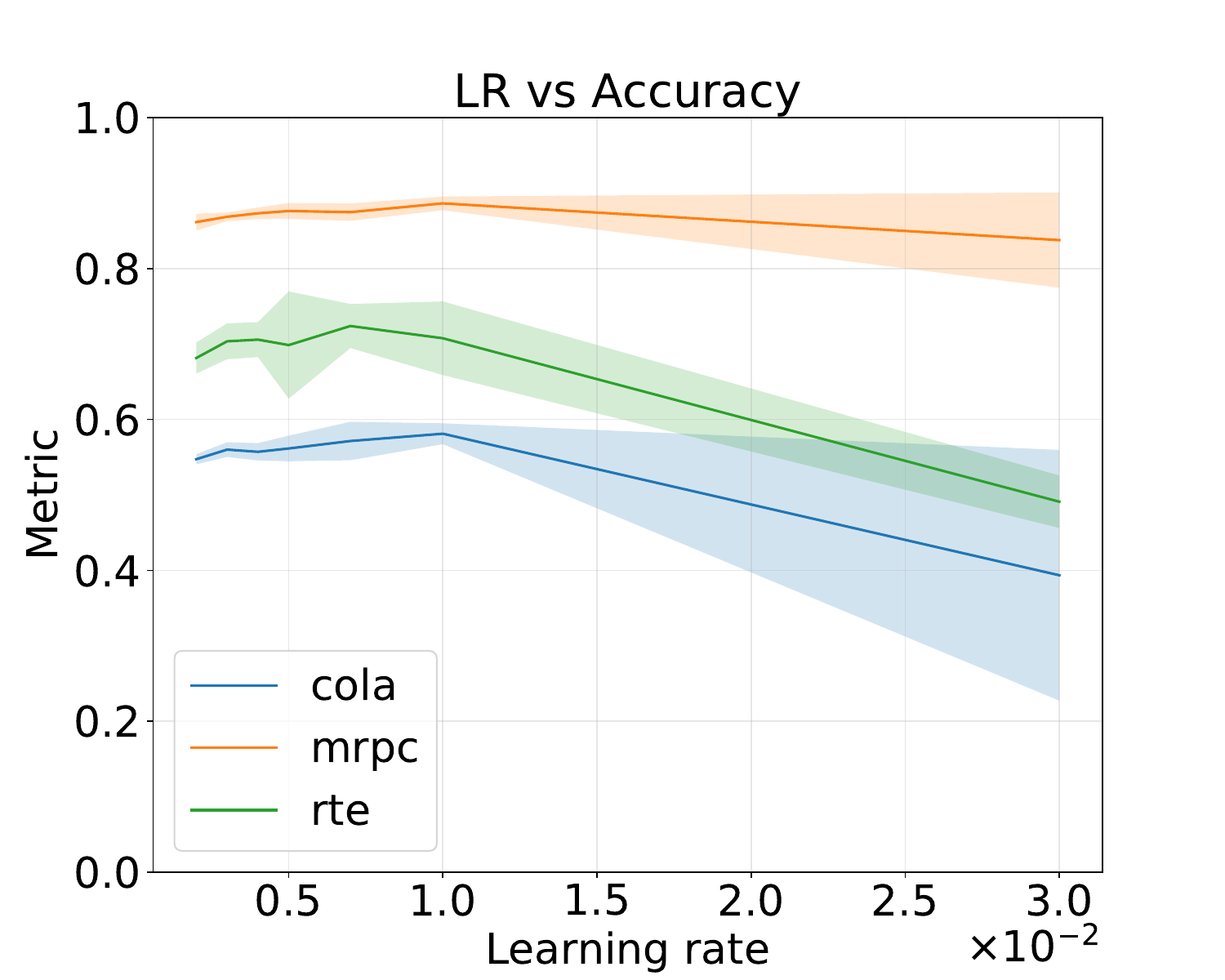}
        \caption{\rbase with VeRA.}
    \end{subfigure}
    \begin{subfigure}[b]{0.4\linewidth}
        \centering
        \includegraphics[width=\linewidth]{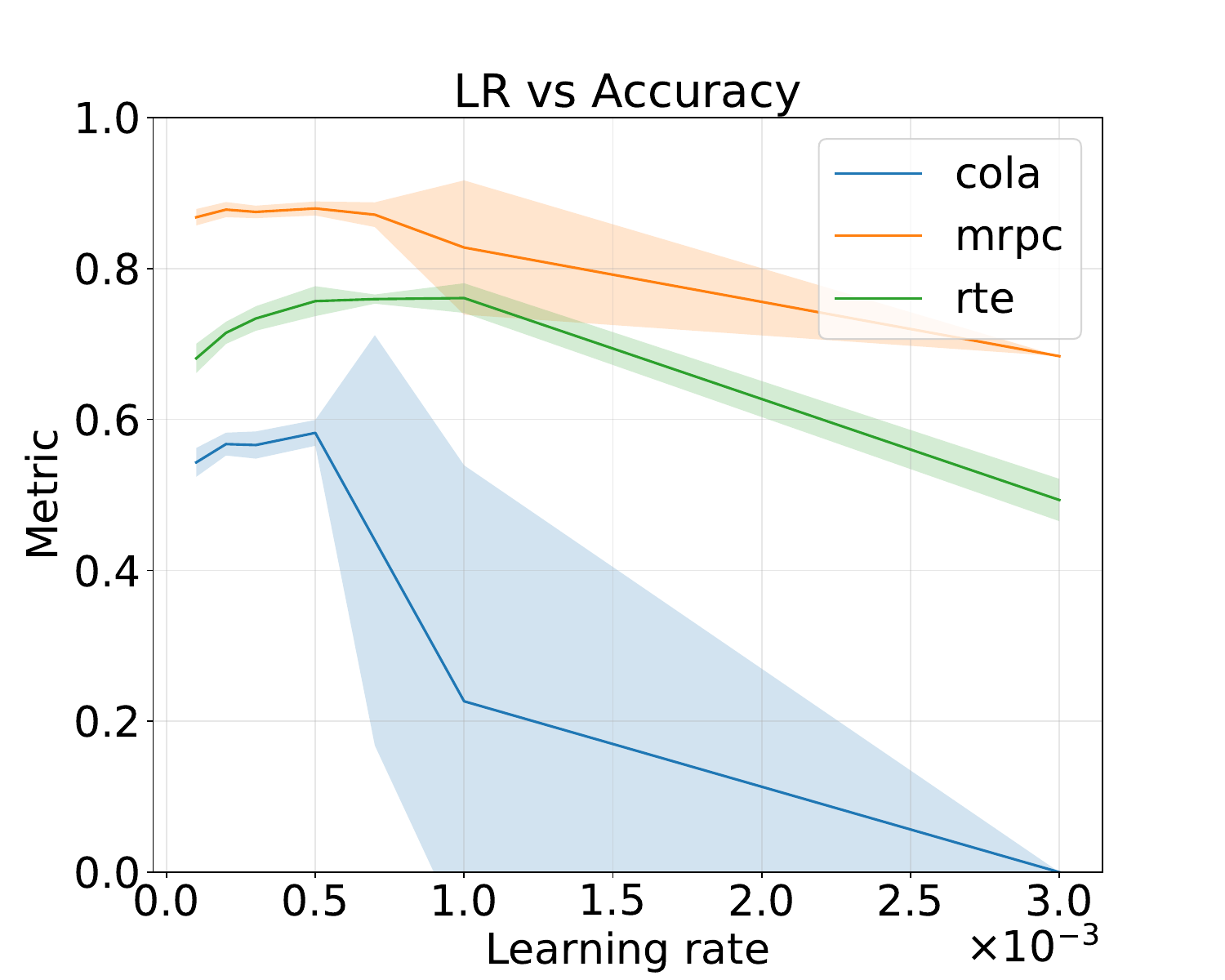}
        \caption{\rbase with MetaTT-4D.}
    \end{subfigure}
    \begin{subfigure}[b]{0.4\linewidth}
        \centering
        \includegraphics[width=\linewidth]{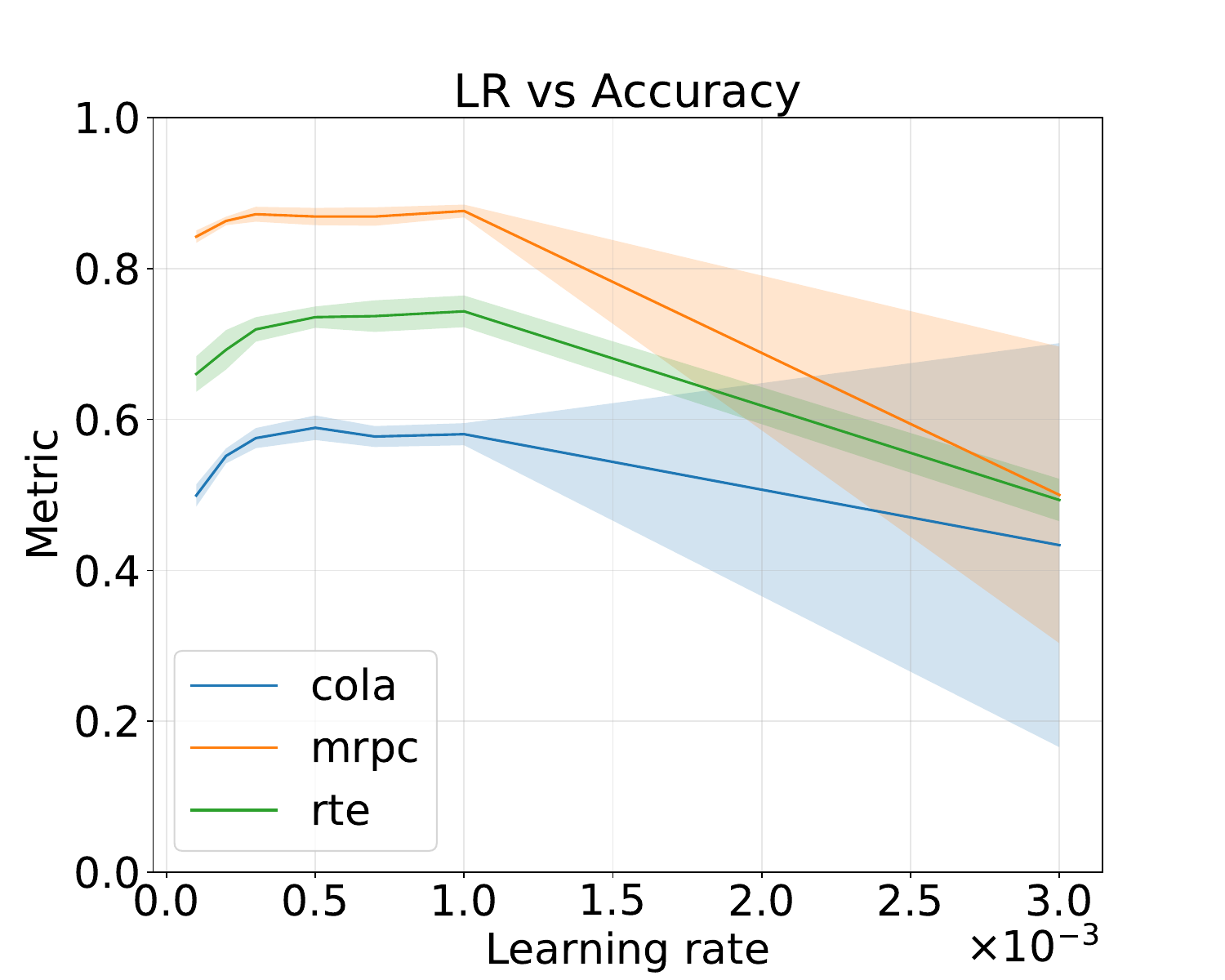}
        \caption{\rbase with MetaTT-5D.}
    \end{subfigure}
    \caption{\textbf{Learning rate vs accuracy.} We plot final accuracy of \rbase when trained with specific adapters on specific glue tasks. We observe that in general variants of MetaTT decays somewhat at a similar rate when compared to LoRA and VeRA. This is inspite of the fact that for LoRA and VeRA the parameter $\alpha$ are tied to the ranks and so varied across runs (or in case of VeRA already absorbed in the learning parameters), while $\alpha$ remains an independent parameter for variants of MetaTT and was treated asfixed across learning rates.}
    \label{fig:stability-across-LRs-base} 
\end{figure}

\begin{figure}[!h]
    \centering
    \begin{subfigure}[b]{0.4\linewidth}
        \centering
        \includegraphics[width=\linewidth]{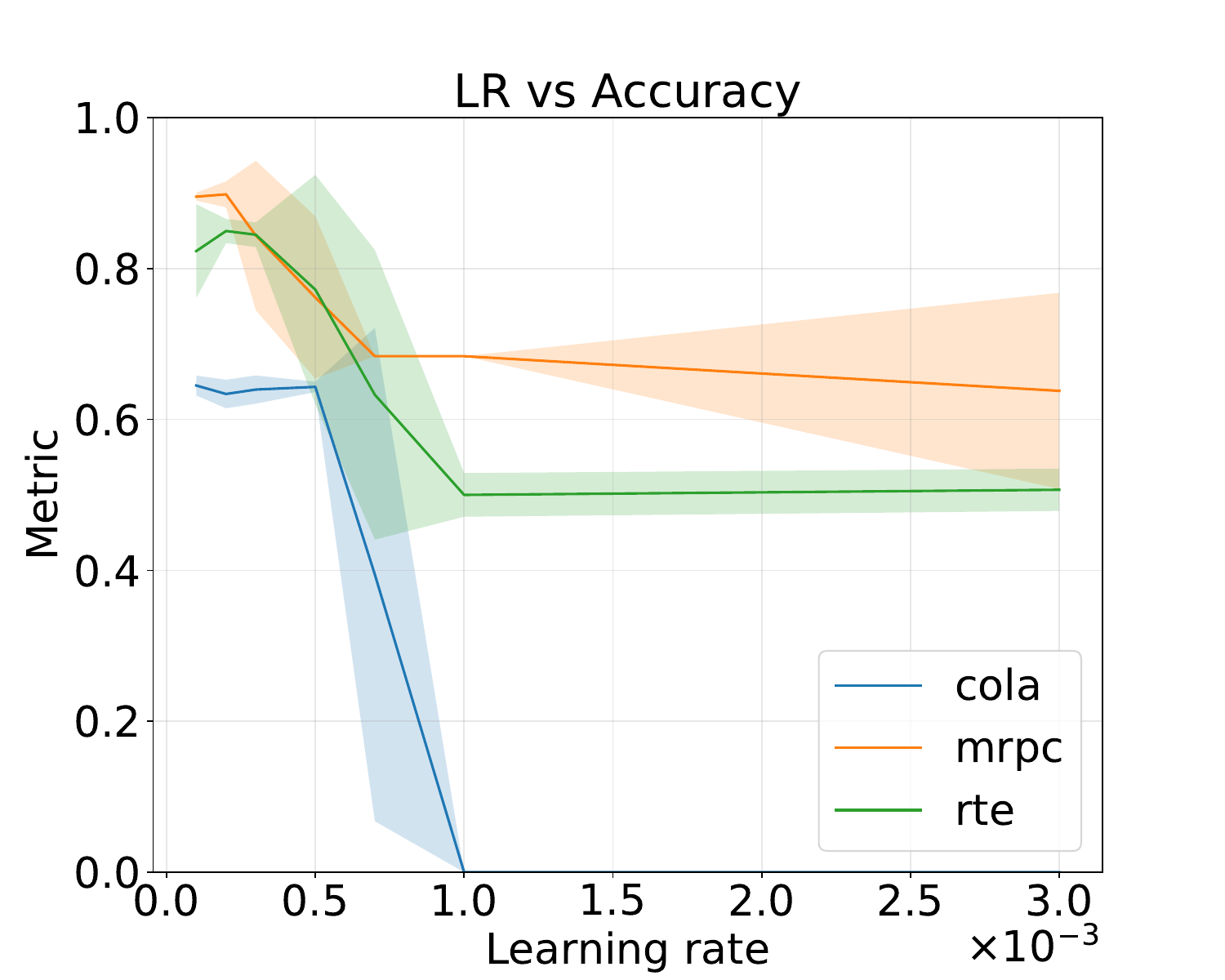}
        \caption{\rlarge with LoRA.}
    \end{subfigure}
    \begin{subfigure}[b]{0.4\linewidth}
        \centering
        \includegraphics[width=\linewidth]{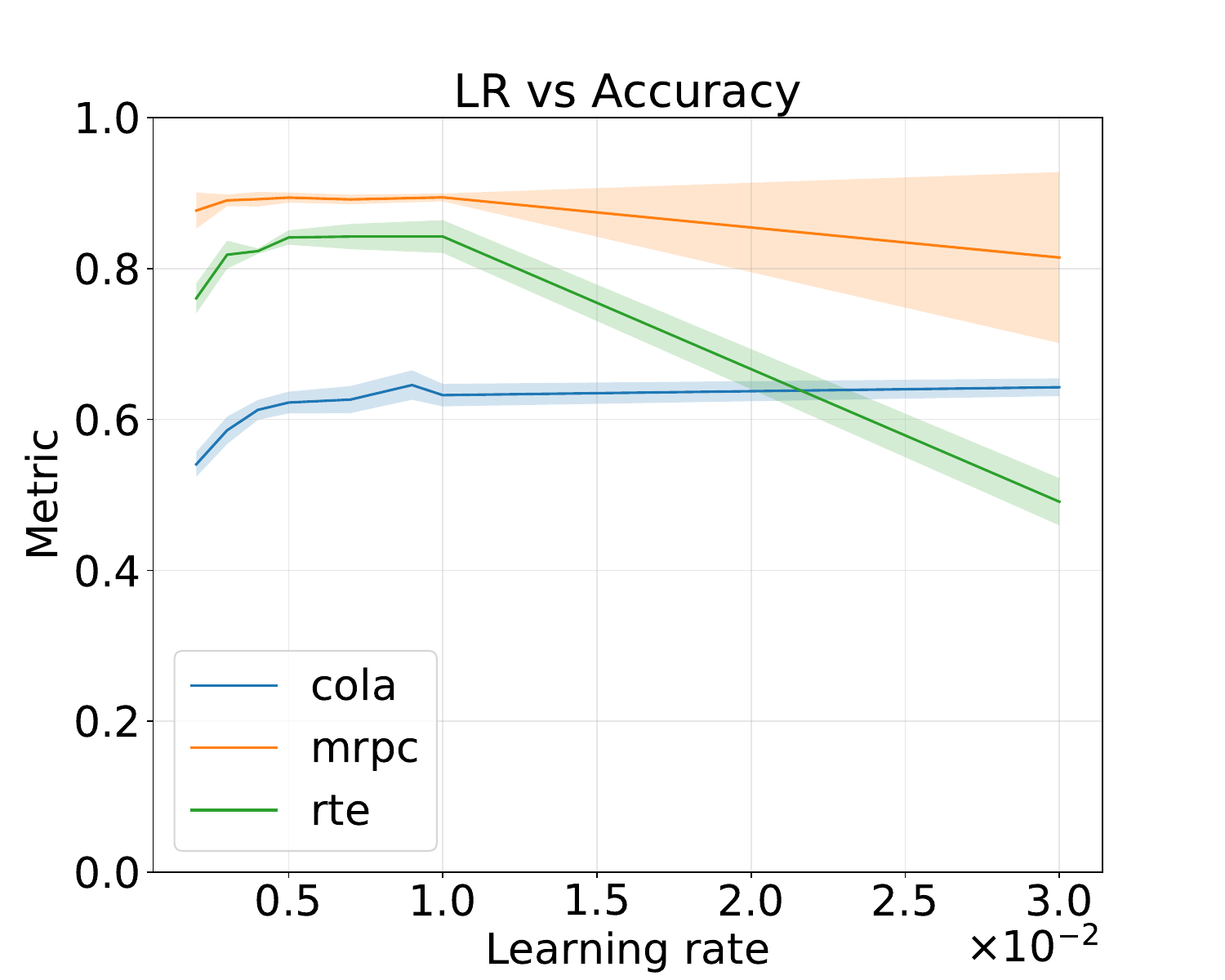}
        \caption{\rlarge with VeRA.}
    \end{subfigure}
    \begin{subfigure}[b]{0.4\linewidth}
        \centering
        \includegraphics[width=\linewidth]{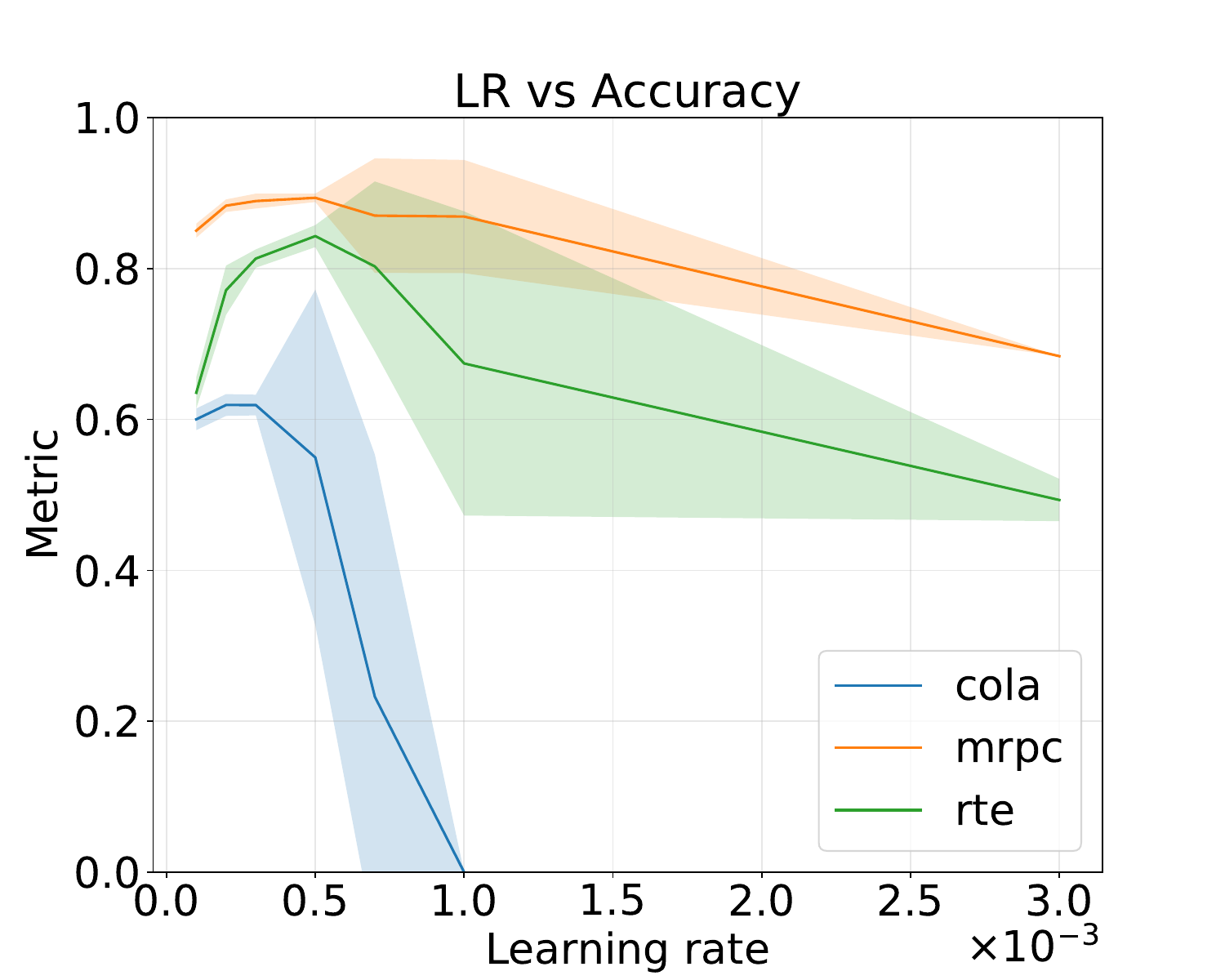}
        \caption{\rlarge with MetaTT-4D.}
    \end{subfigure}
    \begin{subfigure}[b]{0.4\linewidth}
        \centering
        \includegraphics[width=\linewidth]{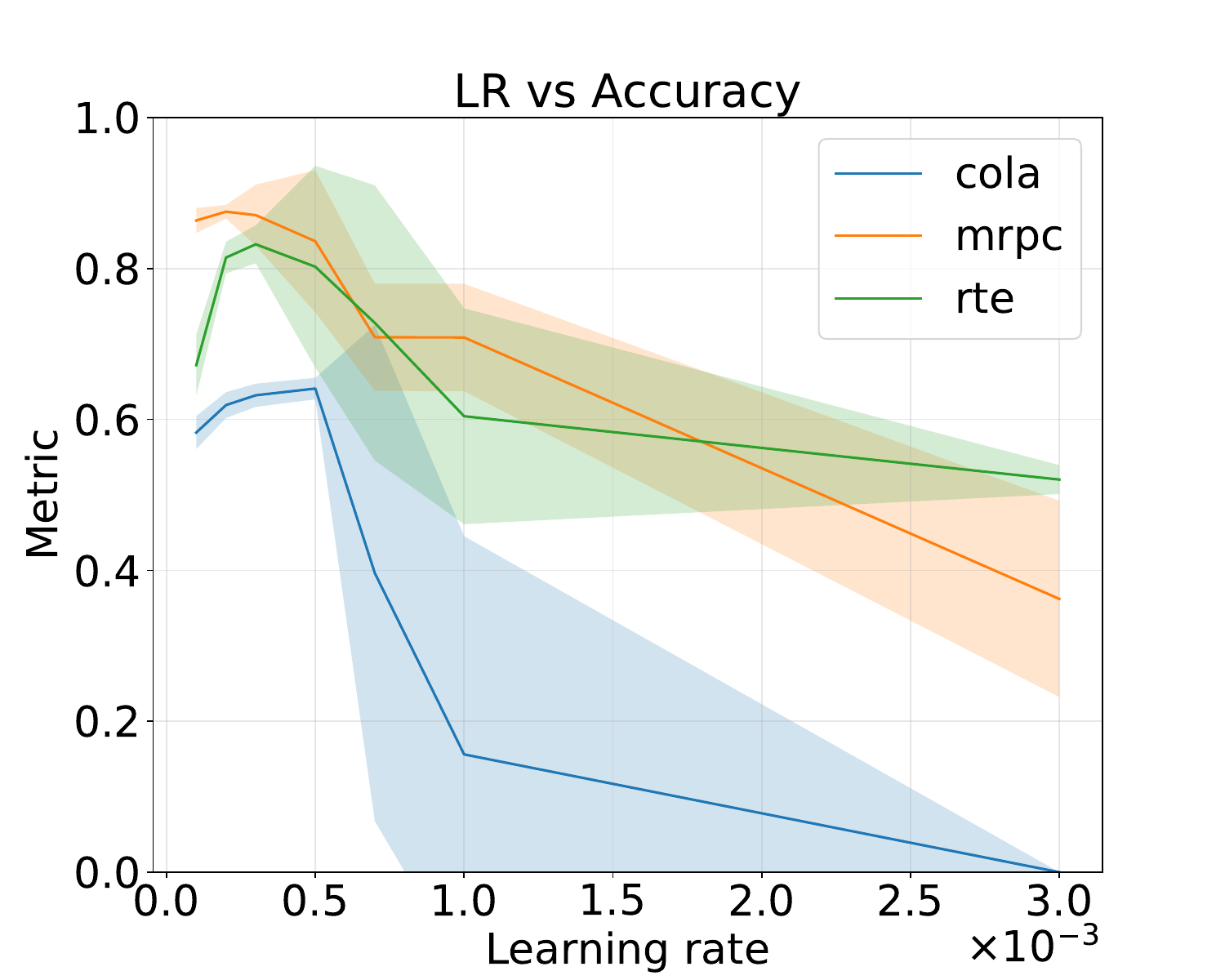}
        \caption{\rlarge with MetaTT-5D.}
    \end{subfigure}
    \caption{\textbf{Learning rate vs accuracy.} We plot final accuracy of \rlarge when trained with specific adapters on specific glue tasks. We observe that for MRPC the rate of performance decay for VeRA is better than other methods, and for RTE the rate of performance decay for MetaTT-5D is better than other methods. }
    \label{fig:stability-across-LRs-large}
\end{figure}

We also plot the performance of each of the adapters when the learning rate is varied while other hyper-parameters are kept constant in \cref{fig:stability-across-LRs-base} and \cref{fig:stability-across-LRs-large} (we choose the best reported hyper-parameters for these sweeps in \cref{app:exp-deets}). We observe that when finetuning \rbase and \rlarge on MRPC and RTE, the decay in performance after attaining the best rate is similar in LoRA and VeRA, and MetaTT-4D also performs similar to these adapters (approximately similar slope of the mean). For fine-tuning on CoLA, MetaTT-4D and LoRA behave similarly (with larger variance in case of MetaTT-4D). One must note that the $\alpha$ is varied for LoRA (as $2\times \operatorname{rank}$) and VeRA doesn't depend on $\alpha$, which makes this comparison harder for both variants of MetaTT. We observe similar trends across tasks for \rlarge for LoRA and MetaTT-4D.

\subsection{Stability across ranks}

\begin{figure}[!h]
    \centering
    \begin{subfigure}[b]{0.4\linewidth}
        \centering
        \includegraphics[width=\linewidth]{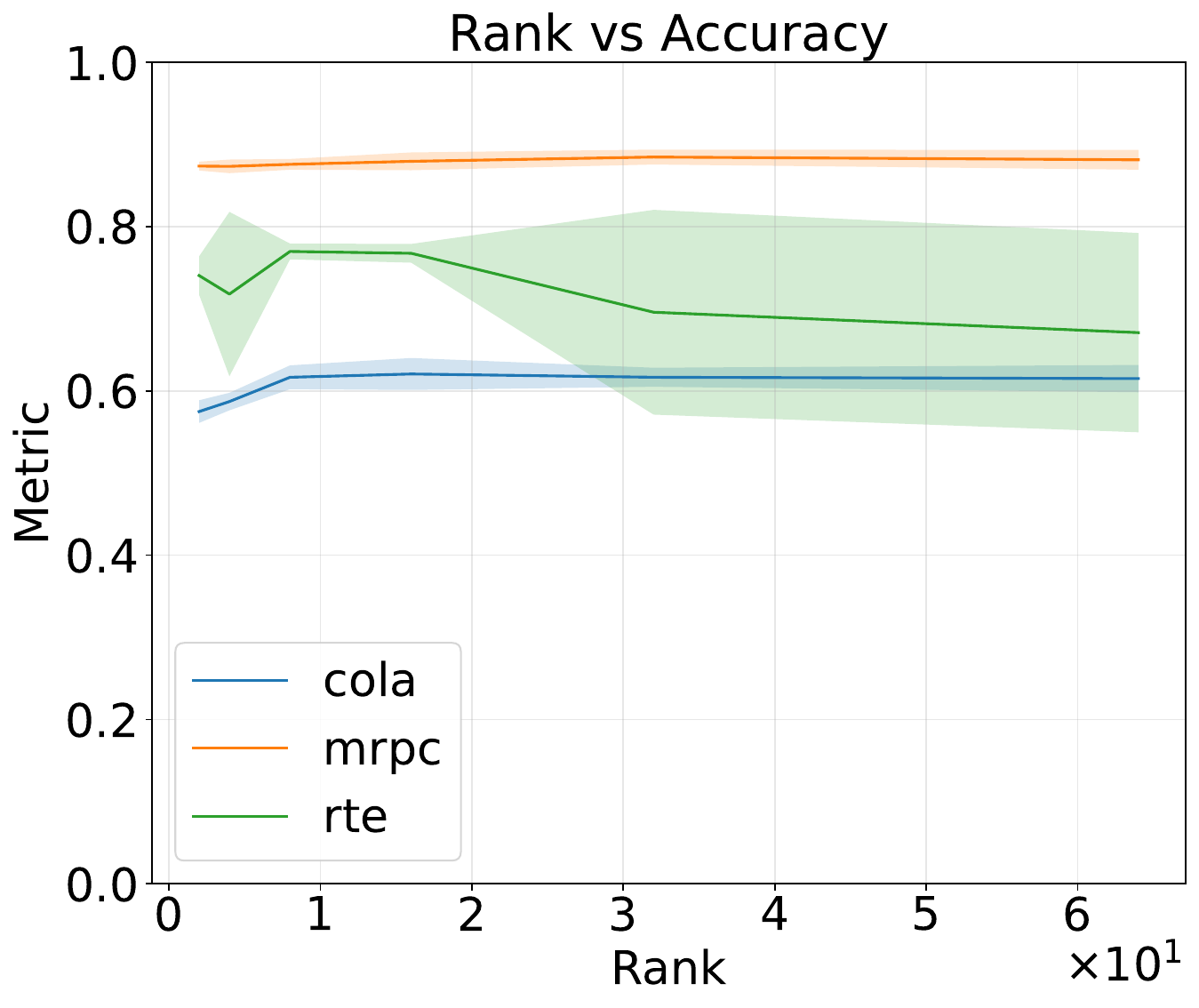}
        \caption{\rbase with LoRA.}
    \end{subfigure}
    \begin{subfigure}[b]{0.4\linewidth}
        \centering
        \includegraphics[width=\linewidth]{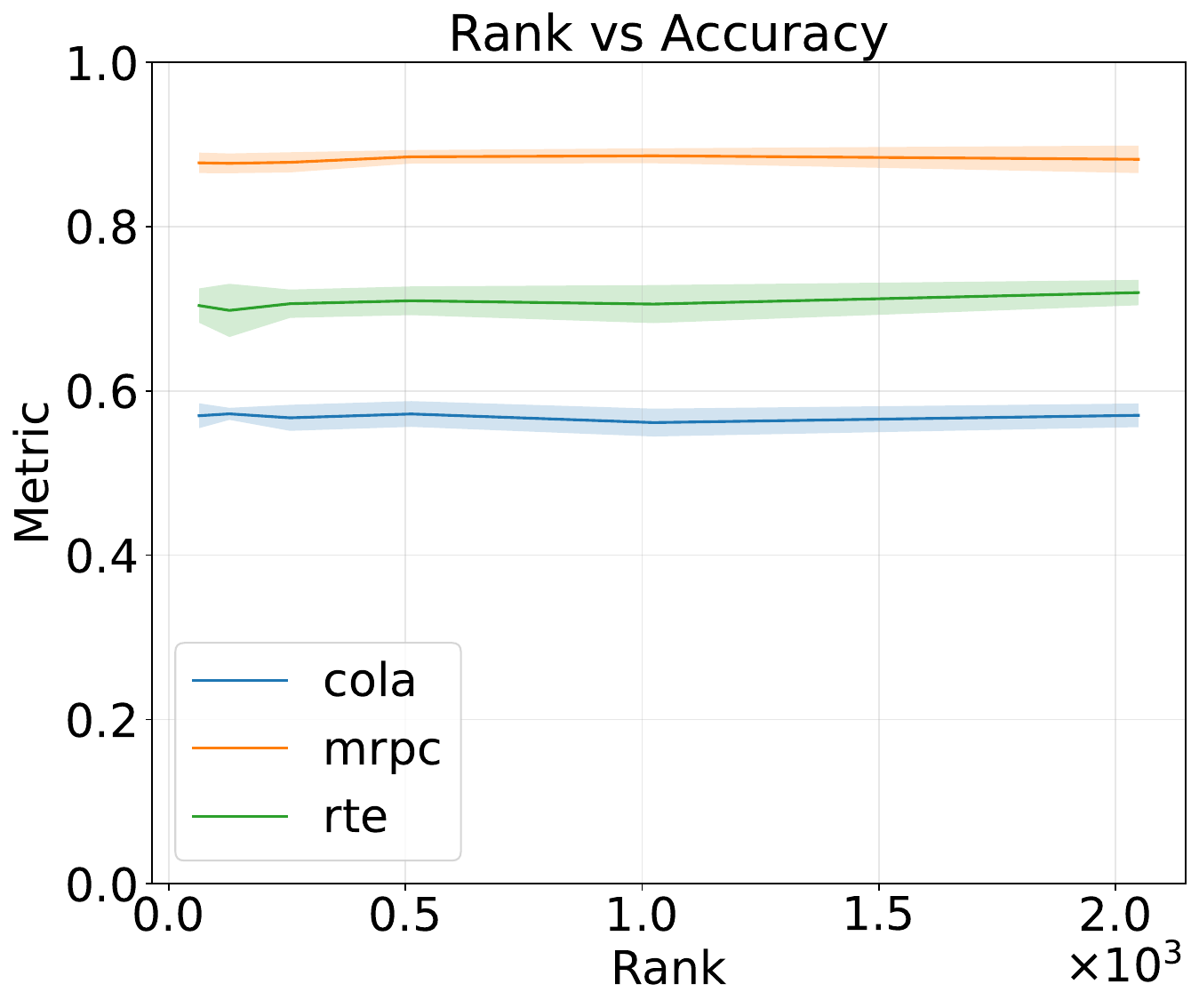}
        \caption{\rbase with VeRA.}
    \end{subfigure}
    \begin{subfigure}[b]{0.4\linewidth}
        \centering
        \includegraphics[width=\linewidth]{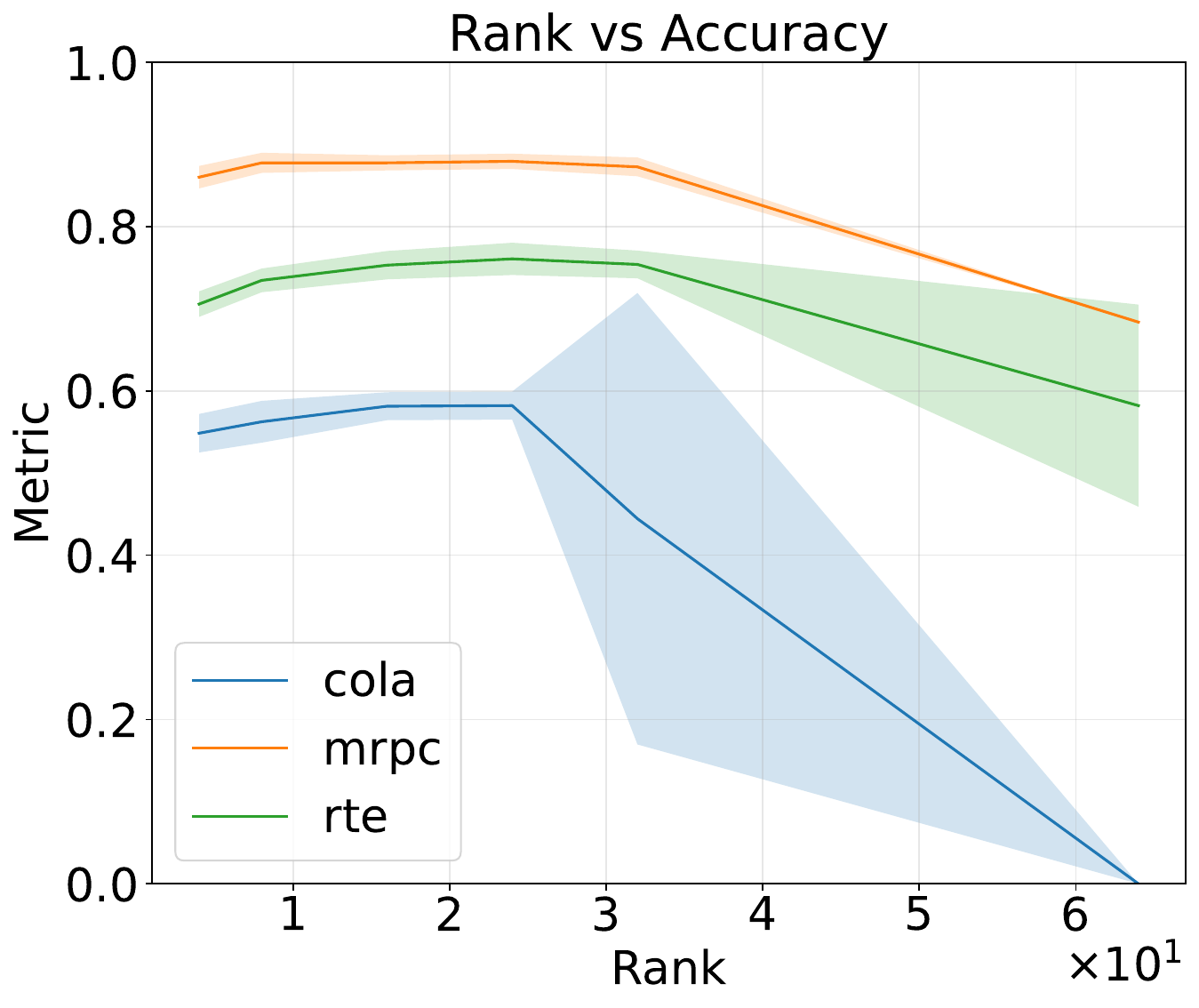}
        \caption{\rbase with MetaTT-4D.}
    \end{subfigure}
    \begin{subfigure}[b]{0.4\linewidth}
        \centering
        \includegraphics[width=\linewidth]{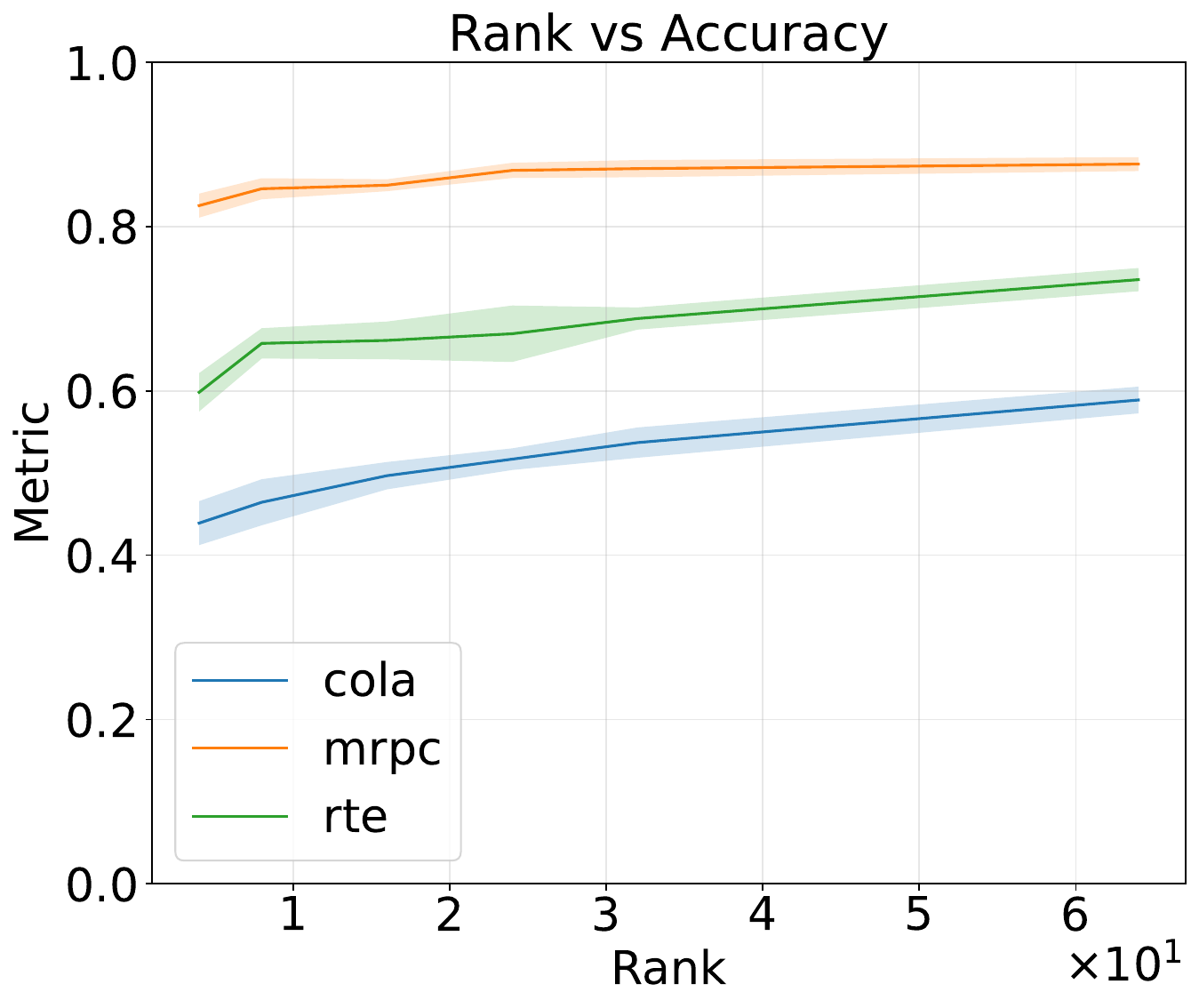}
        \caption{\rbase with MetaTT-5D.}
    \end{subfigure}
    \caption{\textbf{Rank vs accuracy.} We plot final accuracy of \rbase when trained with specific adapters on specific glue tasks when keeping other hyper-parameters fixed and varying ranks. We observe that both LoRA and VeRA maintains performance (hinting a little at model capacity). MetaTT-4D's performance gets somewhat worse at higher rank, hinting at the requirement to find the right pair of ranks and $\alpha$, while MetaTT-5D starts worse and keeps improving across ranks. 
    }
    \label{fig:rank-v-acc-base}
\end{figure}

\begin{figure}[!h]
    \centering
    \begin{subfigure}[b]{0.4\linewidth}
        \centering
        \includegraphics[width=\linewidth]{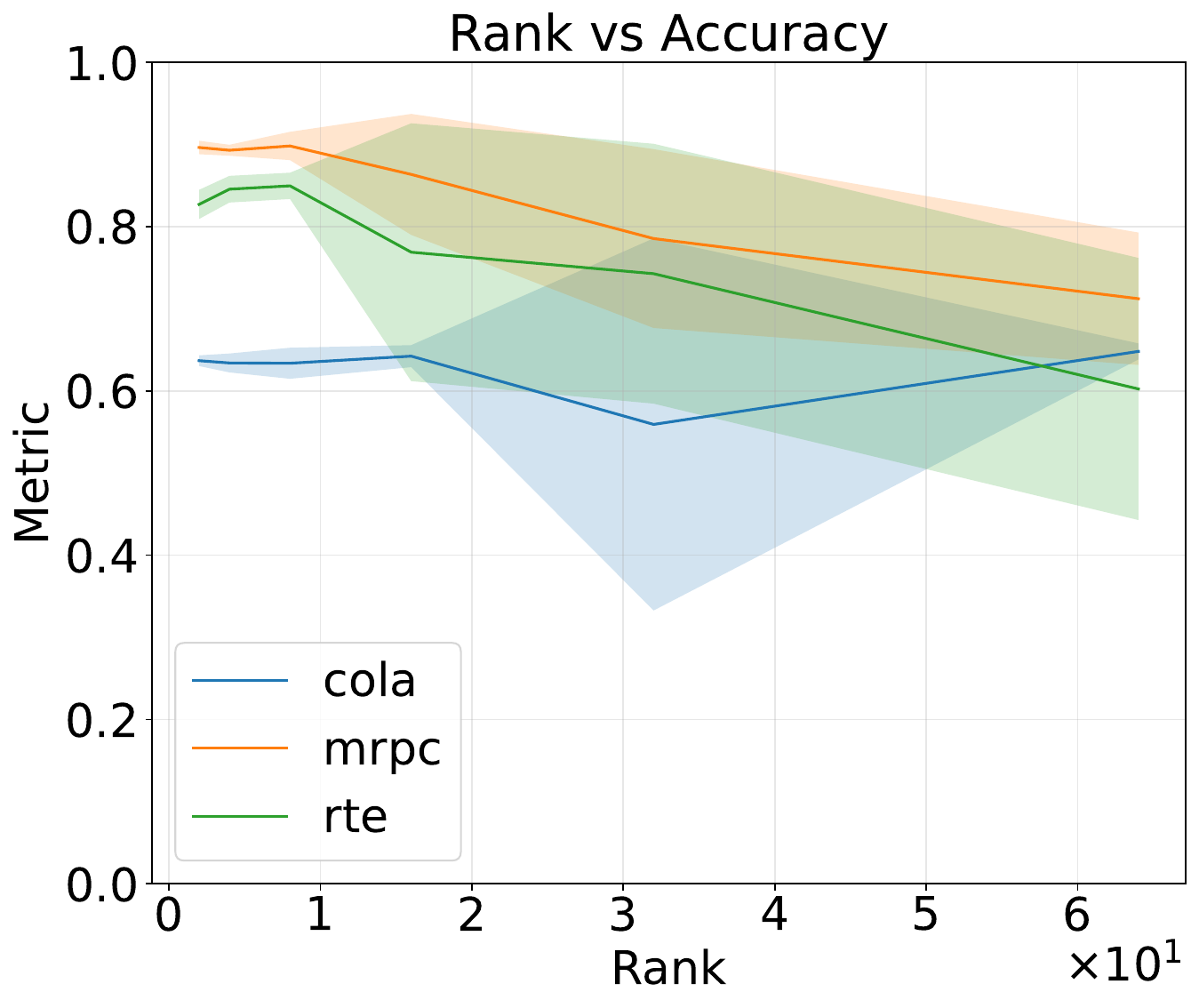}
        \caption{\rlarge with LoRA.}
    \end{subfigure}
    \begin{subfigure}[b]{0.4\linewidth}
        \centering
        \includegraphics[width=\linewidth]{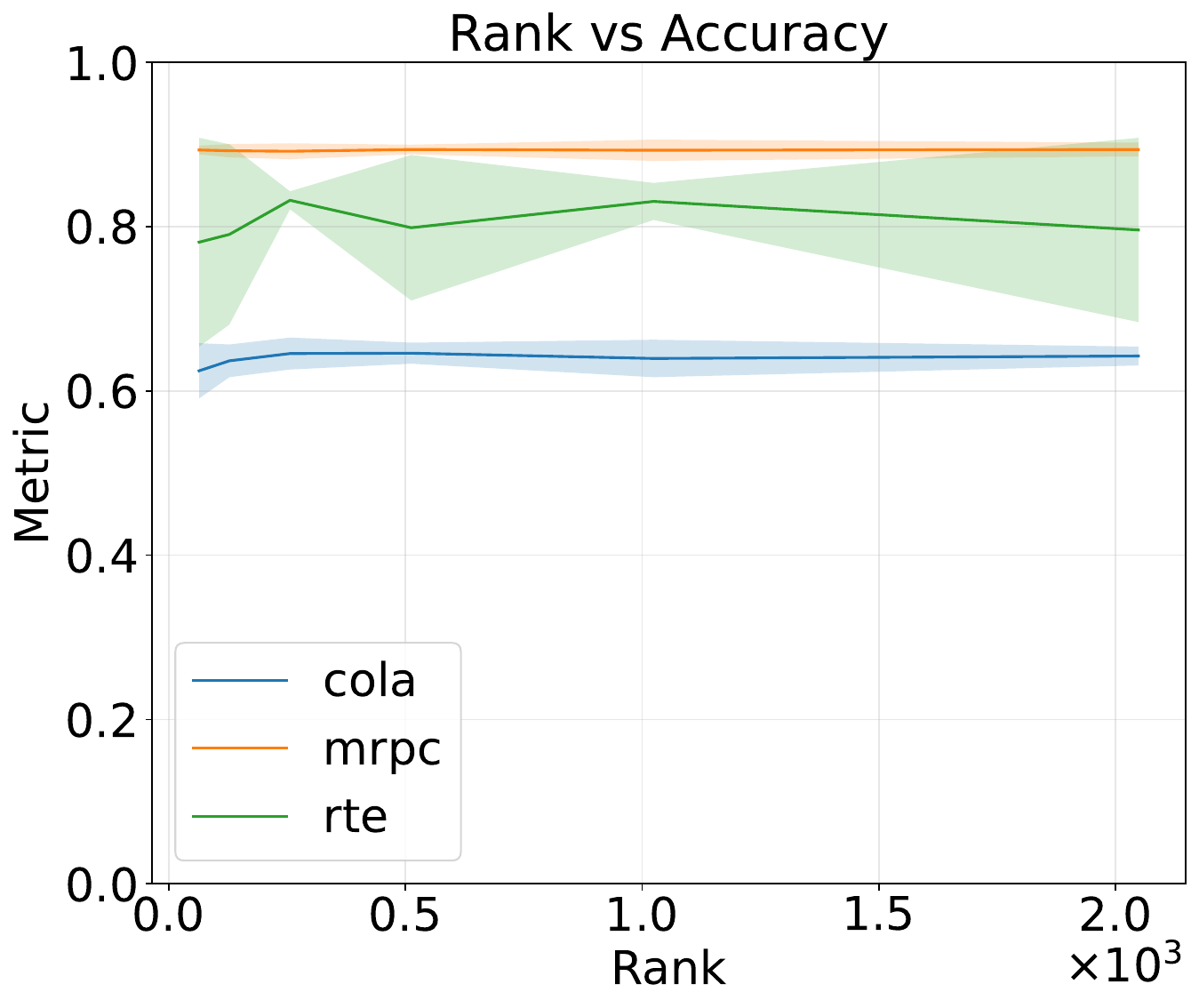}
        \caption{\rlarge with VeRA.}
    \end{subfigure}
    \begin{subfigure}[b]{0.4\linewidth}
        \centering
        \includegraphics[width=\linewidth]{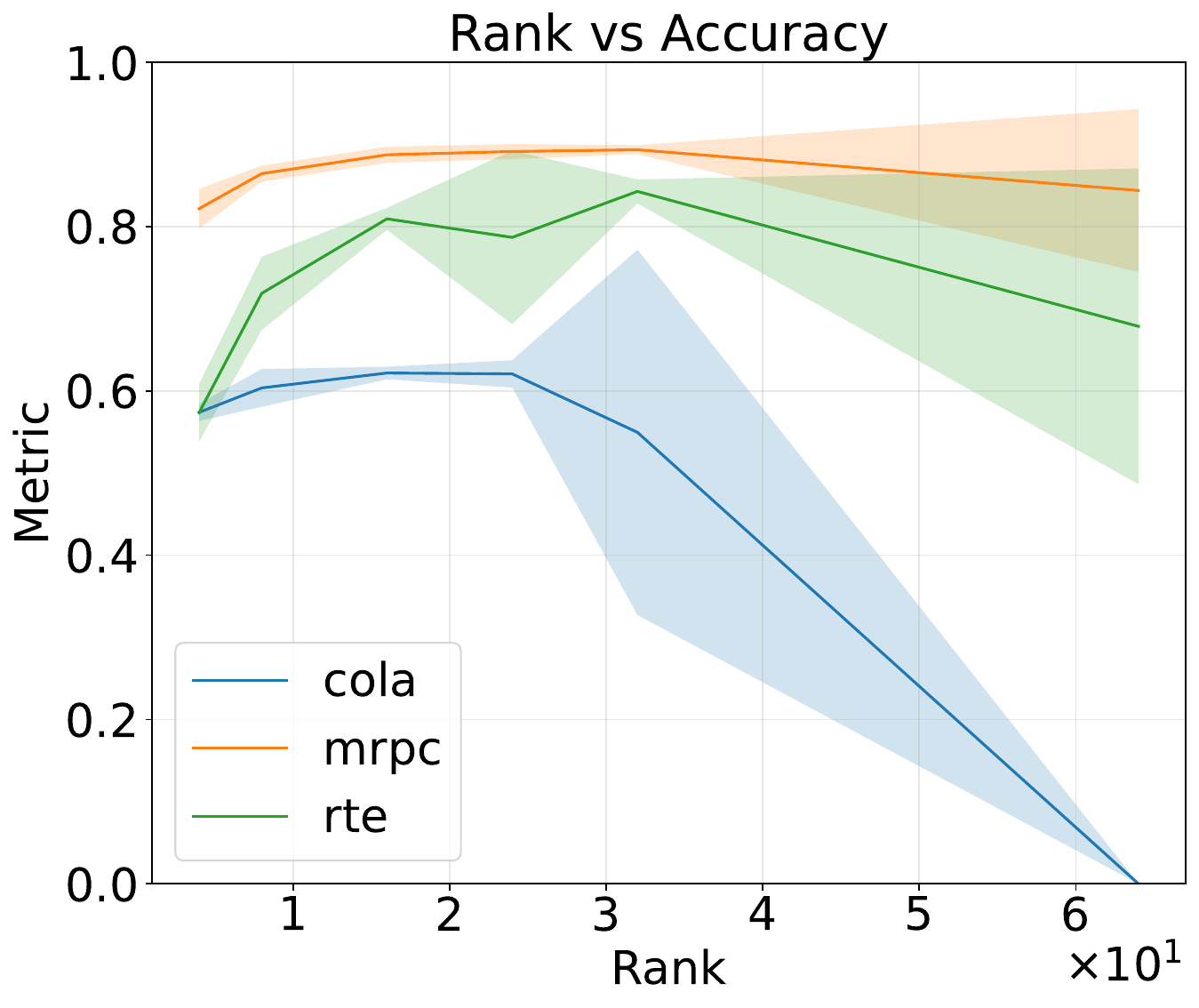}
        \caption{\rlarge with MetaTT-4D.}
    \end{subfigure}
    \begin{subfigure}[b]{0.4\linewidth}
        \centering
        \includegraphics[width=\linewidth]{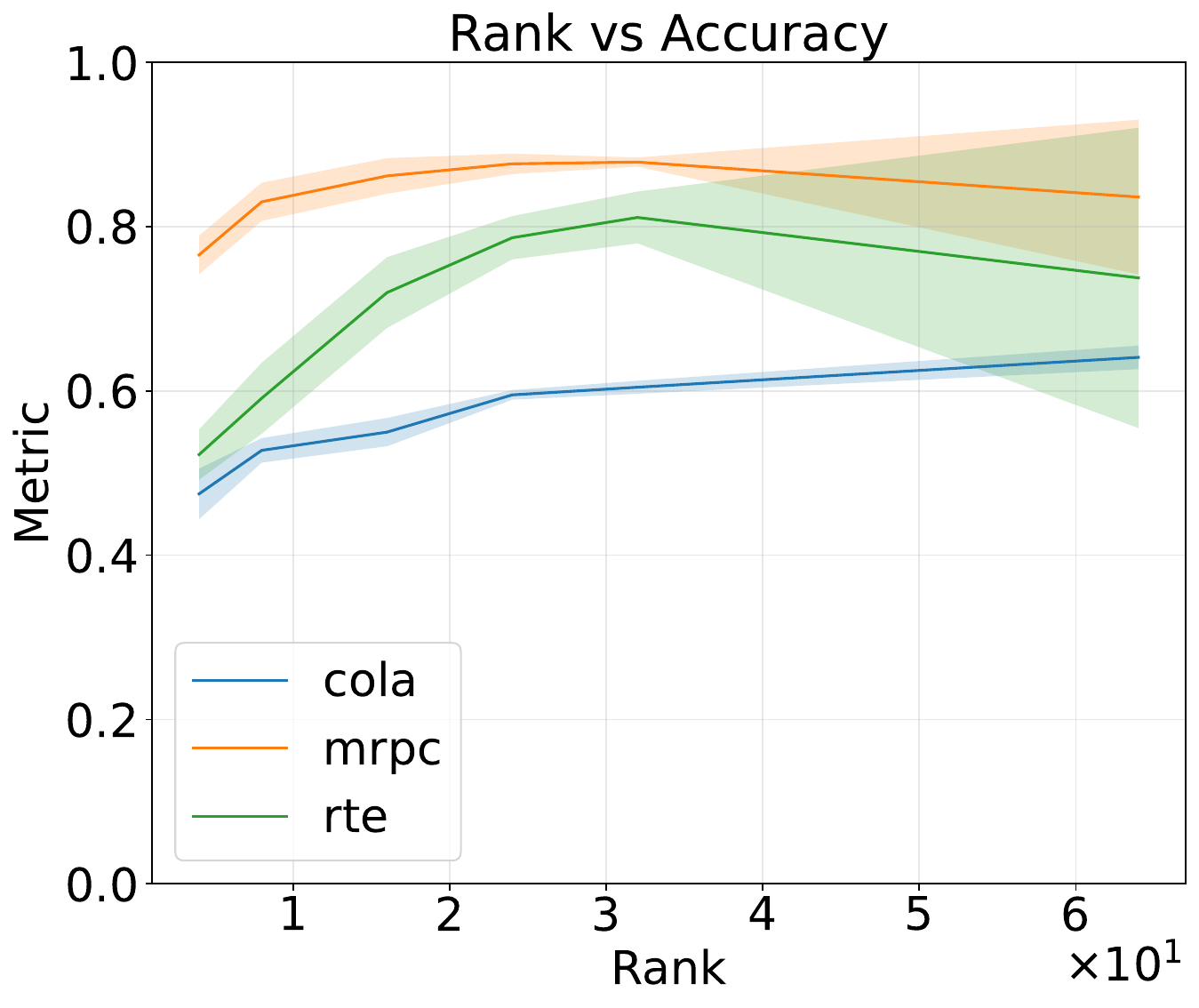}
        \caption{\rlarge with MetaTT-5D.}
    \end{subfigure}
    \caption{\textbf{Rank vs accuracy.} We plot final accuracy of \rlarge when trained with specific adapters on specific glue tasks when keeping other hyper-parameters fixed and varying ranks. We observe that similar to \cref{fig:rank-v-acc-base}, LoRA and VeRA maintains performance across ranks. MetaTT-4D gets worse on CoLA while almost maintaining performance on MRPC and RTE, while MetaTT-5D starts worse but improves as we increase ranks. 
    }
    \label{fig:rank-v-acc-large}
\end{figure}

We also plot the performance of each of the adapters when the rank is varied while other hyper-parameters are kept constant in \cref{fig:rank-v-acc-base} and \cref{fig:rank-v-acc-large} (similar to \cref{sec:stability-lr}). Again note that the parameter $\alpha$ was chosen as $2\times \operatorname{rank}$ for LoRA, and for MetaTT variants were kept as fixed as reported in \cref{app:exp-deets}. We first observe that across tasks and models sizes, LoRA and VeRA performs similarly across ranks. This is somewhat expected as we vary both rank and $\alpha$ (implicitly for VeRA and explicitly for LoRA). However, MetaTT-4D somewhat performs worse at higher ranks when both models are fine-tuned with CoLA and somewhat maintains performance on other tasks. \rev{For MetaTT-5D, performance improves with ranks across models even with fixed $\alpha$, though this variant is generally less stable than MetaTT-4D as discussed above.} 

\subsection{Different initializations of LoRA}

\begin{figure}[!h]
    \centering
    \begin{subfigure}[b]{0.48\linewidth}
        \includegraphics[width=\linewidth]{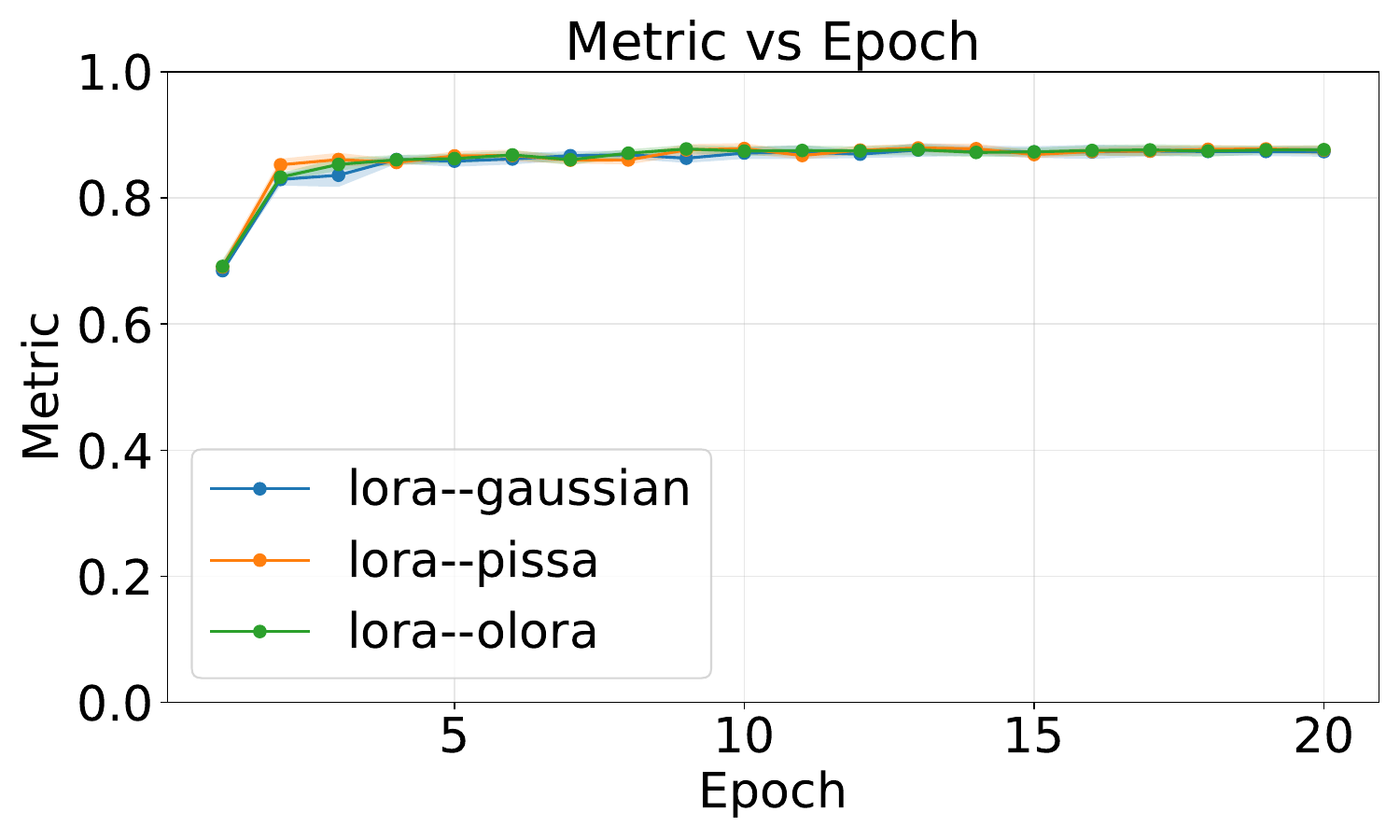}
        \caption{MRPC}
    \end{subfigure}
    \begin{subfigure}[b]{0.48\linewidth}
        \includegraphics[width=\linewidth]{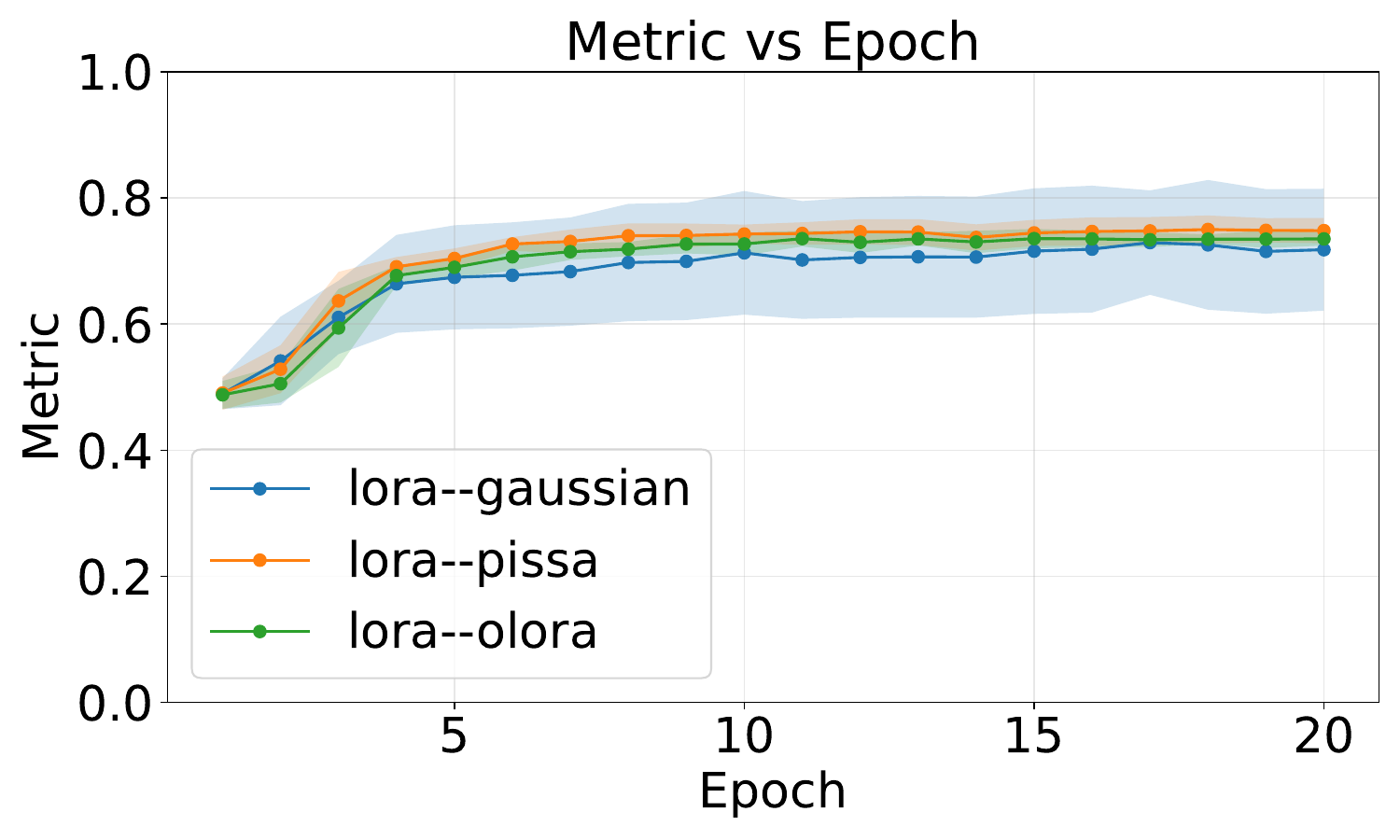}
        \caption{RTE}
    \end{subfigure}
    \caption{\textbf{LoRA using different initializations.} We plot the validation accuracy across training epochs using different initializations for LoRA on RoBERTa$_{\text{Base}}$. We observe that on average these initializations work similar to each other.}
    \label{fig:lora-init-vars}
\end{figure}

Finally, we also plot the evaluation accuracy during training  for several initializations of LoRA. Specifically, we compare the following initializations for LoRA -- 1) Gaussian, 2) Pissa, and 3) OLoRA. We run $8$ independent trials with the hyper-parameters in \cref{tab:lora-adalora-hyperparams} and finetune \rbase on MRPC and RTE tasks. We plot these results in \cref{fig:lora-init-vars} We compare the validation accuracy across training epochs similar to \cref{fig:tt_ini}. We observe that across both MRPC and RTE, LoRA performs similarly when initialized with either Gaussian, Pissa  \cite{meng2024pissa}(matrices initialized as singular vectors of the pre-trained weights), and OLoRA \cite{buyukakyuz2024olora} (base weights are translated with their QR decomposition). Although not as close the performance of these variants, we tested on different initializations of MetaTT-4D in \cref{app:exp-deets}, and observed that the models performed similarly across training epochs. However, tricks like the use of singular vectors of the pre-trained model to initialize adapter weights do not necessarily translate to the TT architecture.

\section{AdaLoRA vs. MetaTT with DMRG-inspired sweeps} \label{sec:ada_vs_dmrg_sec}
In this section we compare AdaLoRA \cite{zhang2023adalora} and the effects of using DMRG-inspired sweeps on MetaTT and comment on the cost of implementing both of these methods for the respective models analytically.

\subsection{Comparing AdaLoRA with LoRA and relating to the improvements via DMRG}
\begin{figure}[!h]
    \centering
    \begin{subfigure}[b]{0.32\linewidth}
        \centering
        \includegraphics[width=\linewidth]{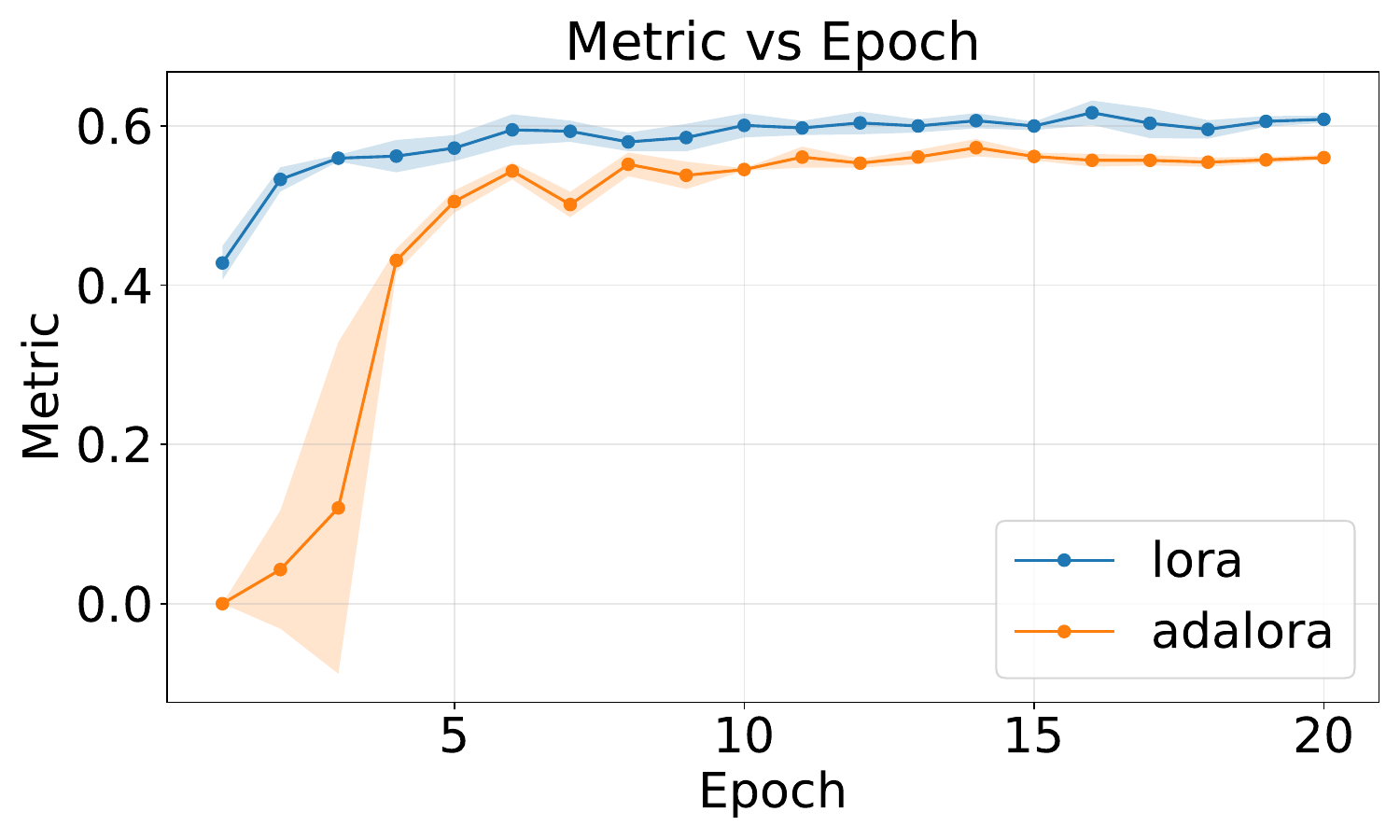}
        \caption{\rbase on CoLA}
    \end{subfigure}
    \begin{subfigure}[b]{0.32\linewidth}
        \centering
        \includegraphics[width=\linewidth]{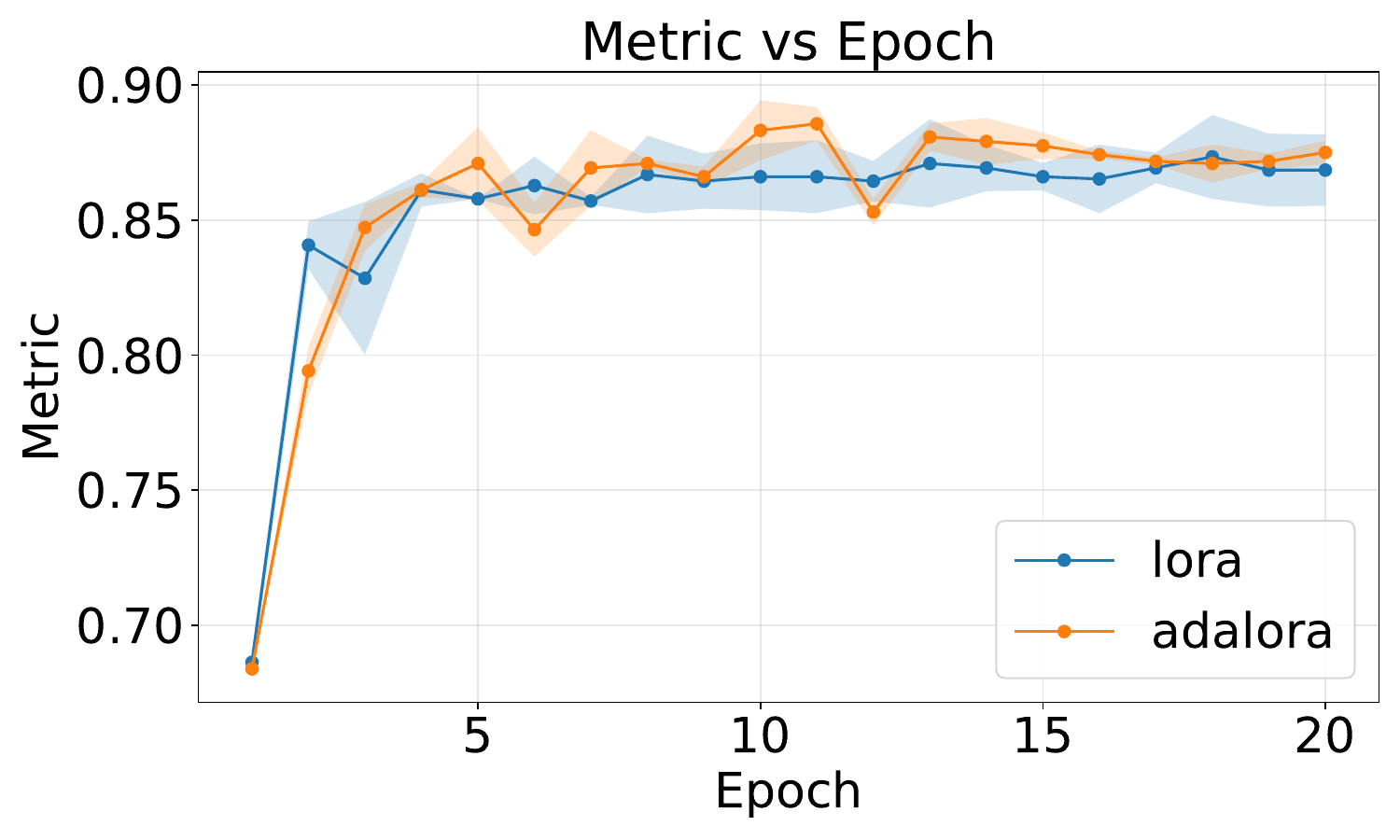}
        \caption{\rbase on MRPC}
    \end{subfigure}
    \begin{subfigure}[b]{0.32\linewidth}
        \centering
        \includegraphics[width=\linewidth]{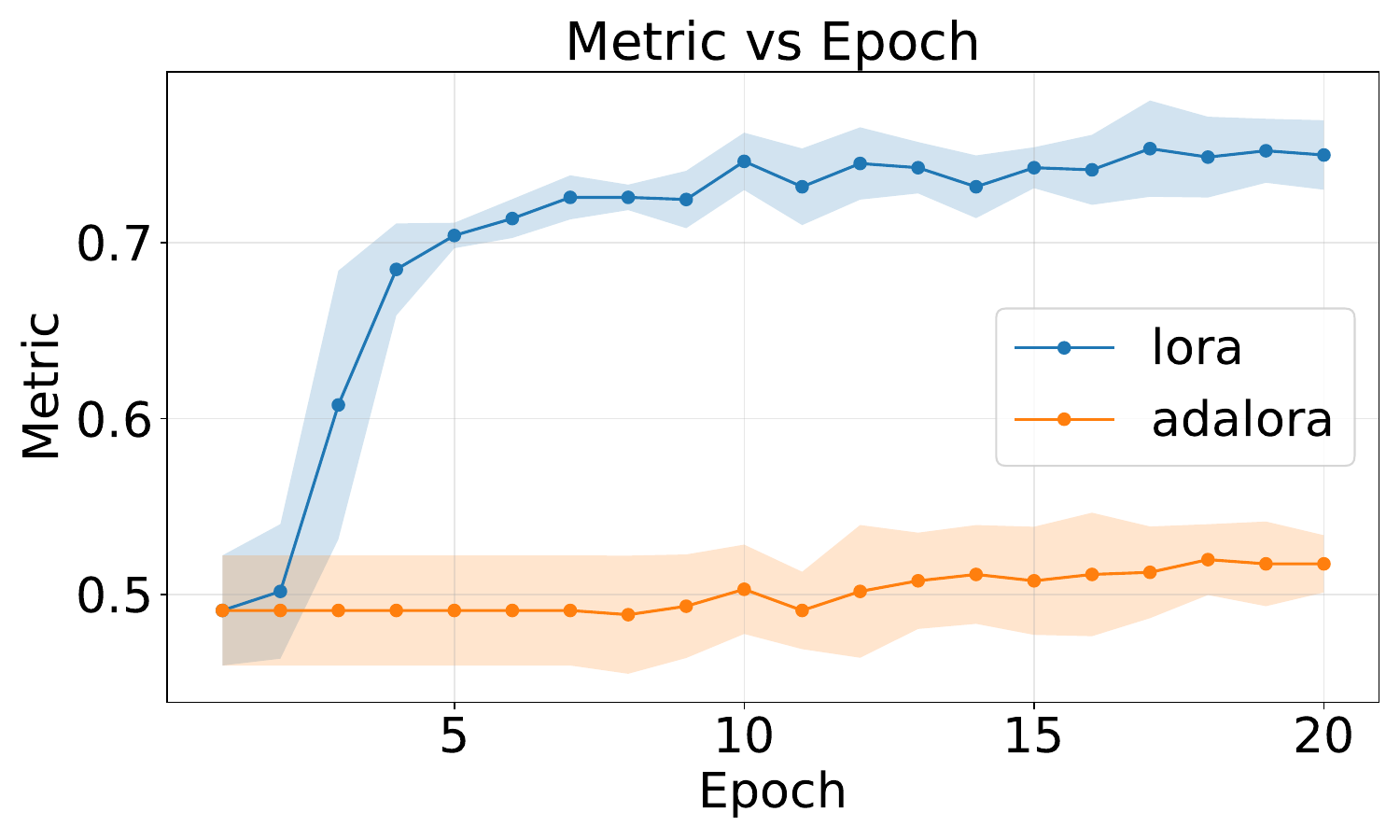}
        \caption{\rbase on RTE}
    \end{subfigure}
    \caption{\textbf{Comparison of LoRA and AdaLoRA during training.} We plot here the validation accuracy achieved using \rbase on CoLA (Matthew's correlation coefficient), MRPC (evaluation accuracy), and RTE (evaluation accuracy) when trained using LoRA with rank $4$ and AdaLoRA with initial rank $8$ and final rank $4$.}
    \label{fig:adalora-v-lora-training}
\end{figure}

In this section we compare the performance of AdaLoRA \cite{zhang2023adalora} with LoRA. We want to specifically understand the capacity of a model trained on AdaLoRA for a fixed target rank to improve upon the performance of the LoRA with same rank. To establish fair comparison to LoRA we fix a target rank of $4$ and report the results in \cref{tab:adaloravlora}. We report the mean and standard deviation of runs corresponding to seeds $[33305628, 2025, 42]$ for \rbase and $[56346, 2025, 42]$ for \rlarge. The best set of hyper-parameters found for AdaLoRA and LoRA for these experiments are reported in \cref{tab:lora-adalora-hyperparams}. In \cref{fig:adalora-v-lora-training} we also plot the validation accuracy during training \rbase using both LoRA and AdaLoRA on three GLUE tasks. As with other experiments reported in our work, we again freeze the classifier and observe that AdaLoRA fails catastrophically for RTE. Moreover, in the cases where on average it outperforms LoRA, the variance of the resulting model is often higher. The fixed settings used for AdaLoRA were -- 1) warmup steps was set at 200, number of steps for final finetuning was set at 1000, time interval between budget allocations was set at 10, hyperparameter for EMA sensitivity smoothing was 0.85 and for uncertainty quantification was 0.85 (used in the original paper), and total training steps was set at 2000. For $\alpha = 16$ and batch size $32$, we searched for best learning rates in range $[1e-4, 1e-3]$ and report the corresponding best set of hyper-parameters.

\begin{table}[h]
    \centering
    \begin{tabular}{|l|l|c|c|}
        \hline
        \textbf{Model} & \textbf{Dataset} & \textbf{LoRA} & \textbf{AdaLoRA}  \\
        \hline
        & CoLA & $\mathbf{60.8(5)}$ & $56.0(4)$\\
        \rbase & MRPC & $87(1) $ & $\mathbf{87.5(4)}$\\
        & RTE & $\mathbf{75(2)}$ & $52(2)$\\
        \hline
        & CoLA & $63.6(4)$ & $\mathbf{64.4(9)}$\\
        \rlarge & MRPC & $89(1)$ & $\mathbf{90.3(4)}$\\
        & RTE & $\mathbf{85(2)}$ & $56(4)$\\
        \hline
    \end{tabular}
    \caption{\textbf{LoRA vs AdaLoRA.} We report the mean accuracy achieved when the target rank for LoRA and AdaLoRA is $4$ while fine-tuning \rbase and \rlarge in some GLUE tasks.}
    \label{tab:adaloravlora}
\end{table}
\begin{table}[h]
    \centering
    \begin{adjustbox}{max width=\textwidth}
    \begin{tabular}{|l|l|c|c|c|l|c|c|c|}
        \hline
        Model & LoRA parameters & CoLA & MRPC & RTE & AdaLoRA parameters & CoLA & MRPC & RTE\\
        \hline
        & Rank & $4$ & $4$ & $4$ & Init \& target ranks & $[8, 4]$ & $[8, 4]$ & $[8, 4]$\\
        \rbase & $\alpha$ & $8.0$ & $8.0$ & $8.0$ & $\alpha$ & $16.0$ & $16.0$ & $16.0$ \\
        & Learning rate & $5e-4$ & $2e-4$ & $2e-4$ & Learning rate & $2e-4$ & $1e-3$ & $2e-4$ \\
        & Batch & $16$ & $8$ & $16$ & Batch & $8$ & $8$ & $8$ \\
        \hline
        & Rank & $4$ & $4$ & $4$ & Init \& target ranks & $[8, 4]$ & $[8, 4]$ & $[8, 4]$ \\
        \rlarge & $\alpha$ & $8.0$ & $8.0$ &  $8.0$ & $\alpha$ & $16.0$ & $16.0$ & $16.0$ \\
        & Learning rate & $2e-4$ & $2e-4$ & $2e-4$ & Learning rate & $2e-4$ & $4e-4$ & $2e-4$ \\
        & Batch & $16$ & $8$ & $16$ & Batch & $8$ & $8$ & $16$ \\
        \hline
    \end{tabular}
    \end{adjustbox}
    \caption{\textbf{LoRA and AdaLoRA hyper-parameters used for fine-tuning.} Here we report the best set of hyper-parameters for LoRA and AdaLoRA found after performing hyper-parameter optimization when target rank for both methods is $4$.}
    \label{tab:lora-adalora-hyperparams}
\end{table}

\subsection{Comparison of AdaLoRA vs. DMRG} \label{sec:adalora_vs_dmrg}

\begin{figure}[!h]
    \centering
    \includegraphics[width=0.6\linewidth]{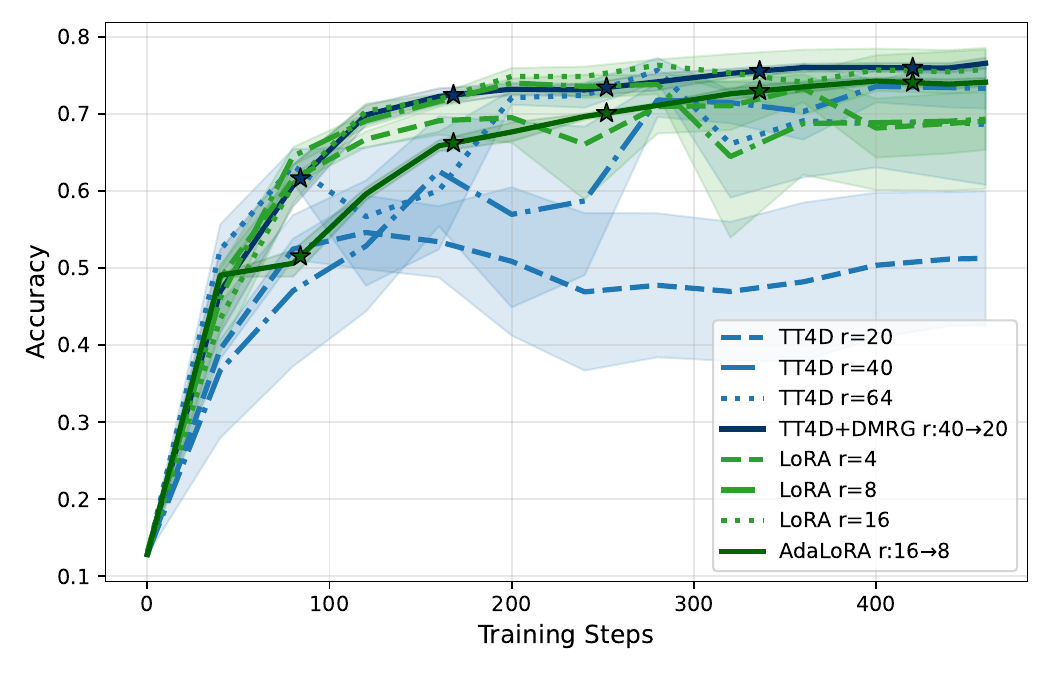}
    \caption{\textbf{Comparison of LoRA and AdaLoRA \textit{vs.} MetaTT-4D and MetaTT-4D with DMRG-style sweeps.} Fine-tuning is performed on Commonsense15k (downsampled from Commonsense170k) with Llama-2-7b as the base model, for one epoch. Results show means and relative errors over 5 independent runs for each method. Star symbols indicate steps at which DMRG/AdaLoRA updates are applied. For MetaTT, $\alpha=1.0$ (fixed); for LoRA, $\alpha=2r$; for AdaLoRA, $\alpha=32$ (twice the initial rank). Hyper-parameter tuning was performed over learning rates $[2\mathrm{e}{-4}, 5\mathrm{e}{-4}]$ for all adapters. See main text for details on the rank schedules used for MetaTT+DMRG and AdaLoRA.}
    \label{fig:adalora_vs_dmrg}
\end{figure}

In \cref{fig:adalora_vs_dmrg}, we compare MetaTT-4D adapters (with and without DMRG-inspired sweeps) against LoRA and AdaLoRA adapters. Experiments are conducted on Commonsense15k, a downsampled version of Commonsense170k from \cite{hu2023llm}, using Llama-2-7b as the base model.

For MetaTT, we employ the following rank schedule $r(i) = r_f + (r_0 - r_f)\left[1 - (i/N)^\gamma\right]$, where $N=5$, $i=1,\ldots,5$, $\gamma=2$, initial rank $r_0=40$, and final rank $r_f=20$. This schedule is motivated by empirical observations that larger rank reductions are more effective at later training steps, after the initial rapid learning phase. This approach closely resembles the cubic schedule used by AdaLoRA \cite{zhang2023adalora}.

We implement AdaLoRA using its HuggingFace implementation with the following parameters:         \texttt{target\_r}=8, \texttt{init\_r}=16, \texttt{tinit}=84, \texttt{tfinal}=39, \texttt{deltaT}=84, \texttt{beta1}=0.85, \texttt{beta2}=0.85, \texttt{orth\_reg\_weight}=0.1.  

Our results show that MetaTT-4D with DMRG-inspired sweeps not only outperforms the counterpart trained simply via AdamW, for a target rank of $r=20$, but also other variants with larger ranks, and more importantly LoRA and AdaLoRA adapters. We find that MetaTT-4D achieves accuracies comparable with LoRA with rank $r=16$, but with $\approx 47$x fewer parameters.

In \cref{tab:adalora_vs_dmrg} we compare times taken for each of the runs on a single A100 GPU. We observe a slight overhead of running MetaTT with DMRG-inspired sweeps. This is mostly a result of a temporary increase in evaluation time immediately following each DMRG update. This artifact arises from PyTorch's need to recompile the computational graph and reinitialize CUDA kernels after the adapter tensor shapes are modified, rather than from the algorithmic complexity of the DMRG procedure itself. This motivates to apply fewer such DMRG steps (3-5 in our experiments). 
\begin{table}[h]
    \centering
    \begin{adjustbox}{max width=\textwidth}
    \begin{tabular}{|c|c|c|c|c|}
    \hline
         & LoRA ($r=16$) & AdaLoRA ($r:16\to 8$) & MetaTT-4D ($r=40$) & MetaTT-4D + DMRG ($r:40\to 20$) \\
    \hline
        Time (s)  & 1725  & 1760  & 1680  & 1920 \\
    \hline
    \end{tabular}
    \end{adjustbox}
    \caption{\textbf{Training times for LoRA and MetaTT based adapters.} End-to-end training times for a single realization used in Fig.~\ref{fig:adalora_vs_dmrg}.}
    \label{tab:adalora_vs_dmrg}
\end{table}

\subsection{Complexity Analysis of SVD-based Rank Adaptation: LoRA vs. MetaTT} \label{sec:compl_ada_vs_dmrg}
A key motivation behind the development of AdaLoRA is the computational expense associated with performing SVDs on all weight matrices to dynamically truncate LoRA ranks during training. This process quickly becomes prohibitive for large models, as the cost of SVD scales cubically with the hidden dimension. To mitigate this, AdaLoRA instead employs a pruning strategy, zeroing out entries deemed irrelevant according to a score that serves as a proxy for singular values.
In contrast, our DMRG-inspired algorithm for MetaTT adapters enables SVD-controlled truncations to be performed efficiently. The TT structure of MetaTT allows for global compression and facilitates rank adaptation via SVD sweeps over a much smaller set of tensor cores, rather than all individual weight matrices. This approach not only reduces computational overhead but also allows for dynamic reduction of matrix sizes during training, which is better exploited by GPUs compared to the sparse matrix operations required by AdaLoRA.
To explicitly quantify the computational benefits of our rank-adaptive scheme over the alternative of performing SVDs on all weight matrices, we present a complexity analysis. The cost of performing a single series of SVDs for LoRA adapters is given by
\begin{equation} \label{eq:loravd} O(LMD^3), \quad \text{(LoRA-SVD)} \end{equation}
where $L$ is the number of layers, $M$ is the number of matrices adapted per layer, and $D$ is the hidden dimension. In contrast, a single DMRG-style sweep (\cref{alg:DMRG}) with initial TT-rank $r$ incurs a cost for MetaTT-4D of
\begin{equation} \label{eq:metattsvd} O(2DLr\min(D,Lr)) + O(2LMr^2\min(Lr,Mr)) + O(2DMr\min(D,Mr)). \quad \text{(MetaTT-SVD)} \end{equation}
Assuming constant factors of O(1) in (\ref{eq:loravd})–(\ref{eq:metattsvd}), the ratio of the LoRA-SVD cost to the MetaTT-SVD cost for the models considered in this work is summarized in \cref{tab:lora_vs_metatt_ratios} for various TT-ranks $r$
\begin{table}[h]
    \centering
    \begin{adjustbox}{max width=\textwidth}
    \begin{tabular}{|c|c|c|c|c|}
    \hline
        TT-rank $r$ & RoBERTa$_\text{Base}$ & RoBERTa$_\text{Large}$ & Llama-2-7b & Llama-2-13b \\
    \hline
        16  & 186  & 169  & 2039  & 2553  \\
        64  & 11 & 16 & 127 & 159 \\
        256 & 2 & 3 & 16 & 20\\
    \hline
    \end{tabular}
    \end{adjustbox}
    \caption{\textbf{Complexity of LoRA-SVD vs. MetaTT-SVD.} We show the ratio of LoRA-SVD to MetaTT-SVD complexity as the ratio of Eqs.~(\ref{eq:loravd}) to (\ref{eq:metattsvd}) with an O(1) constant, for various TT-ranks $r$ and models.}
    \label{tab:lora_vs_metatt_ratios}
\end{table}

As shown in \cref{tab:lora_vs_metatt_ratios}, the computational savings achieved by MetaTT are substantial, especially for moderate TT-ranks. For example, at $r=16$, the MetaTT approach is over two orders of magnitude more efficient than LoRA-SVD for all models considered. This efficiency gain enables practical rank-adaptive training via SVD sweeps, which would otherwise be a computational overhead for large-scale models using conventional LoRA adapters.

\section{Computational comparisons}

In this section, we empirically demonstrate that MetaTT variants are extremely competitive computationally. These results further support our claim that MetaTT adapters do not impose significant computational overhead on the base model's computational graph.

\subsection{Runtime comparisons}\label{app:adapter-runtime-comps}
In \cref{tab:runtime-comparison} we compare the world clock runtimes of LoRA, VeRA, MetaTT-4D and MetaTT-5D for both forward pass and one gradient step in \rbase and \rlarge when fine-tuning on MRPC. For this we use a $1$ A100 GPU node, which has $55$ Intel Xeon Platinum CPUs, and $500$ GBs RAM. For each adapter, we choose the best set o hyper-parameters, as these are the setting in which the respective adapters will be used in practice. 

We take a random sample of 5 batches and make forward and backward passes. We do this $5$ times and report the mean and standard deviation. For \rbase and \rlarge, we observe that LoRA and MetaTT-4D are the two of the fastest adapters in real world clock time. Note, the ranks chosen here are the ranks that were used to report results in \cref{tab:roberta-combined}.
\begin{table}[h]
    \centering
    \begin{tabular}{|l|l|c|c|c|c|}
        \hline
        \textbf{Model} & \textbf{Adapter} & \textbf{Rank} & \textbf{Batch} &\textbf{Forward pass (secs)} & \textbf{Backward pass (secs)}  \\
        \hline 
         & LoRA & 8 & 64 & $\mathbf{0.1859 (1)}$ & $\mathbf{0.2014(1)}$\\
         \rbase & VeRA & 1024 & 64 & 0.2539(0) & 0.2718(1)\\
         & MetaTT-4D & $24 \times3$ & 64 & $\mathbf{0.1866(1)}$ & $\mathbf{0.2031(0)}$ \\
         & MetaTT-5D & $64\times 4$ & 64 & 0.1921(1) & 0.2146(1)\\
        \hline
         & LoRA & 8 & 64 & $\mathbf{0.62(2)}$ & $\mathbf{0.6575(1)}$\\
         \rlarge & VeRA & 256 & 64 & 0.6465(2) & 0.7035(3) \\
         & MetaTT-4D & $32\times 3$ & 64 & $\mathbf{0.6087(0)}$ & $\mathbf{0.6612(1)}$\\
         & MetaTT-5D & $64 \times 4$ & 64 & 0.63(1) & 0.6917(2)\\
        \hline
    \end{tabular}
    \caption{\textbf{World clock runtime comparison of different PEFT adapters.} We report the average time required to make 5 batches pass through the model and corresponding adapter and then compute gradients, $5$ independent trials. We observe that in general LoRA and MetaTT-4D are the fastest among the other methods reported here for their respective best performing ranks. We note that even for similar number of trainable parameters in VeRA, the matrix-vector-vector-matrix operation is significantly larger than the matrix-matrix operations in LoRA and variants of MetaTT, leading to gains in the forward and backward pass across batch and trials. }
    \label{tab:runtime-comparison}
\end{table}

\subsection{Memory overhead}

Beyond runtimes, we also profile the memory requirement of various adapters when fine-tuning \rbase and \rlarge on MRPC and report them in \cref{tab:memory-overhead}. Since we already understand the speed of the adapters when the best set of hyper-parameters are chosen from \cref{app:adapter-runtime-comps}, we now profile these adapters when target rank is fixed to either 16 or 64. We train the models for 1 epoch, and log the memory requirement every 10 steps into fine-tuning. Throughout we fix the batch size at 64.

We observe that across ranks and model sizes for RoBERTa, LoRA and MetaTT-4D consistently require lower memory (both GPU reserved and GPU allocated) than the other adapters. This coupled with the fact that for a fixed rank, MetaTT variants can learn a higher order function at any given rank, demonstrates that variants of MetaTT can be used without incurring much computational overhead when compared to the state-of-the-art PEFT adapters.
\begin{table}[h]
    \centering
    \begin{adjustbox}{max width=\textwidth}
    \begin{tabular}{|l|l|c|c|c|c|}
    \hline
    Model & Adapter & Rank & Batch & GPU allocated (MB) & GPU reserved (MB)\\
    \hline
    & LoRA & 16 & 64 & \textbf{7466.54 (635.93)} & \textbf{7773.93 (0.53)} \\
    \rbase & VeRA & 16 & 64 & 8088.68 (693.07) & 8416.00 (0.00)\\
    & MetaTT-4D & 16 & 64 & \textbf{7505.36 (640.28)} & \textbf{7816.00 (0.00)} \\
    & MetaTT-5D & 16 & 64 & 7789.85 (666.28) & 8104.00 (0.00) \\
    \hline
    & LoRA & 64 & 64 & \textbf{7557.95 (640.52)} & \textbf{7838.00 (0.00)} \\
    \rbase & VeRA & 64 & 64 & 8160.37 (697.59) & 8500.00 (0.00)\\
    & MetaTT-4D & 64 & 64 & \textbf{7720.70 (656.26)} & \textbf{7997.97 (0.26)}\\
    & MetaTT-5D & 64 & 64 & 8859.09 (760.26) & 9149.97 (0.26) \\
    \hline
    & LoRA & 16 & 64 & \textbf{19431.54 (1646.91)} & \textbf{19897.79 (1.58)} \\
    \rlarge & VeRA & 16 & 64 & 21011.46 (1791.78) & 21513.97 (0.26) \\
    & MetaTT-4D & 16 & 64 & \textbf{19505.04 (1655.60)} & \textbf{19976.00 (0.00)}\\
    & MetaTT-5D & 16 & 64 & 20263.81 (1724.93) & 20744.00 (0.00)\\
    \hline
    & LoRA & 64 & 64 & \textbf{19627.44 (1658.08)} & \textbf{20098.03 (0.26)} \\
    \rlarge & VeRA & 64 & 64 & 21154.22 (1804.05) & 21673.97 (0.26)\\
    & MetaTT-4D & 64 & 64 & \textbf{19934.48 (1692.18)} & \textbf{20411.97 (0.26)} \\
    & MetaTT-5D & 64 & 64 & 22970.08 (1969.73) & 23483.97 (0.26) \\
    \hline
    \end{tabular}
    \end{adjustbox}
    \caption{\textbf{Comparison of memory requirements across adapters.} We provide the GPU memory requirement for some PEFT adapters when rank is fixed at either 16 or 64, batch size is 64, while fine-tuning \rbase and \rlarge on MRPC. The numbers reported here are mean and standard deviation over 1 epoch computed ever 10 steps. The lowest two in each category are shown in bold.}
    \label{tab:memory-overhead}
\end{table}

\section{Experimental Details}\label{app:exp-deets}
In this section we include experimental details not covered in the previous sections. 

\subsection{Methodology for Hyper-parameter Search}
During hyper-parameter tuning, we conducted a manual grid search without fixing random seeds, as the goal was to identify promising regions in the search space rather than produce final reportable results. The details of the grid search are given in \cref{app:exp-deets}. For the final evaluation, we selected the best-performing configurations and ran them across three different, fixed random seeds (see \cref{app:exp-deets} for the final set of hyper-parameters used) to ensure stability and reproducibility. This allowed us to balance exploration efficiency with reliable performance reporting. Generally, we follow \cite{bershatsky2024lotr} for hyper-parameter tuning. For very large datasets ($\geq 500$k data-points, e.g., MNLI and QQP), we do the hyper-parameter tuning for only $1$ epoch (for example MNLI which has $\approx 390$k entries). For smaller datasets (e.g., CoLA and MRPC), we train for $20$ epochs.

\paragraph{Seeds.} We run $3$ trials for most of our experiments unless the datasets are huge, in which case we run only $2$ trials. All experiments with \rbase use the following seeds $\{33305628, 2025, 42\}$, and the experiments with \rlarge use the following seeds $\{56346, 2025, 42\}$.

\subsection{Choice of Projection Matrices}
All experimental results in the main text were obtained by adapting $Q,V$ matrices, as these were the ones used in LoTR \cite{bershatsky2024lotr}, LoRA \cite[Tables 2, 3]{hu2021lora}, and VeRA \cite{kopiczko2023vera}. Just like several other PEFT adapters, MetaTT allows for fine-tuning any arbitrary subset of attention and projection matrices in a transformer architecture, including the MLP matrices (upon a proper reshaping). Since the number of projection matrices to be adapted $M$ (per layer) factorizes separately from other variables in MetaTT (including number of layers and input/output dimensions), higher compression rates can be achieved by considering this quantity larger. In line with previous works, we found that capturing $Q,K,V$ matrices at once did not improve over capturing only $Q,V$ matrices. We leave for future work a detailed study of the role of MLP layers and output projection matrices $O$.

\subsection{Implementation Environment}
\paragraph{Implementation details for MetaTT variants and other baselines.} To construct our training and benchmarking suite, we employed a range of technologies. HuggingFace provides a wrapper, known as HuggingFace Transformers \cite{wolf2019huggingface}, which extends existing deep learning libraries like PyTorch \cite{paszke2019pytorch} with additional NLP functionalities. This library offers a unified interface for tasks such as input tokenization, model configuration, inference pipelines, and output decoding. We utilized HuggingFace's Transformers and PEFT \cite{peft} to facilitate the design and training of our adapters, specifically taking advantage of the \textit{Trainer} and \textit{TrainingArguments} features available within the library.

\paragraph{Implementation details for DMRG-inspired sweep.} Similar to the single task and multi-task learning, we leverage HuggingFace's transformers library \cite{wolf-etal-2020-transformers} to load the models and HuggingFace datasets \cite{lhoest21} to load the datasets. However, we do not leverage the \textit{Trainer} here and instead fall back to custom PyTorch training loops as we wish to have more precise control over the training loop (this is because we are changing the model itself during the run). Doing this using a custom PyTorch loop is much cleaner than using \textit{TrainerCallbacks}.

\paragraph{Machine configuration and coding environment.} We run our benchmarks on a machine with the following configuration: dual Intel Xeon Platinum 8275CL CPUs with 96 cores, 192 threads, and 1.1 TB of RAM and 8 A100 GPUs with 40GB memory each (an AWS P5 instance). At any given point on any GPU, only 1 model is being trained against one dataset.

\subsection{MetaTT}\label{app:meta-tt-exp-deets}

\begin{figure}[h!]
    \centering
    \includegraphics[trim={0.3cm 0cm 0cm 0.3cm}, clip, width=1\linewidth]{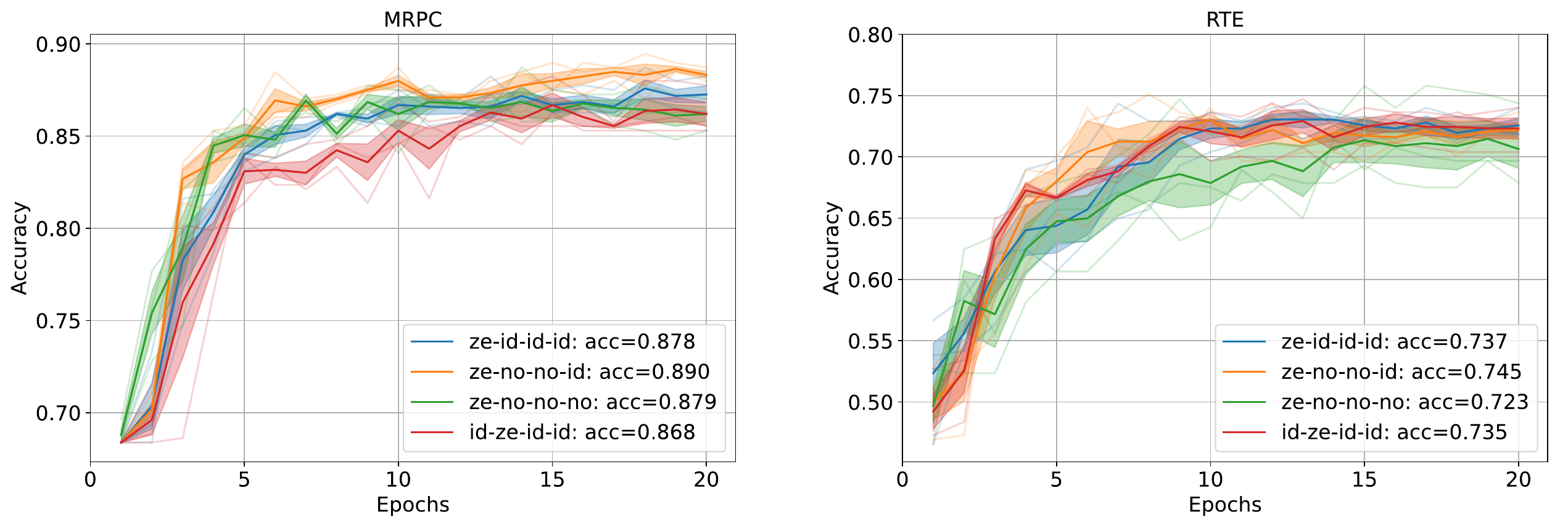}
    \vspace{-1em}
    \caption{\textbf{TT initialization performance.} Shown are the accuracies in MRPC (left) and RTE (right) when training MetaTT-4D on \rbase with different initialization strategies along with mean of best accuracies over $20$ epochs across $3$ different trials shown in the legend. Each pair of letters correspond to a different initialization strategy: `\texttt{ze}' sets a given core to zero, `\texttt{id}' sets each matrix slice of a core to the identity matrix and `\texttt{no}' to a normal distribution with $\text{mean} = 0$ and $\text{standard deviation} = 0.2$. The order of pairs of letters follows the order of how each of the cores are initialized in MetaTT-4D. We choose the sequence \texttt{ze-id-id-id} (blue line) since it generally performs well on average across multiple datasets.}
    \label{fig:tt_ini}
\end{figure}

\paragraph{Initialization.} An important component for running MetaTT successfully is the initialization strategy. There is freedom in choosing how to initialize each core, as long as the TT contraction $\mathcal{G}[i_1,\cdots, i_d]=0$ along each slice. This is required to guarantee $\Delta W_{l,m}=0$ everywhere at the beginning \cite{hu2021lora}. All the RoBERTa experiments in this work initialized the first core $\mathcal{G}_1$ to zero, and the rest to the identity along each slice. I.e., $\mathcal{G}_i[j]=\mathbf{1}$. \rev{For the Llama experiments we instead initialized the middle two cores as Gaussians with std=0.2 and zero mean (keeping the first core as zero, and the last as identity along the main diagonal).} This choice was done for simplicity and ease of reproducibility of results, and found to work well across all datasets that we experimented with. In \cref{fig:tt_ini}, we compare this scheme against other initialization strategies on MRPC and RTE. Note that further improvements on the numbers quoted in \cref{tab:roberta-combined} can be achieved by optimizing initialization choices, as shown in \cref{fig:tt_ini}.

\paragraph{Hyper-parameters for MetaTT results of \cref{tab:roberta-combined}.} In \cref{tab:metatt4d_stl_hyperparams} and \cref{tab:metatt5d_stl_hyperparams} we list the exhaustive set of hyper-parameters required to replicate the results in \cref{tab:roberta-combined} for MetaTT-4D and MetaTT-5D respectively.
For final evaluations, we run multiple trials across all the datasets for each transformer (for $20$ epochs). For CoLA, MRPC, RTE and STS-B, we do $3$ trials. For MLNI, QNLI, QQP and SST2, we run $2$ trials due to their large cardinality. 
\begin{table}[h]
    \centering
    \begin{adjustbox}{max width=\textwidth}
    \begin{tabular}{|c|c!{\vrule width 1.5pt}l|c|c|c|c|c|c|c|c|}
        \hline
         \textbf{Model} & \textbf{Rank} & \textbf{Params} & 
          \textbf{CoLA} & \textbf{MNLI} & \textbf{MRPC} & \textbf{QNLI} & \textbf{QQP} & \textbf{RTE} & \textbf{SST2} & \textbf{STS-B} \\
         \hline
         \multirow{9}{*}{\rotatebox{90}{RoBERTa$_\text{Base}$}}
         & \multirow{3}{*}{$4$} & $\alpha$ & $4$ & $4$ & $0.5$ & $4$ & $4$ & $4$ & $0.5$ & $4$\\
         \cline{3-11}
         &  & LR & $0.001$ & $0.001$ & $0.001$ & $0.001$ & $0.001$ & $0.001$ & $0.001$ & $0.001$\\
         \cline{3-11}
         & & Batch & $8$ & $8$ & $8$ & $8$ & $16$ & $16$ & $8$ & $8$\\
         \cline{2-11}
         & \multirow{3}{*}{$24$} & $\alpha$ & $4$ & $4$ & $4$ & $0.5$ & $0.5$ & $0.5$ & $4$ & $0.5$\\
         \cline{3-11}
         & & LR & $0.0005$ & $0.001$ & $0.0005$ & $0.001$ & $0.001$ & $0.001$ & $0.0005$ & $0.001$\\
         \cline{3-11}
         & & Batch & $8$ & $32$ & $16$ & $16$ & $32$ & $16$ & $32$ & $16$\\
         \cline{2-11}
         & \multirow{3}{*}{$64$} & $\alpha$ & $0.5$ & $0.5$ & $0.5$ & $0.5$ & $0.5$ & $0.5$ & $0.5$ & $0.5$\\
         \cline{3-11}
         & & LR & $0.001$ & $0.0005$ & $0.0005$ & $0.001$ & $0.001$ & $0.0005$ & $0.001$ & $0.0005$\\
         \cline{3-11}
         & & Batch & $32$ & $8$ & $32$ & $16$ & $32$ & $8$ & $8$ & $8$\\
         \Xhline{3\arrayrulewidth}
         \multirow{6}{*}{\rotatebox{90}{RoBERTa$_\text{Large}$}}
         & \multirow{3}{*}{$16$} & $\alpha$ & $0.5$ & $0.5$ & $0.5$ & $0.5$ & $0.5$ & $0.5$ & $4$ & $0.5$\\
         \cline{3-11}
         &  & LR & $0.001$ & $0.001$ & $0.0005$ & $0.001$ & $0.001$ & $0.0005$ & $0.001$ & $0.001$\\
         \cline{3-11}
         & & Batch & $8$ & $32$ & $32$ & $32$ & $16$ & $8$ & $16$ & $32$\\
         \cline{2-11}
         & \multirow{3}{*}{$32$} & $\alpha$ & $4$ & $0.5$ & $0.5$ & $0.5$ & $0.5$ & $0.5$ & $4$ & $0.5$\\
         \cline{3-11}
         & & LR & $0.0005$ & $0.001$ & $0.001$ & $0.001$ & $0.0005$ & $0.0005$ & $0.0005$ & $0.001$\\
         \cline{3-11}
         & & Batch & $32$ & $16$ & $32$ & $32$ & $8$ & $32$ & $32$ & $16$\\
         \hline
    \end{tabular}
    \end{adjustbox}
    \vspace{0.5em}
    \caption{\textbf{Hyper-parameters for RoBERTa for MetaTT-4D.} We list here the hyper-parameters that can be used to replicate the results for MetaTT-4D reported in \cref{tab:roberta-combined}.}
    \label{tab:metatt4d_stl_hyperparams}
\end{table}
\begin{table}[h]
    \centering
    \begin{adjustbox}{max width=\textwidth}
    \begin{tabular}{|c|c!{\vrule width 1.5pt}l|c|c|c|c|c|c|c|c|}
        \hline
         \textbf{Model} & \textbf{Rank} & \textbf{Params} & 
          \textbf{CoLA} & \textbf{MNLI} & \textbf{MRPC} & \textbf{QNLI} & \textbf{QQP} & \textbf{RTE} & \textbf{SST2} & \textbf{STS-B} \\
         \hline
         \multirow{6}{*}{\rotatebox{90}{RoBERTa$_\text{Base}$}}
         & \multirow{3}{*}{$16$} & $\alpha$ & $0.5$ & $0.5$ & $0.5$ & $0.5$ & $0.5$ & $0.5$ & $0.5$ & $0.5$\\
         \cline{3-11}
         &  & LR & $0.0005$ & $0.001$ & $0.001$ & $0.001$ & $0.001$ & $0.001$ & $0.0005$ & $0.001$\\
         \cline{3-11}
         & & Batch & $32$ & $16$ & $8$ & $8$ & $8$ & $8$ & $16$ & $8$\\
         \cline{2-11}
         & \multirow{3}{*}{$64$} & $\alpha$ & $0.5$ & $0.5$ & $0.5$ & $0.5$ & $0.5$ & $0.5$ & $0.5$ & $0.5$\\
         \cline{3-11}
         & & LR & $0.0005$ & $0.0005$ & $0.001$ & $0.001$ & $0.0005$ & $0.0005$ & $0.001$ & $0.001$\\
         \cline{3-11}
         & & Batch & $32$ & $8$ & $16$ & $16$ & $16$ & $32$ & $16$ & $8$\\
         \Xhline{3\arrayrulewidth}
         \multirow{6}{*}{\rotatebox{90}{RoBERTa$_\text{Large}$}}
         & \multirow{3}{*}{$32$} & $\alpha$ & $0.5$ & $0.5$ & $0.5$ & $0.5$ & $0.5$ & $0.5$ & $0.5$ & $0.5$\\
         \cline{3-11}
         &  & LR & $0.001$ & $0.001$ & $0.0005$ & $0.001$ & $0.001$ & $0.0005$ & $0.0005$ & $0.001$\\
         \cline{3-11}
         & & Batch & $32$ & $8$ & $32$ & $32$ & $16$ & $8$ & $8$ & $16$\\
         \cline{2-11}
         & \multirow{3}{*}{$64$} & $\alpha$ & $0.5$ & $0.5$ & $0.5$ & $0.5$ & $0.5$ & $0.5$ & $0.5$ & $0.5$\\
         \cline{3-11}
         & & LR & $0.0005$ & $0.001$ & $0.0005$ & $0.0005$ & $0.0005$ & $0.0005$ & $0.0005$ & $0.0005$\\
         \cline{3-11}
         & & Batch & $16$ & $32$ & $16$ & $16$ & $16$ & $16$ & $16$ & $8$\\
         \hline
    \end{tabular}
    \end{adjustbox}
    \vspace{0.5em}
    \caption{\textbf{Hyper-parameters for RoBERTa for MetaTT-5D.} We list here the hyper-parameters that can be used to replicate the results for MetaTT-5D reported in \cref{tab:roberta-combined}.}
    \label{tab:metatt5d_stl_hyperparams}
\end{table}

\paragraph{Hyper-parameter search grid.} We also list the hyper-parameter grids we used to search for the set of hyper-parameters we reported for fine-tuning RoBERTa using MetaTT-4D and MetaTT-5D on GLUE benchmark datasets, in \cref{tab:metatt4d_stl_hyperparams} and \cref{tab:metatt5d_stl_hyperparams}, in \cref{tab:hyperparam-grid}. 
Across both models and methods, we use $0.0$ as weight decay, warmup ratio of $0.6$, and set the sequence length at $256$. 
\begin{table}[h]
\centering
\begin{tabular}{|c|c|c|}
\hline
& MetaTT-4D & MetaTT-5D\\
\hline
\textbf{Hyper-parameter} & \textbf{Values} & \textbf{Values}\\
\hline
\hline
Rank ($r$) & $4, 8, 16, 24, 32, 48, 64$ & Base -- $\{16, 24, 32, 48, 64\}$. Large -- $\{32, 64, 96\}$\\
\hline
Alpha ($\alpha$) & $0.5, 4$ & $0.5, 4$ \\
\hline
Learning Rate ($\eta$) & $1\times10^{-3}$, $5\times10^{-4}$ & $1\times10^{-3}$, $5\times10^{-4}$ \\
\hline
Batch Size & $8, 16, 32$ & $8, 16, 32$\\ \hline 
\end{tabular}
\vspace{0.5em}
\caption{Hyper-parameter grid used for \rbase and \rlarge on GLUE benchmark datasets to fine-tune with MetaTT-4D and MetaTT-5D PEFT adapters.}
\label{tab:hyperparam-grid}
\end{table}

\paragraph{Hyper-parameter search for Llama.} Since fine-tuning on Llama models is computationally more demanding, we restrict the search of hyper-parameters over coarser grids in conjunction with some heuristics. Precisely, we perform a heuristic search over the grid spanned by the TT-ranks $r\in\{8,16,32,64,128,256\}$, alpha values $\alpha \in \{1.0, 2.0, 3.0\}$, learning rates $\eta \in \{1e-4, 2e-4, 5e-4\}$, and over two epochs. All Llama results were obtained by initializing the two middle cores as Gaussians with std$=0.2$ and mean 0, and the right core being set to identity (the left core being set to zero). For MetaTT and other baselines we use AdamW as optimizer along with a linear scheduler, a warmup ratio of 0.06, and effective batch size of 32.

\paragraph{Hyper-parameters for MTL.} We used a fixed learning rate of $5e-4$, and a weight decay schedule of $0.0$ for LoRA and the variants of MetaTT.

\subsection{Baselines}

Several of the baselines reported in \cref{tab:roberta-combined} had extensively reported the set of hyper-parameters used to benchmark against LoRA: VeRA \cite[Tables 8, 9]{kopiczko2023vera}, LoRETTA \cite[Tables 12, 13]{yang2024loretta}, LoRTA \cite[Tables 11, 12]{hounie2024lorta}, LoTR \cite[\S D]{bershatsky2024lotr}, MTL-LoRA and MoE-LoRA \cite[Table 7]{yang2025mtl}). However, except for LoTR and LoRTA, all other methods report accuracy after fine-tuning the weights of both the classifier head and the shared parameters. However, allowing the whole classifier head to be trainable significantly blows up the total number of trainable parameters (e.g., adds about $400$K parameters for \rbase in case of VeRA with sequence length of $1024$), effectively hiding the sole impact of the shareable hyper-parameters. As such we re-run the benchmarking by freezing the classifier heads. We believe that this is necessary for a fair comparison. We also report the new set of hyper-parameters for replicating these results in the following subsections.

\subsubsection{LoRA \texorpdfstring{\protect{\cite{hu2021lora}}}{}}
The reported results for fine-tuning Llama-2 models with LoRA were obtained through hyper-parameter tuning over a grid with ranks $r \in \{8, 16, 32, 64\}$, alpha $\alpha = 2r$, and learning rates $\eta \in \{1e\text{-}1, 1e\text{-}2, 1e\text{-}5\}$. The values presented correspond to the best performance achieved on this grid search over 2 epochs. 

For RoBERTa, we directly report the results from \cite{bershatsky2024lotr}.
\subsubsection{VeRA \texorpdfstring{\protect{\cite{kopiczko2023vera}}}{}}
For each of the GLUE benchmark datasets we use weight decay $0.0$ and warmup ratio $0.06$. Furthermore, to be consistent with our other experiments, for \rbase and \rlarge, we use a sequence length of $256$. We tried different batch sizes from the set $\{4,8,16,32\}$ out of which $32$ consistently worked well across all of the datasets. 
The best performing learning rates for both the models and corresponding datasets are reported in \cref{tab:vera_hyperparams}. To find these learning rates, we searched in the range $[0.0001, 0.1]$ across all datasets. Finally, consistent with the experiments in \cite{kopiczko2023vera}, we set VeRA rank for \rbase as $1024$, and for \rlarge as $256$.
\begin{table}[h]
    \centering
    \begin{tabular}{|l!{\vrule width 1.5pt}c|c|c|c|c|c|c|c|}
        \hline
         \textbf{Model} & 
          \textbf{CoLA} & \textbf{MNLI} & \textbf{MRPC} & \textbf{QNLI} & \textbf{QQP} & \textbf{RTE} & \textbf{SST2} & \textbf{STS-B} \\
         \hline
         \rbase & $0.005$ & $0.0008$ & $0.01$ & $0.015$ & $0.025$ & $0.004$ & $0.01$ & $0.003$\\
         \hline
         \rlarge & $0.009$ & $0.01$ & $0.004$ & $0.006$ & $0.01$ & $0.005$ & $0.01$ & $0.003$\\
         \hline
    \end{tabular}
    \vspace{0.5em}
    \caption{\textbf{Learning rates for VeRA experiments.} Since we only train the attention layers and keep the classifier weights frozen, we report the learning rates to fine-tune \rbase and \rlarge on the GLUE benchmark datasets.}
    \label{tab:vera_hyperparams}
\end{table}

For Llama experiments we perform a coarser grid search given runs are substantially more expensive. For each of the two models we searched over the grid formed by ranks $r \in \{256, 1024\}$ and learning rates $\eta \in \{2e-4,5e-4\}$. We picked the best out of these parameters over the span of two epochs. 

Other than rank and learning rate, for both RoBERTa and Llama experiments we use the default set of hyper-parameters from HuggingFace's implementation of VeRA.

\subsubsection{LoRETTA \texorpdfstring{\cite{yang2024loretta}}{}}

Similar to the original paper, for LoRETTA$_{adp}$ we tried on two batch sizes $16, 32$ and found $32$ to work best in all tasks of the GLUE suite. The bottleneck dimension was set at $64$. Adapter dropout was set to $0$, and the scaling parameter ($\alpha$) was set at $1.0$. 
Weight decay was set to $0.01$ and sequence length was set at $256$ for both the methods. Furthermore, among $2,5,10,20$ tensor ranks, $5$ had the right balance of parameters and performance across GLUE tasks for both the methods.
The learning rates used across the tasks, method, and the models are reported in \cref{tab:LoRETTA_hyperparams}. Each of the dataset was trained on $20$ epochs, except MNLI and QQP which were trained for $10$ epochs.

\paragraph{Tensor decomposition.} Following \cite{yang2024loretta}, each tensorized adapter projection (bottleneck $b=64$, constant TT-rank $r=5$) reshapes its $d\cdot b$ entries into the mode shapes
\begin{equation}
    [8,8,12,8,8] \quad (d=768,\ \text{base}), \qquad [4,8,8,8,8,4] \quad (d=1024,\ \text{large}),
\end{equation}
where $d$ is the hidden size. Together with the adapter placement, these shapes determine the trainable-parameter counts reported in \cref{tab:roberta-combined}; we refer the reader to \cite{yang2024loretta} and the accompanying code for the full parameter accounting.

\begin{table}[h]
    \centering
    \textbf{LoRETTA$_{adp}$} \\
    \begin{tabular}{|l@{\hspace{1.5pt}\vrule width 1.5pt\hspace{1.5pt}}c|c|c|c|c|c|c|c|}
        \hline
         \textbf{Model} & 
         \textbf{CoLA} & \textbf{MNLI} & \textbf{MRPC} & \textbf{QNLI} & \textbf{QQP} & \textbf{RTE} & \textbf{SST2} & \textbf{STS-B} \\
         \hline
         RoBERTa$_{\text{Base}}$ & $1e-3$ & $4e-4$ & $8e-4$ & $2e-3$ & $3e-3$ & $2e-3$ & $3e-4$ & --(--)\\
         \hline
    \end{tabular}\\
    \vspace{0.25em}
    \textbf{LoRETTA$_{rep}$}\\
    \begin{tabular}{|l@{\hspace{1.5pt}\vrule width 1.5pt\hspace{1.5pt}}c|c|c|c|c|c|c|c|}
        \hline
         \textbf{Model} & 
         \textbf{CoLA} & \textbf{MNLI} & \textbf{MRPC} & \textbf{QNLI} & \textbf{QQP} & \textbf{RTE} & \textbf{SST2} & \textbf{STS-B} \\
         \hline
         RoBERTa$_{\text{Base}}$ & $5e-4$ & $1e-4$ & $7e-4$ & $1e-3$ & --(--) & $4e-4$ & $3e-4$ & --(--)\\
         \hline
    \end{tabular}
    \vspace{0.5em}
    \caption{\textbf{Learning rates for LoRETTA$_\text{adp}$ experiments.} Since we only train the attention layers and keep the classifier weights frozen, we report the learning rates to fine-tune RoBERTa$_{\text{Base}}$ and RoBERTa$_{\text{Large}}$ on the GLUE benchmark datasets.}
    \label{tab:LoRETTA_hyperparams}
\end{table}

\subsubsection{LoTR \texorpdfstring{\cite{bershatsky2024lotr}}{}}

For RoBERTa experiments we directly quote the results of LoTR from the original work. For Llama models we search exhaustively over the grid spanned by the learning rates $\eta \in \{2e-4, 5e-4\}$ and ranks $r \in \{16, 64\}$. The choice of scaling factor $\alpha$ is set to $2.0$ and we initialize the cores analogously to MetaTT: middle core drawn from a Gaussian with std$=0.2$ and mean 0, and right core (output dimension leg) set to identity. We report the best accuracy on this grid over two epochs.

\subsubsection{LoRTA \texorpdfstring{\cite{hounie2024lorta}}{}}

We chose to rerun GLUE experiments for LoRTA since in \cite[Table 2]{hounie2024lorta} only the best results are reported. For completeness, we restate the hyperparamter grids used to report results in \cite{hounie2024lorta} for GLUE baselines in \cref{tab:lorta-deets}. We train the models on the $3$ seeds mentioned at the beginning of this section. For COLA, MRPC, STS-B, and RTE tasks we fine-tune RoBERTa based models for $20$ epochs, and for MNLI, SST2, QNLI, and QQP tasks we fine-tune RoBERTa based models for $10$ epochs. Similar to other experiments we set sequence length at $256$. Weight decay is set at $0.0$ for fine-tuning both models. The learning rate grid is $[5e-4, 1e-3, 2e-3, 3e-3, 4e-3, 5e-3, 6e-3, 7e-3, 1e-2, 1.5e-2, 2e-2]$ for \rbase, and $[1e-3, 5e-3, 7e-3, 8e-3, 9e-3, 1e-2, 2e-2]$ for \rlarge. The final learning rates used are reported in \cref{tab:lorta-lrs}.
\begin{table}[h]
    \centering
    \begin{tabular}{|c|c|c|}
         \hline
         \textbf{Hyper-parameter} & \textbf{\rbase} & \textbf{\rlarge} \\
         \hline
         $\alpha$ & $[0.5, 1.0, 2.0, 8.0]$ & $[0.5, 1.0, 2.0, 8.0]$\\
         Scheduler & Linear & Linear\\
         Optimizer & AdamW & AdamW\\
         Batch size & $[32, 64]$ & $[32, 64]$\\
         Warmup ratio & $0.06$ & $0.06$\\
         \hline
    \end{tabular}
    \caption{\textbf{Hyper-parameter configurations for \rbase and \rlarge for LoRTA.} We restate and update some of the hyper-parameters from \cite{hounie2024lorta} for completeness.}
    \label{tab:lorta-deets}
\end{table}
\begin{table}[h]
    \centering
    \begin{tabular}{|c|c|c|c|c|c|c|c|c|}
         \hline
         \textbf{Model} & \textbf{COLA} & \textbf{MNLI} & \textbf{MRPC} & \textbf{QNLI} & \textbf{QQP} & \textbf{RTE} & \textbf{SST2} & \textbf{STS-B}\\
         \hline
         \rbase & $1e-2$ & $1e-2$ & $1e-2$ & $1e-2$ & $1.5e-2$ & $1e-2$ & $4e-3$ & $4e-3$\\
         \rlarge & $1e-2$ & $1e-2$ & $1e-2$ & $1e-2$ & $8e-3$ & $2e-2$ & $1e-2$ & $2e-2$\\
         \hline
    \end{tabular}
    \caption{\textbf{Final learning rates used for fine-tuning \rbase and \rlarge using LoRTA.} We state the final learning rates used to fine-tune RoBERTa based pre-trained models on the GLUE tasks.}
    \label{tab:lorta-lrs}
\end{table}

\subsubsection{MoE-LoRA \texorpdfstring{\cite{liu2024moe}}{} and MTL-LoRA \texorpdfstring{\cite{yang2025mtl}}{}}

The codebase for MTL-LoRA\footnote{\url{https://github.com/pUmpKin-Co/MTL-LoRA}} had implementation only for Llama based models. As such we adapted this codebase for experimenting MTL on RoBERTa based models for fine-tuning on some of the GLUE tasks. For both MoE-LoRA and MTL-LoRA, and both RoBERTa based models, we train for $20$ epochs. The number of experts were set to $4$ for MoE-LoRA (used a grid of $[4,8]$, $4$ worked reliably well), and rank $r$ of LoRA was set to $4$ (grid used was $[4,8]$, $4$ gave the best balance between number of trainable parameters and performance) for both methods. We set a batch size of $32$ across both methods and the learning rate grid used was $[1e-4, 2e-4, 3e-4, 5e-4, 7e-4, 9e-4, 1e-3]$ for both the methods. The final learning rates used are reported in \cref{tab:mtl-lora-lrs}.
\begin{table}[h]
    \centering
    \begin{tabular}{|c|c|c|c|}
         \hline
         \textbf{Tasks} & \textbf{Model} & \textbf{MoE-LoRA} & \textbf{MTL-LoRA} \\
         \hline
         \multirow{2}{*}{MRPC, RTE, CoLA} & \rbase & $7e-4$ & $5e-4$\\
         & \rlarge & $9e-4$ & $5e-4$\\
         \hline
         \multirow{2}{*}{MRPC, RTE, CoLA, QNLI} & \rbase & $1e-3$ & $7e-4$\\
         & \rlarge & $1e-3$ & $3e-4$\\
         \hline
    \end{tabular}
    \caption{\textbf{Final learning rates for MTL experiments using methods from \cite{liu2024moe, yang2025mtl}.} In this table we report the final learning rates used to generate the baseline results in \cref{tab:MTL} and \cref{tab:MTL-4-task} using MoE-LoRA and MTL-LoRA.}
    \label{tab:mtl-lora-lrs}
\end{table}

\section{MetaTT Adapter Implementation}\label{app:metatt-codes}

In this section, we give an example of one of the MetaTT adapters (MetaTT-4D) using pseudo-code written in python.

\paragraph{Configuration.} We start by stating the configuration class of MetaTT-4D in \cref{alg:metatt4d-config}. The different inputs to the class function are -- \texttt{rank}: a $1\times 3$ array (e.g., $[8, 8, 8]$) corresponding to the different ranks of the tensor-train, \texttt{alpha}: scaling factor for the adapter, \texttt{target\_modules}: type of matrices to be fine-tuned using the algorithm, and \texttt{use\_bias}: flag to choose whether to add bias as a parameter.
\begin{algorithm}[ht]
    \caption{MetaTT-4D configuration class file}\label{alg:metatt4d-config}
    \begin{lstlisting}[style=mystyle]
    class MetaTT4DConfig:
        def __init__(self, ranks, alpha=1.0, target_modules=["query", "key", "value"], use_bias=False):
            self.ranks = ranks
            self.alpha = alpha  # scaling factor for the adapter
            self.target_modules = target_modules
            self.use_bias = use_bias
    \end{lstlisting}
\end{algorithm}

\paragraph{Adapter.} We then list an example of an adapter for MetaTT-4D in \cref{alg:metatt4d-adapter}. In the attached pseudo-code, the $\mathcal{G}_1$ core is initialized to zero, the $\mathcal{G}_2,~\mathcal{G}_3,~\mathcal{G}_4$ cores are initialized as the identity matrix. One of the cores is generally set to $\mathbf{0}$-tensor so that output of the corresponding adapter is zero at the beginning of the training similar to \cite{hu2021lora}.
\begin{algorithm}[ht]
    \caption{MetaTT-4D adapter}\label{alg:metatt4d-adapter}
    \begin{lstlisting}[style=mystyle]
    class MetaTT4DAdapter(nn.Module):
        def __init__(self, hidden_dim: int, num_layers: int, tt_config: MetaTT4DConfig):
            super().__init__()
            self.hidden_dim = hidden_dim
            self.num_layers = num_layers
            self.tt_config = tt_config
            self.num_projs = len(tt_config.target_modules)

            # initialize the tensor cores
            self.G1 = Parameter(torch.empty(self.hidden_dim, \
                tt_config.ranks[0]), requires_grad=True)
            
            self.G2 = ParameterList[nn.init.eye_(Parameter(torch.zeros\
                (self.tt_config.ranks[0], self.tt_config.ranks[1]),\
                requires_grad=True)) for _ in range(self.num_layers)]
            
            self.G3 = ParameterList[nn.init.eye_(Parameter(torch.empty\
                (self.tt_config.ranks[1], self.tt_config.ranks[2]),\
                requires_grad=True)) for _ in range(self.num_projs)]
                
            self.G4 = Parameter(torch.empty(tt_config.ranks[2],\
                self.hidden_dim), requires_grad=True)
    
            nn.init.zeros_(self.G1)
            nn.init.eye_(self.G4)
    \end{lstlisting}
\end{algorithm}

\paragraph{Linear adapter and forward function.} \cref{alg:metatt4d-adapter} can be used to create a linear adapter and the corresponding forward function as shown in \cref{alg:metatt4d-forward}. The inputs to this adapter are -- \texttt{original\_layer}: the layers of the pre-trained model, \texttt{tt\_config}: the corresponding initialized MetaTT configuration class, \texttt{M1, M2, M3, M4}: the tensor core slices along each of the four dimensions (corresponding to matrices). During the forward pass, the input batch is first passed through the original layer that is frozen. This batch is also multiplied by \texttt{M1} trough \texttt{M4}, and scaled by $\alpha$. Note that this order of matrix multiplication will be optimal if the rank is smaller than the batch size. The outputs of the original and the adapter are then added and returned.
\begin{algorithm}[ht]
    \caption{MetaTT-4D linear adapter}\label{alg:metatt4d-forward}
    \begin{lstlisting}[style=mystyle]
        class MetaTT4DLinearAdapter(nn.Module):
            def __init__(self, original_layer: nn.Module, tt_config: MetaTT4DConfig, M1, M2, M3, M4):
                super().__init__()
                # set requires_grad to False for original weights
                self.original_layer = original_layer.\
                                    requires_grad_(False)
                self.tt_config = tt_config
                self.M1 = M1 
                self.M2 = M2 
                self.M3 = M3 
                self.M4 = M4
        
            def forward(self, X: torch.tensor) -> torch.Tensor:
                original_output = self.original_layer(X)
                return original_output + self.tt_config.alpha * (((X @ self.M1) @ self.M_2) @ self.M3) @ self.M4 
    \end{lstlisting}
\end{algorithm}

\paragraph{PEFT model.} Finally, the pseudo-code for developing our model with PEFT philosophy so that it can be used as a drop-in within any model quickly is given in \cref{alg:metatt4d-peft}. The inputs to this algorithm is just the pre-trained model and the configuration class for MetaTT. \cref{alg:metatt4d-peft} uses the cores initialized in \cref{alg:metatt4d-adapter} to set up the trainable layer using \cref{alg:metatt4d-forward}. Finally, these layers are set-up such that one can dynamically update them during runtime. Once all the layers are initialized correctly, the model is assigned to the corresponding training device and returned.
\begin{algorithm}[ht]
    \caption{MetaTT-4D PEFT model for RoBERTa}\label{alg:metatt4d-peft}
    \begin{lstlisting}[style=mystyle]
    def get_meta_tt_4d_model(model, config):
        # set pre-trained model weights to be non-trainable
        for param in model.parameters():
            param.requires_grad = False 
        # grab device ID from model
        device = model.device
    
        num_layers = model.config.num_hidden_layers
        hidden_dim = model.config.hidden_size
        
        # initialize the MetaTT adapter
        meta_tt_adapter = MetaTT4DAdapter(hidden_dim, num_layers,\
                                          config)

        # go through each layer and corresponding projection 
        # matrices of the roberta model
        for layer_idx, layer in \
                enumerate(model.roberta.encoder.layer):
            for proj_idx, proj_matrix in \
                enumerate(config.target_modules):
                if proj_matrix in ("query", "key", "value"):
                    original_layer = layer.attention.self
                    original_matrix = getattr(original_layer,\
                                              proj_matrix)
                elif proj_matrix == "dense":
                    original_layer = layer.attention.output
                    original_matrix = getattr(original_layer,\
                                              proj_matrix)
                else:
                    raise ValueError(f\
                        "Unexpected proj_matrix value: {proj_matrix}")    
                        
                # set-up the MetaTT layer
                meta_tt_layer = MetaTT4DLinearAdapter(\
                                original_matrix, config, 
                                meta_tt_adapter.G1, 
                                meta_tt_adapter.G2[layer_idx],
                                meta_tt_adapter.G3[proj_idx],
                                meta_tt_adapter.G4)
    
                setattr(original_layer, proj_matrix, meta_tt_layer)
                
        return model.to(device)
    \end{lstlisting}
\end{algorithm}

\section{Limitations and broader impacts}

\subsection{MetaTT-5D Results}\label{sec:5d-results}
\rev{While the MetaTT framework naturally extends to 5D by further decomposing the output dimension along the head dimension and number of heads, we found that in practice MetaTT-4D is the more reliable variant. MetaTT-5D can be more parameter-efficient than MetaTT-4D at low ranks (when $r < D/H(1-1/H)$), but exhibits greater training instability, particularly at higher ranks. We report MetaTT-5D results on the GLUE benchmark in \cref{tab:roberta-5d} for completeness. On \rbase, MetaTT-5D r=64 achieves competitive results but uses more parameters than MetaTT-4D r=24 for similar accuracy. On \rlarge, MetaTT-5D r=64 suffers from a catastrophic failure on STS-B (65(23)\%), illustrating the instability issue. Based on these observations, we recommend MetaTT-4D as the default variant.}

\begin{table}[h]
\centering
\begin{adjustbox}{max width=\textwidth}
\begin{tabular}{|c|lcc*{8}{c}|}
\hline
\multirow{2}{*}{} & \multirow{2}{*}{\textbf{Method}} & \multirow{2}{*}{\shortstack{\textbf{Param}\\$\times 10^3$}} & \multirow{2}{*}{\textbf{Rank}} & \multicolumn{8}{c|}{\textbf{Metric (\%)}} \\
& & & &  \textbf{CoLA} &  \textbf{MNLI} &  \textbf{MRPC} &  \textbf{QNLI} &  \textbf{QQP} &  \textbf{RTE} &  \textbf{SST2} &  \textbf{STS-B} \\
\hline
\multirow{2}{*}{\rotatebox{90}{\small Base}}
& \multirow{2}{*}{MetaTT-5D} & $20$ & $16$ & $50(2)$ & $84.0(1)$ & $88.2(5)$ & $89.7(1)$ & $87.0(1)$ & $73.6(8)$ & $93.2(3)$ & $88.6(3)$ \\
& & $160$ & $64$ & $60.4(3)$ & $85.8(1)$ & $88.8(2)$ & $91.3(2)$ & $88.3(1)$ & $74.8(8)$ & $93.8(1)$ & $89.5(4)$ \\
\hline
\multirow{2}{*}{\rotatebox{90}{\small Large}}
& \multirow{2}{*}{MetaTT-5D} & $78$ & $32$ & $63.2(5)$ & $89.8(1)$ & $89.6(1)$ & $93.4(0)$ & $88.7(1)$ & $73(7)$ & $94.6(0)$ & $91.5(2)$ \\
& & $242$ & $64$ & $64.9(2)$ & $90.0(1)$ & $90.0(4)$ & $93.4(1)$ & $89.1(1)$ & $74(9)$ & $95.2(1)$ & $65(23)$\footnotemark \\
\hline
\end{tabular}
\end{adjustbox}
\caption{\textbf{MetaTT-5D results on GLUE.} MetaTT-5D further decomposes the output dimension along the head dimension. While it can achieve competitive results, it exhibits greater training instability than MetaTT-4D, particularly at higher ranks on \rlarge (e.g., STS-B).}
\label{tab:roberta-5d}
\end{table}

\paragraph{Limitations.}
MetaTT is sensitive to parameter initialization. We have observed that for certain choice of hyper-parameters MetaTT would fail to train. \rev{This is more prevalent when the number of TT cores increases (e.g., in 5D decompositions), which is why we recommend MetaTT-4D as the default variant.} Finding better initialization heuristics would improve the robustness of MetaTT.
\rev{Additionally, the DMRG-inspired rank schedule is sensitive to the training regime. While multi-step DMRG sweeps are effective on smaller-scale settings (RoBERTa/GLUE; Commonsense15k with Llama-2-7b), we found that at the Commonsense170k scale, multi-step compression is harmful---reducing average accuracy to near zero-shot levels---whereas a single well-timed compression step is benign. The optimal schedule (number of steps, timing, rank trajectory) thus appears to be task- and scale-dependent, and developing principled guidelines for choosing DMRG schedules remains an open problem.}

\paragraph{Broader Impacts.}
The development of a reparameterization adapter using tensor trains and DMRG-inspired techniques offers significant potential in advancing compressed adapter fine-tuning as well as model training. By leveraging these methods, models can be compressed as they are being trained, significantly reducing the final parameter count while maintaining high accuracy.
This leads to ultra-compressed models while training, and compressed adapters while fine-tuning. The ability to compress models as they are training ensures that the compression does not compromise the model's performance, as there are opportunities for correction during the training process itself. 

Further, our work opens up new possibilities for deploying advanced scalable models in resource-constrained settings -- where our techniques could allow for maintaining a high accuracy during training as opposed to approaches where a model is compressed post-training.

\end{document}